%% file: main.tex
\documentclass[twocolumn]{article}

\usepackage{booktabs}
\usepackage{wrapfig}
\usepackage{multicol}
\usepackage[noadjust]{cite}

\usepackage{amsmath,amssymb,amsfonts}
\usepackage{graphicx}
\usepackage{textcomp}
\usepackage{times}
\usepackage{epsfig}
\usepackage{subcaption}
\usepackage{array,multirow}
\usepackage{float}
\usepackage{algorithm2e}
\usepackage{comment}
\usepackage[table,xcdraw]{xcolor}
\usepackage{hyperref}
\usepackage{soul}
\usepackage{authblk}
\usepackage[margin=0.7in]{geometry}

\def\BibTeX{{\rm B\kern-.05em{\sc i\kern-.025em b}\kern-.08em
    T\kern-.1667em\lower.7ex\hbox{E}\kern-.125emX}}

\graphicspath{{./images/}}
\DeclareGraphicsExtensions{.png , .jpg, .jpeg, .eps}

\newcommand{\ie}{\emph{i.e.}}
\newcommand{\eg}{\emph{e.g.}}

\newcommand{\Sect}[1]{Sect. \ref{sec:#1}}

\newcommand{\citetal}[2]{#1 et al. \cite{#2}}

\newcommand{\lab}{l}
\newcommand{\ulab}{u}
\newcommand{\dataset}{\mathcal{X}}
\newcommand{\bblabel}{\mathcal{Y}}
\newcommand{\image}{\mbox{I}}
\newcommand{\plab}{\hat{l}}
\newcommand{\XL}{\dataset^{\lab}}
\newcommand{\IL}{\image^{\lab}_i}
\newcommand{\YL}{\bblabel^{\lab}_i}
\newcommand{\XU}{\dataset^{\ulab}}
\newcommand{\IU}{\image^{\ulab}_i}
\newcommand{\XP}{\dataset^{\plab}}
\newcommand{\IP}{\image^{\plab}_i}
\newcommand{\YP}{\bblabel^{\plab}_i}
\newcommand{\CNN}{\Phi}
\newcommand{\OD}{\phi}
\newcommand{\HP}{\mathcal{H}}
\newcommand{\HPStop}{\HP_{{\tiny stp}}}
\newcommand{\HPSeq}{\HP_{{\tiny seq}}}
\newcommand{\HPOD}{\HP_{\CNN}}
\newcommand{\HPCoT}{\HP_{sl}}
\newcommand{\HPST}{\HP_{sl}}

\newcommand{\STOP}[4]{\mbox{{\small Stop?}}({#1},{#2},{#3},{#4})}
\newcommand{\Select}[3]{\mbox{{\small Select}}({#1},{#2},{#3})}
\newcommand{\TD}[4]{\mbox{{\small TrainDetector}}({#1},{#2},{#3},{#4})}
\newcommand{\RD}[3]{\mbox{{\small RunDetector}}({#1},{#2},{#3})}
\newcommand{\Fuse}[2]{\mbox{{\small Fuse}}({#1},{#2})}
\newcommand{\Rand}[3]{\mbox{{\small Rand}}({#1},{#2}{#3})}

\newcommand{\model}{\phi}
\newcommand{\domain}{\mathcal{D}}
\newcommand{\train}{{tr}}
\newcommand{\test}{{tt}}
\newcommand{\source}{{S}}
\newcommand{\target}{{T}}
\newcommand{\XS}{\dataset_{\source}}
\newcommand{\XStr}{\dataset_{\source}^{\train}}
\newcommand{\XStrL}{\dataset_{\source}^{\lab,\train}}
\newcommand{\XStt}{\dataset_{\source}^{\test}}
\newcommand{\XT}{\dataset_{\target}}
\newcommand{\XTtrL}{\dataset_{\target}^{\lab,\train}}

\newcommand{\MS}{\model_{\source}}
\newcommand{\MT}{\model_{\target}}
\newcommand{\DS}{\domain_{\source}}
\newcommand{\DT}{\domain_{\target}}

\newcommand{\Xtrain}{\dataset^{\train}}
\newcommand{\XLtrain}{\dataset^{\lab,\train}}

\newcommand{\XPtrain}{\dataset^{\plab,\train}}
\newcommand{\Xtest}{\dataset^{\test}}

\newcommand{\Vds}{\mathcal{V}}
\newcommand{\Kds}{\mathcal{K}}
\newcommand{\Wds}{\mathcal{W}}
\newcommand{\Kdstrain}{\Kds^{\train}}
\newcommand{\Wdstrain}{\Wds^{\train}}
\newcommand{\Kdstest}{\Kds^{\test}}
\newcommand{\Wdstest}{\Wds^{\test}}
\newcommand{\GAN}{\mathcal{G}}
\newcommand{\VGds}{\Vds_{\GAN}}
\newcommand{\GANKds}{{\GAN}_{\Kds}}
\newcommand{\VGKds}{\Vds_{\GANKds}}
\newcommand{\GANWds}{{\GAN}_{\Wds}}
\newcommand{\VGWds}{\Vds_{\GANWds}}

\newcommand{\centeredcell}[1]{\begin{tabular}{l} #1 \end{tabular}}

\providecommand{\keywords}[1]{\textbf{\textit{Index terms---}} #1}

\newcommand\blfootnote[1]{%
  \begingroup
  \renewcommand\thefootnote{}\footnote{#1}%
  \addtocounter{footnote}{-1}%
  \endgroup
}

\begin{document}

\title{Co-training for On-board Deep Object Detection}

\author[1]{\uppercase{Gabriel Villalonga}}
\author[1]{\uppercase{Antonio M. L\'opez}}

\affil[1]{Computer Vision Center (CVC) and Computer Science Dpt., Universitat Aut\`onoma de Barcelona (UAB), Spain.}


\maketitle

\begin{abstract}
Providing ground truth supervision to train visual models has been a bottleneck over the years, exacerbated by domain shifts which degenerate the performance of such models. This was the case when visual tasks relied on handcrafted features and shallow machine learning and, despite its unprecedented performance gains, the problem remains open within the deep learning paradigm due to its data-hungry nature. Best performing deep vision-based object detectors are trained in a supervised manner by relying on human-labeled bounding boxes which localize class instances ({\ie} objects) within the training images. Thus, object detection is one of such tasks for which human labeling is a major bottleneck. In this paper, we assess co-training as a semi-supervised learning method for self-labeling objects in unlabeled images, so reducing the human-labeling effort for developing deep object detectors. Our study pays special attention to a scenario involving domain shift; in particular, when we have automatically generated virtual-world images with object bounding boxes and we have real-world images which are unlabeled. Moreover, we are particularly interested in using co-training for deep object detection in the context of driver assistance systems and/or self-driving vehicles. Thus, using well-established datasets and protocols for object detection in these application contexts, we will show how co-training is a paradigm worth to pursue for alleviating object labeling, working both alone and together with task-agnostic domain adaptation.
\end{abstract}


\blfootnote{The authors acknowledge the financial support received for this work from the Spanish TIN2017-88709-R (MINECO/AEI/FEDER, UE) project. Antonio acknowledges the financial support to his general research activities given by ICREA under the ICREA Academia Program. The authors acknowledge the support of the Generalitat de Catalunya CERCA Program as well as its ACCIO agency to CVC's general activities.}


\input{sections/introduction}

\input{sections/relatedwork}

\input{sections/method}

\input{sections/experiments}

\input{sections/conclusion}

\bibliographystyle{ieee}
\bibliography{references}

\end{document}

%% file: sections/introduction.tex
\section{Introduction}
\label{sec:introduction}

Since more than two decades ago, machine learning (ML) has been the enabling technology to solve computer vision tasks. In the last decade, traditional ML, {\ie} based on relatively shallow classifiers and hand-crafted features, has given way to deep learning (DL). Thanks to DL models based on convolutional neural networks (CNNs), DL approaches significantly outperformed traditional ML in all kinds of computer vision tasks, such as image classification, object detection, semantic segmentation, etc. A major key for the usefulness of DL models is to train them in a supervised way. In other words, the raw data (still images and videos) need to be supplemented by ground truth; nowadays, usually collected via crowd-sourced labeling. In practice, due to the data-hungry nature of CNNs, data labeling is considered a major bottleneck. Therefore, approaches to minimize labeling effort are of high interest; or put in another way, approaches to automatically leverage the large amounts of available unlabeled raw data must be pursued. Thus, we can find different families of algorithms, which address human-labeling minimization under different working assumptions. 

For instance, \emph{active learning} (AL) approaches \cite{Abramson:2005, Settles:2012, Roy:2018} assume an initial model and an unlabeled dataset to be labeled by a human (oracle) following an interactive procedure. In particular, the current model processes the unlabeled data providing so-called \emph{pseudo-labels} (at image, object, or pixel level); then, these results are inspected, either automatically or directly by the oracle, to select the next data to be labeled. Afterwards, the model is trained again and the process repeated until fulfilling a stop criterion. AL assumes that with less labeling budget than the required to have the oracle labeling data at random, one can train models that perform at least similarly. In practice, it is usually hard to clearly outperform the oracle's random labeling strategy. In contrast to AL, other approaches do not assume human oracles in the labeling loop. For instance, this is the case of semi-supervised learning (SSL) algorithms \cite{Chapelle:2006, Triguero:2015, Engelen:2020}, which assume the availability of a large number or raw unlabeled data together with a relatively small number of labeled data. Then, a model must be trained using the unlabeled and labeled data (without human intervention), with the goal of being more accurate than if only the labeled data were used. The so-called \emph{self-training} and \emph{co-training} algorithms, which are of our main interest in this paper, fall in the SSL paradigm \cite{Engelen:2020}. In AL and SSL, the most common idea is to efficiently label the unlabeled data, either via an oracle (AL) or automatically (SSL). \emph{Self-supervised learning} (SfSL) follows a different approach which consists in providing supervision in the form of additional (relatively simple) tasks, known as pretext tasks, for which automatic ground truth can be easily generated ({\eg} solving jigsaw puzzles \cite{gidaris:2018,Kim:2018,Kolesnikov:2019}). 

Within these approaches, the most classical assumption is that the labeled and the unlabeled data are drawn from the same distribution and one aims at using the unlabeled data (via annotations or pretext tasks) to solve the same tasks for which the labeled data was annotated. However, in practice, we may require to solve new tasks, leading to \emph{transfer learning} (TL) \cite{Weiss:2016,Zhuang:2020}, or solving the same tasks in a new domain, leading to \emph{domain adaptation} (DA) \cite{Csurka:2017,Wang-Deng:2018,Wilson:2020}. Beyond specific techniques to tackle TL or DA, we can leverage solutions/ideas from AL, SSL, or SfSL. For instance, AL \cite{Vazquez:2014}, SSL \cite{Zou:2018,Kim:2019}, and SfSL \cite{Xu:2019} algorithms have been used to address DA problems. In this paper, we use SSL (comparing self-training and co-training) to address DA too.

From the application viewpoint, in this paper, we focus on vision-based on-board deep object detection. Note that this is a very relevant visual task for driving, since detecting objects ({\eg} vehicles, pedestrians, etc.) along the route of a vehicle is a key functionality for perception-decision-action pipelines in the context of both advanced driver assistance systems (ADAS) and self-driving vehicles. Moreover, nowadays, most accurate vision-based models to detect such objects are based on deep CNN architectures \cite{Ren:2017,Liu:2016,Redmon:2017,Liu:2020}. In addition, in this application context, it is possible to acquire innumerable quantities of raw images, for instance, from cameras installed in fleets of cars. Thus, methods to minimize the effort of manually labeling them are of great relevance.

Among the SSL approaches, co-training \cite{Blum:1998,Guz:2007} has been explored for deep image classification \cite{Chen:2011,Qiao:2018} with promising results. However, up to the best of our knowledge, it remains unexplored for deep object detection. Note that, while image classification aims at assigning image-level class/attribute labels, object detection is more challenging since it requires to localize and classify objects in images, {\ie} placing a bounding box (BB) per object, together with the class label assigned to it. Moreover, object detection from on-board images, {\ie} detection of vehicles, pedestrians, etc., is especially difficult since, to the inherent intra-class differences ({\eg} due to vehicle models, pedestrian clothes, etc.), acquisition conditions add a vast variability because the objects appear in a large range of distances to the camera (resolution and focus variations), under different illumination conditions (from strong shadows to direct sunlight), and they usually move (blur, occlusion, view angle, and pose variations).  

Originally, co-training relies on the \emph{agreement} of two trained models performing predictions from different \emph{features} (views) of the data \cite{Blum:1998}. These predictions are taken as pseudo-labels to keep improving the models incrementally. Later, prediction \emph{disagreement} was shown to improve co-training results in applications related to natural language processing \cite{Guz:2007,Tur:2009}. Thus, in this paper, we propose a co-training algorithm inspired by prediction \emph{disagreement}. On the other hand, since co-training was proposed as a SSL method aiming at avoiding the drift problem of self-training \cite{Blum:1998}, which incrementally re-trains a single model from its previously most confident predictions, we also elaborate a strong self-training baseline for our deep object detection problem. 

We assess the effectiveness of the two self-labeling methods, {\ie} self-training and co-training, in two different practical situations. First, following most classical SSL, we will assume access to a small dataset of labeled images, $\XL$, together with a larger dataset of unlabeled images $\XU$; where here the labels are object BBs with class labels. Second, we will address a relevant setting that resorts DA. In particular, we will assume access to a dataset of virtual-world images with automatically generated object labels. Therefore, eventually, $\XL$ can be larger than $\XU$, since labeling these images does not require human intervention; however, $\XU$ is composed of real-world images, thus, between $\XL$ and $\XU$ there is a domain shift \cite{Vazquez:2014}. In this case, we also compare self-labeling techniques with task-agnostic DA, in particular, using GAN-based image-to-image translation \cite{Zhu:2017}. As especially relevant cases, we will focus on on-board detection of vehicles and pedestrians using a deep CNN architecture. Our experiments will show how, indeed, our co-training algorithm is a good SSL alternative for on-board deep object detection.
Co-training clearly boosts detection accuracy in regimes where the size of $\XL$ is just the $5\%/10\%$ of the labeled data used to train an upper-bound object detector. Moreover, under the presence of domain shift, we will see how image-to-image GAN-based translation and 
co-training complement each other, allowing to reach almost upper-bound performances without human labeling.

Hence, the main contributions of this paper are: (1) proposing a co-training algorithm for deep object detection; (2) designing this algorithm to allow addressing domain shift via GAN-based image-to-image translation; (3) showing its effectiveness by developing a strong self-training baseline and relying on publicly available evaluation standards and on-board image datasets. Alongside, we will also contribute with the public release of the virtual-world dataset we generated for our experiments. To show these achievements we organize the rest of the paper as follows. \Sect{relatedwork} reviews the most related works to ours, highlighting commonalities and differences. \Sect{method} details our self-training and co-training algorithms. \Sect{experiments} draws our experimental setting, and discuss the obtained results. Finally, \Sect{conclusion} summarizes the presented work, suggesting lines of continuation.

%% file: sections/relatedwork.tex

\section{Related Work}
\label{sec:relatedwork}
Our main goal is to train a vision-based deep object detector without relying on human labeling. Following our work line \cite{Lopez:2017}, we propose to leverage labeled images from virtual worlds and unlabeled ones from the real world, in such a way that we can automatically label the real-world images by progressively re-training a deep object detector that substitutes human annotators. Note that the labeling step is needed due to the domain shift between virtual and real-world images, otherwise, we could just use the virtual-world images to train object detectors and, afterwards, deploy them in the real world expecting a reliable performance.  
Accordingly, we explore the combination of the co-training idea for self-labeling objects and GAN-based image-to-image translation in the role of task-agnostic DA. On the other hand, our proposal is agnostic to the object detection architecture supporting self-labeling. Thus, in the remaining of this section we review related works on self-labeling and DA. 

\subsection{Self-labeling}
Self-labeling algorithms are examples of SSL \emph{wrappers} \cite{Engelen:2020}, which work as meta-learners for supervised ML algorithms. The starting point consists of a labeled dataset, $\XL$, and an unlabeled dataset, $\XU$, where it is supposed that the cardinality of $\XU$ is significantly larger than the cardinality of $\XL$, and both datasets are drawn from the same domain, $\domain$. The goal is to learn a predictive model, $\model$, whose accuracy would be relatively poor by only using $\XL$, but becomes significantly higher by also leveraging $\XU$. Briefly, self-labeling is an incremental process where $\model$ is first trained with $\XL$; then $\XU$ is processed using $\model$ in a way that the predictions are taken as a new pseudo-labeled dataset $\XP$, which in turn is used to retrain $\model$. The pseudo-labeling/retraining cycle is repeated until reaching some stop criterion, when $\model$ is expected to be more accurate than at the beginning of the process. The main differences between self-labeling algorithms arise from how $\XP$ is formed and used to retrain $\model$ in each cycle.

\paragraph{Self-training}
In self-training, introduced by Yarowsky \cite{Yarowsky:1995} for word sense disambiguation, $\XP$ is formed by collecting the most confident predictions of $\model$ in $\XU$, updating $\XU$ to contain only the remaining unlabeled data. Then, $\model$ is retrained using supervision from $\XL\cup\XP$, {\ie} pseudo-labels are taken as ground truth. Before the DL era, \citetal{Rosenberg}{Rosenberg:2005} already showed promising results when applying self-training to eye detection in face images. More recently, \citetal{Jeong}{Jeong:2019}, proposed an alternative to collect $\XP$ for deep object detection, which consists of adding a consistency loss for training $\model$ as well as eliminating predominant backgrounds. If $\IU$ is an unlabeled image, the consistence loss is based on the idea that $\model(\IU)$ and $\model((\IU)^\Lsh)$, where $()^\Lsh$ stands for horizontal mirroring, must provide corresponding detections. Experiments are conducted on PASCAL VOC and MS-COCO datasets, and results
are on pair with other state-of-the-art methods combining AL and self-training \cite{WangYan:2018}. The reader is referred to \cite{Oliver:2018} for a review on loss-based SSL methods for deep image classification. On the other hand, note that this SSL variations are not agnostic to $\model$, since its training loss is modified. This is also the case in \cite{Lokhande:2020}, where, in the context of deep image classification, the activation functions composing $\model$ must be replaced by Hermite polynomials. In this paper, we use a self-training strategy as SSL baseline for on-board deep object detection, thus, keeping agnosticism regarding  $\model$. 

\paragraph{Co-training}
Co-training was introduced by Blum and Mitchell \cite{Blum:1998} in the context of web-page classification, as alternative to the self-training of Yarowsky \cite{Yarowsky:1995}, in particular, to avoid model drift. In this case, two models, $\model_1$ and $\model_2$, are trained on different conditionally independent features of the data, called \emph{views}, assuming that each view is sufficiently good to learn an accurate model. Each model is trained by following the same idea as with self-training, but in each cycle the data samples self-labeled by $\model_1$ and $\model_2$ are aggregated together. Soon, co-training was shown to outperform other state-of-the-art methods including self-training \cite{Nigam:2000}, and the conditional independence assumption was shown not to be essential in practice \cite{Balcan:2004,Wang:2007,Wang:2010}. Later, in the context of sentence segmentation, \citetal{Guz}{Guz:2007} introduced the \emph{disagreement} idea for co-training, which was refined by Tur \cite{Tur:2009} to jointly tackle DA in the context of natural language processing. In this case, samples self-labeled with high confidence by $\model_i$ but with low confidence by $\model_j$, $i,j\in\{1,2\},i \neq j$, are considered as part of the new pseudo-labeled data in each cycle. In fact, disagreement-based SSL became a subject of study on its own at that time \cite{Zhou:2010}.

Before the irruption of DL, \citetal{Levin}{Levin:2003} applied co-training to detect vehicles in video-surveillance images, so removing background and using different training data to generate different views. More recently, \citetal{Qiao}{Qiao:2018} used a co-training setting for deep image classification, based on several views. Each view corresponds to a different CNN, $\model_i$, trained by including samples generated to be mutually adversarial. The idea is to use different training data for each $\model_i$ to prevent them to prematurely collapse in the same network. This implies to link the training of the $\model_i$'s at the level of the loss function. In this paper, we force the use of different training data for each $\model_i$, without linking their training at the level of loss functions, again, keeping co-training agnostic to the used $\model_i$. Moreover, we address objected detection, which involves not only predicting class instances as in image classification, but also localizing them within the images, so that the background becomes a large source of potential false positives.

Finally, to avoid confusion, it is worth to mention the so-called \emph{co-teaching}, recently introduced by \citetal{Han}{Han:2018}, and its variant \emph{co-teaching+} introduced by \citetal{Yu}{Yu:2019}. These algorithms are designed to address situations with noisy labels on $\XL$, both demonstrated on deep image classification. Indeed, these algorithms stem ideas from  co-training (the classical one from \cite{Han:2018}, the disagreement-based one from \cite{Yu:2019}), however, reproducing the words of \citetal{Han}{Han:2018}, \emph{co-training is designed for SSL, and co-teaching is for learning with noisy (ground truth) labels (LNL); as LNL is not a special case of SSL, we cannot simply translate co-training from one problem setting to another problem setting}.

\subsection{Domain adaptation}
ML algorithms assume that training and testing data are drawn from the same domain, $\domain$. When this is not the case, the trained models suffer from \emph{domain shift}. In other words, we have data, $\XS$, drawn from a \emph{source domain}, $\DS$, as well as data, $\XT$, drawn from a \emph{target domain}, $\DT, \DT\neq\DS$. We can assume that $\XS=\XStr\cup\XStt, \XStr\cap\XStt=\emptyset$, where $\XStr$ is used to train some predictive model $\MS$. It turns out that the prediction accuracy of $\MS$ in $\XStt$ is much higher than in $\XT$, a phenomenon known as domain shift. Addressing this problem is the goal of DA techniques, under the assumption that there is some (unknown) correlation between $\DS$ and $\DT$, since DA is not possible otherwise.

The core idea is to use $\XT$ to obtain a new model, $\MT$, being clearly more accurate than $\MS$ in $\DT$. While doing this, the human-based labeling effort in $\XT$ must be minimized. \emph{Supervised DA} (SDA) assumes access to a relatively small set of labeled target-domain data $\XTtrL\subset\XT$. If we do not have access to $\XStrL$, then the challenge is to leverage from $\MS$ and $\XTtrL$ to obtain $\MT$. Otherwise, we can combine $\XStrL$ and $\XTtrL$ to train $\MT$. In \emph{unsupervised DA} (UDA), $\XT$ is unlabeled; thus, we address the more challenging situations of using $\XT$ with either $\MS$ or $\XStrL$ to train $\MT$. For a review of the DA corpus we advise the reader to consult \cite{Csurka:2017,Wang-Deng:2018,Wilson:2020}. 

In this paper, we assume an UDA setting. Moreover, our source data comes from a virtual world with automatically generated labels, so we have $\XStrL$. Since we aim at assessing self-labeling by co-training to address the UDA problem, we have $\XL=\XStrL$, $\XU=\XT$, thus, $\XP\subseteq\XT$. 

Following this line of work, \citetal{Kim}{Kim:2019} addressed USD for deep object detection by proposing a combination of a weak self-training and a special treatment of backgrounds via a loss component used during the training of the object detector, with PASCAL VOC as $\XS$ and art-style datasets (Clipart1K, Watercolor2K, Comic2K) as $\XT$. \citetal{Zou}{Zou:2019} also use a self-training strategy for UDA in the context of deep image classification and semantic segmentation (including a virtual-to-real setting),
where the core idea is to perform confidence and model regularization of the trained classifiers. Since in our experimental setting we will use Faster R-CNN to obtain $\model$, it is also worth to mention the work by \citetal{Chen}{Chen:2018}, where an UDA method was specifically designed for Faster R-CNN and demonstrated on car detection under a virtual-to-real setting too.
Since Faster R-CNN is a two-stage classifier, the proposed UDA involves an image-level adaptation for the region proposal stage, and an instance-level adaptation for the BB prediction stage. Finally, focusing on traditional ML and \emph{Amazon reviews} datasets, \citetal{Chen}{Chen:2011b} showed that co-training is a promising algorithm for UDA, providing better performance than self-training. 

Accordingly, beyond self-training-style and model-specific strategies for UDA, in this paper, we are interested in assessing co-training as a meta-learning UDA strategy. To the best of our knowledge, this is an under-explored and relevant setting. 
Moreover, even we are going to use Faster R-CNN because its usually outstanding accuracy, our proposal neither is specific for it, nor requires to modify its losses.
In other words, our co-training-based UDA works at the training-data level. This allows to complement it with other UDA working at the same level. In particular, we combine it with GAN-based image-to-image translation, since such task-agnostic approach can transform $\XStrL$ to be more similar to $\XT$ before starting co-training. By using the CycleGAN implementation of \citetal{Zhu}{Zhu:2017}, we will see how, indeed, GAN-based image-to-image translation combined with co-training outperforms the use of each method in isolation.

%% file: sections/method.tex
\section{Methods}
\label{sec:method}

In this section, we detail our self-training and co-training meta-learning proposals as Algorithms \ref{alg:self-training} and \ref{alg:co-training}, respectively. Our main interests is to assess the performance of co-training in vision-based object detection, but we need also to develop a strong self-training baseline. Since they share functional components, we first introduce those and then detail how they are used for both self-training and co-training. Finally, we will see how, depending on the input data, these SSL algorithms can be used even in a context where there is a domain shift between already existing labeled images and the images to be labeled automatically. To refer to both, self-training and co-training indistinctly, we will use the term \emph{self-labeling} in the rest of this section. 

\begin{algorithm}
\SetAlgoLined
\SetKwInOut{Input}{input}\SetKwInOut{Output}{output}
\Input{{\small Labeled images:} $\XL=\{<\IL,\YL>\}$\\ 
       {\small Unlabeled images:} $\XU=\{\IU\}$\\ 
       {\small Object detection architecture:} $\CNN$\\
       {\small $\CNN$ Training hyper-par.:} $\HPOD$\\
       {\small Slf-tr. hyp.-p.:} $\HPST=\{T,N,n,\HPStop[,\HPSeq]\}$\\}
\Output{{\small New labeled images:} $\XP=\{<\IP,\YP>\}$}
$<\XP,k>\,\,\leftarrow\,\,<\emptyset,0>$\;
$\OD \leftarrow \TD{\CNN}{\HPOD}{\XL}{\XP}$\;
$\XP_{new} \leftarrow \RD{\OD}{\XU}{T}$\;
\Repeat{$\STOP{\HPStop}{\XP_{old}}{\XP_{new}}{k\mbox{++}}$}{
$\XP_{old} \leftarrow \XP_{new}$\;
$\XP_{\uparrow} \leftarrow \Select{\uparrow}{n}{\Rand{\XP_{new}}{N}{[,\HPSeq,k]}}$\;
$\XP \leftarrow \Fuse{\XP}{\XP_{\uparrow}}$\;
$\OD \leftarrow \TD{\CNN}{\HPOD}{\XL}{\XP}$\;
$\XP_{new} \leftarrow \RD{\OD}{\XU}{T}$\;
}
$\XP \leftarrow \XP_{new}$\;
\KwRet{$\XP$}\;
\caption{Self-labeling by self-training (see main text).}
\label{alg:self-training}
\end{algorithm}

\begin{algorithm}[!t]
\SetAlgoLined
\SetKwInOut{Input}{input}\SetKwInOut{Output}{output}
\Input{{\small Labeled images:} $\XL=\{<\IL,\YL>\}$\\ 
       {\small Unlabeled images:} $\XU=\{\IU\}$\\
       {\small Object detection architecture:} $\CNN$\\
       {\small $\CNN$ Training hyper-par.:} $\HPOD$\\
       {\small Co-tr. hyp.-p.:} $\HPCoT=\{T,N,n,m,\HPStop[,\HPSeq]\}$\\}
\Output{{\small New labeled images:} $\XP=\{<\IP,\YP>\}$}
$<\XP_1,\XP_2,k>\,\,\leftarrow\,\,<\emptyset,\emptyset,0>$\;
$<\XL_1,\XL_2>\,\,\leftarrow\,\,<\XL,(\XL)^\Lsh>$\;
$\OD_1 \leftarrow \TD{\CNN}{\HPOD}{\XL_1}{\XP_1}$\;
$\OD_2 \leftarrow \TD{\CNN}{\HPOD}{\XL_2}{\XP_2}$\;
$\XP_{1,new} \leftarrow \RD{\OD_1}{\XU}{T}$\;
$\XP_{2,new}     \leftarrow \RD{\OD_2}{\XU}{T}$\;
\Repeat{$\STOP{\HPStop}{\XP_{old}}{\XP_{1,new}}{k\mbox{++}}$}{
$\XP_{old} \leftarrow \XP_{1,new}$\;
$\XP_{1,\uparrow} \leftarrow \Select{\uparrow}{m}{\Rand{\XP_{1,new}}{N}{[,\HPSeq,k]}}$\;
$\XP_{2,\uparrow} \leftarrow \Select{\uparrow}{m}{\Rand{\XP_{2,new}}{N}{[,\HPSeq,k]}}$\;
$\XP_{1,\downarrow} \leftarrow \Select{\downarrow}{n}{\RD{\OD_1}{\XP_{2,\uparrow}}{T}}$\;
$\XP_{2,\downarrow} \leftarrow \Select{\downarrow}{n}{\RD{\OD_2}{\XP_{1,\uparrow}}{T}}$\;
$\XP_1 \leftarrow \Fuse{\XP_1}{\XP_{1,\downarrow}}$\;
$\XP_2 \leftarrow \Fuse{\XP_2}{\XP_{2,\downarrow}}$\;
$\OD_1 \leftarrow \TD{\CNN}{\HPOD}{\XL_1}{\XP_1}$\;
$\OD_2 \leftarrow \TD{\CNN}{\HPOD}{\XL_2}{\XP_2}$\;
$\XP_{1,new} \leftarrow \RD{\OD_1}{\XU}{T}$\;
$\XP_{2,new} \leftarrow \RD{\OD_2}{\XU}{T}$\;
}
$\XP \leftarrow \XP_{1,new}$\;
\KwRet{$\XP$}\;
\caption{Self-labeling by co-training (see main text).}
\label{alg:co-training}
\end{algorithm}

\subsection{Self-labeling functional components}
\label{ssec:s-l}

\paragraph{Input and output parameters}
Since we follow a SSL setting \cite{Engelen:2020}, we assume access to a set of images ({\eg} acquired on-board a car) with each object of interest ({\eg} vehicles and pedestrians) labeled by a BB and a class label, as well as to a set of unlabeled images. The former is denoted as $\XL=\{<\IL,\YL>\}$, where $\IL$ is a particular image of the labeled set, being $\YL$ its corresponding labeling information; {\ie}, for each object of interest in $\IL$, $\YL$ includes its BB and its class label. The unlabeled set is $\XU=\{\IU\}$, where $\IU$ is a particular unlabeled image. We assume also a given object detection architecture to perform self-labeling, denoted as $\CNN$, with corresponding training hyper-parameters denoted as $\HPOD$. After self-labeling, we expect to obtain a new set of automatically labeled images, which we denote as $\XP=\{<\IP,\YP>\}$, where $\IP\in\XU$ and has $\YP$ as so-called pseudo-labels, which in this case consists of a BB and a class label for each detected object in $\IP$. The variables introduced so far are generic in SSL, {\ie} they are not specific of our proposals. We denote as $\HPCoT$ the specific set of hyper-parameters we require for self-labeling. In fact, $\HPCoT$ includes $\{T,N,n,\HPStop,\HPSeq\}$ as common parameters for both self-training and co-training, but the latter requires an additional parameter $m$ that we will explain in the context of co-training. If $K$ is the number of classes to be considered, then $T=\{t_1,\ldots,t_K\}$ is a set of per-class detection thresholds, normally used by object detectors to ensure a minimum per-class confidence to accept potential detections. Since self-labeling must perform object detection on unlabeled images, these thresholds are needed. During a self-labeling cycle, $N$  self-labeled images are randomly selected, from which $n$ are kept to retrain the object detector. Self-training and co-training use different criteria to select such $n$ self-labeled images. 
$\HPStop$ consists of the parameters required to evaluate whether self-labeling should stop or not. Finally, $\HPSeq$ is an optional set of parameters, only required if $\XU$ consists of sequences rather than isolated images. Note that, in this case, we can easily avoid to introduce too similar training samples coming from the same object instances in consecutive frames, which has shown to be useful in AL approaches \cite{Aghdam:2019}.

\paragraph{$\TD{\CNN}{\HPOD}{\XL}{{\XP}}:\,\, \OD$} This function returns an object detector, $\OD$, by training a CNN architecture $\CNN$ ({\eg} Faster R-CNN \cite{Ren:2017}) according to the training hyper-parameters $\HPOD$ ({\eg} optimizer, learning rate, mini-batch size, training iterations, etc.). Note that this is just the standard manner of training $\OD$, but as part of our self-labeling procedures, we control the provided training data. In particular, we use labels (in $\XL$) and pseudo-labels (in $\XP$) indistinctly. However, we only consider background samples based on $\XL$, since, during self-labeling, $\XP$ can contain false negatives which could be erroneously taken as hard-negatives when training $\OD$. Despite this control over the training data, note that we neither require custom modifications of the loss function used to train $\OD$, nor architectural modifications of $\CNN$.

\paragraph{$\RD{\OD}{\XU}{T}:\,\, \XP$} This function returns a set of self-labeled images, $\XP$, obtained by running the object detector $\OD$ on the unlabeled set of images $\XU$; in other words, the pseudo-labels correspond to the object detections. Each detection $D$ consists of a BB $B$ and a class label $c$; moreover, being a detection implies that the confidence $\OD(D)$ fulfils the condition $\OD(D)\geq t_c, t_c\in T$. 

\paragraph{$\Rand{\XP}{N}{[,\HPSeq,k]}:\,\, \XP_{s}$} This function returns a set $\XP_{s}$ of $N$ self-labeled images randomly chosen from $\XP$.  $\HPSeq$ and $k$ are optional parameters. If they are provided, it means that $\XP$ consists of sequences and we want to ensure not to return self-labeled consecutive frames, since they may contain very similar samples coming from the same object instances and we want to favor variability in the samples. Moreover, in this way we can minimize the inclusion of spurious false positives that may persist for several frames. In this case, $\HPSeq=\{{\Delta t}_1,{\Delta t}_2\}$; where ${\Delta t}_1$ is the minimum distance between frames with pseudo-labels generated in the current cycle, $k$, and ${\Delta t}_2$ is the minimum distance between frames with pseudo-labels generated in cycle $k$ and frames with pseudo-labels generated in previous cycles ($<k$). ${\Delta t}_1$ condition is applied first, then ${\Delta t}_2$ one. The final random selection is performed on the frames remaining after applying these distance conditions.

\paragraph{$\Select{s}{n}{\XP}:\,\, \XP_{s}$} This function returns a sub-set $\XP_{s}\subset\XP$ of $m$ self-labeled images, selected according to a criterion $s$. If $s=\uparrow$, the top $n$ most confident images are selected; while for $s=\downarrow$, the $n$ least confident images are selected. We resume the confidence of an image from the confidences of its detections. For the sake of simplicity, since $\XP=\{<\IP,\YP>\}$, we can assume that $\YP$ not only contains the BB and class label for each detected object in $\IP$, but also the corresponding detection confidence. Then, the confidence assigned to $\IP$ is the average of the confidences in $\YP$. 

\paragraph{$\Fuse{\XP_{old}}{\XP_{new}}:\,\, \XP$} The goal of this function is to avoid redundant detections between the self-labeled sets $\XP_{old}$ and $\XP_{new}$ and preserving different ones,  producing a new set of self-labeled images, $\XP$. Thus, on the one hand, $\XP$ will contain all the self-labeled images in $\XP_{old}\cup\XP_{new} - \XP_{old}\cap\XP_{new}$; on the other hand, for those images in $\XP_{old}\cap\XP_{new}$, only the detections in $\XP_{new}$ are kept, so the corresponding information is added to $\XP$. 

\paragraph{$\STOP{\HPStop}{\XP_{old}}{\XP_{new}}{k}:\,\, {\small \mbox{Boolean}}$} This function decides whether to stop self-labeling or not. The decision relies on two conditions. The first one is if a minimum number of self-labeling cycles, $K_{min}$, have been performed, where current number is $k$. The second condition relies on the similarity of $\XP_{old}$ and $\XP_{new}$, computed as the mAP (mean average precision) \cite{Geiger:2012} by using $\XP_{old}$ in the role of ground truth and $\XP_{new}$ in the role of results to be evaluated. We compute the absolute value difference between the mAP at current cycle and previous one, keeping track of these differences in a sliding window fashion. If these differences are below a threshold, $T_{\Delta_{mAP}}$, for more than a given number of consecutive cycles, $\Delta K$, self-labeling will stop. Therefore, $\HPStop=\{K_{min},T_{\Delta_{mAP}},\Delta K\}$. 

\subsection{Self-training}
\label{ssec:s-t}

Algorithm \ref{alg:self-training} describes self-training based on the functional components introduced in Section \ref{ssec:s-l}. In the following, we highlight several points of this algorithm. 

As $\Select{}{}{}$ is run in the loop, it implies that in each cycle the $n$ most confident self-labeled images are kept for retraining $\OD$. Moreover, potentially redundant object detections arising from different cycles are resumed by running the $\Fuse{}{}{}$ function. As is called $\Fuse{}{}{}$ in the loop, when a previous detection and a new one overlap enough, the new one is kept since it is based on the last trained version of $\OD$, which is expected to be more accurate than previous ones. When calling $\Select{}{}{}$, not all the available self-labeled images are considered, but just a random set of them of size $N$; which are chosen taking into account the selection conditions introduced in case of working with sequences. This prevents the same highly accurate detections to be systematically selected across cycles, which would prevent the injection of variability when retraining $\OD$. The random selection over the entire $\XU$ was also proposed in the original co-training algorithm \cite{Blum:1998}. As we have mentioned before, when running $\TD{}{}{}{}$, background samples are only collected from $\XL$ to avoid introducing false positives. Moreover, 
we set $\HPOD$ to ensure that all the training samples are visited at least once (mini-batch size $\times$ number of iterations $\geq$ number of training images). Therefore, we can think of $\XL\cup\XP$ as a mixing of data, where $\XL$ acts as a regularising factor during training, which aims to prevent $\OD$ to become an irrecoverably bad detector. Thinking in a virtual-to-real UDA setting, we note that from traditional ML algorithms to modern deep CNNs, mixing virtual and real data has been shown to be systematically beneficial across different computer vision tasks \cite{Vazquez:2014,Ros:2016,DeSouza:2019,Jaipuria:2020}. Finally, we can see that to run the stopping criterion we rely on the full self-labeled data available at the beginning and end of each cycle, which results from applying the last available version of $\OD$ to the full $\XU$ set.

\subsection{Co-training}
\label{ssec:c-t}

Algorithm \ref{alg:co-training} describes our co-training proposal based on the functional components introduced in Section \ref{ssec:s-l}. In the following, we highlight several points of this algorithm.

Our co-training strategy tries to make $\OD_1$ and $\OD_2$ different by training them on different data. Note how each $\OD_i$ is trained on $\XL_i$ and $\XP_i$, $i\in\{1,2\}$, at each cycle. The $\XL_i$'s do not change across cycles and we have $\XL_1=\XL$ and $\XL_2=(\XL)^\Lsh$, thus, one of the labeled sets is a horizontal mirroring of the other. Moreover, the $\XP_i$'s are expected to be different from each other and changing from cycle to cycle. On the other hand, even technically not using the same exact data to train $\OD_1$ and $\OD_2$, there will be a cycle when they converge and are not able to improve each other. This is what we check to stop, {\ie} as for self-training, but focusing on the performance of $\OD_1$ since it is the detector using $\XL$ (the original labeled dataset).

Another essential question is how each $\XP_i$ is obtained. As we have mentioned before, we follow the idea of disagreement. Thus, from the random set of images self-labeled by $\OD_i$, {\ie} $\XP_{i,new}$, $i\in\{1,2\}$, we set to $\XP_{i,\uparrow}$ the $m$ with overall higher confident pseudo-labels; then, the images in $\XP_{i,\uparrow}$ are processed by $\OD_j$, $j\in\{1,2\}, i \neq j$, and the $n$ with the overall lower confident pseudo-labels are set to $\XP_{j,\downarrow}$. Finally, each $\XP_{i,\downarrow}$ is fused with the $\XP_i$ accumulated from previous cycles to update $\XP_i$. Therefore, in each cycle, each $\OD_i$ is retrained with the images containing the less confident pseudo-labels for current $\OD_i$, selected among the images containing the most confident pseudo-labels for current $\OD_j$. 
In this way, the detectors trust each other, and use the samples that are more difficult for them for improving. 

Note also that the most costly processing in self-training and co-training cycles is the $\TD{}{}{}{}$ procedure. It is called once per cycle for self-training, and twice for co-training. However, in the latter case the two executions can run totally in parallel, therefore, with the proper hardware resources, self-training and co-training cycles can be performed in a very similar time.

\subsection{Self-labeling for UDA}
\label{ssec:s-l}

Once introduced our self-training and co-training algorithms, we see how using them for UDA is just a matter of providing the proper input parameters; {\ie}, $\XL=\XStrL$, $\XU=\XT$, being $\XStrL$ the source-domain labeled dataset, and $\XT$ the target-domain unlabeled one. We will draw the former from a virtual world, and the latter from the real world.

%% file: sections/experiments.tex
\section{Experiments}
\label{sec:experiments}

\begin{table}
\centering
\caption{Datasets ($\dataset$): train ($\Xtrain$) and test ($\Xtest$) information, $\dataset=\Xtrain\cup\Xtest, \Xtrain\cap\Xtest=\emptyset$. We show the number of images/frames (sequences), vehicle BBs, pedestrian BBs, and whether the datasets consists of video sequences or not.}\label{tab:datasets}
\resizebox{1.0\linewidth}{!}{
\begin{tabular}{@{}lccccccc@{}}
\toprule
                      & \multicolumn{3}{c}{$\Xtrain$}    & \multicolumn{3}{c}{$\Xtest$}    &  \\
Dataset ($\dataset$)  & Images (seq.) & Vehicles & Pedestrians  & Images (seq.) & Vehicles & Pedestrians &  Seq.? \\
\toprule
VIRTUAL ($\Vds$)      & 19,791 & 43,326  & 44,863        &        &          &             &  No \\
\midrule
KITTI ($\Kds$)        & 3,682 & 14,941  & 3,154        
& 3,799 & 18,194   & 1,333      &  No \\
\midrule
WAYMO ($\Wds$)        & 9,873 (50) & 64,446  & 9,918
& 4,161 (21) & 24,600   & 3,068      &  Yes \\
\bottomrule
\end{tabular}
}
\end{table}

\subsection{Experimental setup}
\label{ssec:experimentalsetup}

\begin{figure}
\centering
\includegraphics[width=\columnwidth]{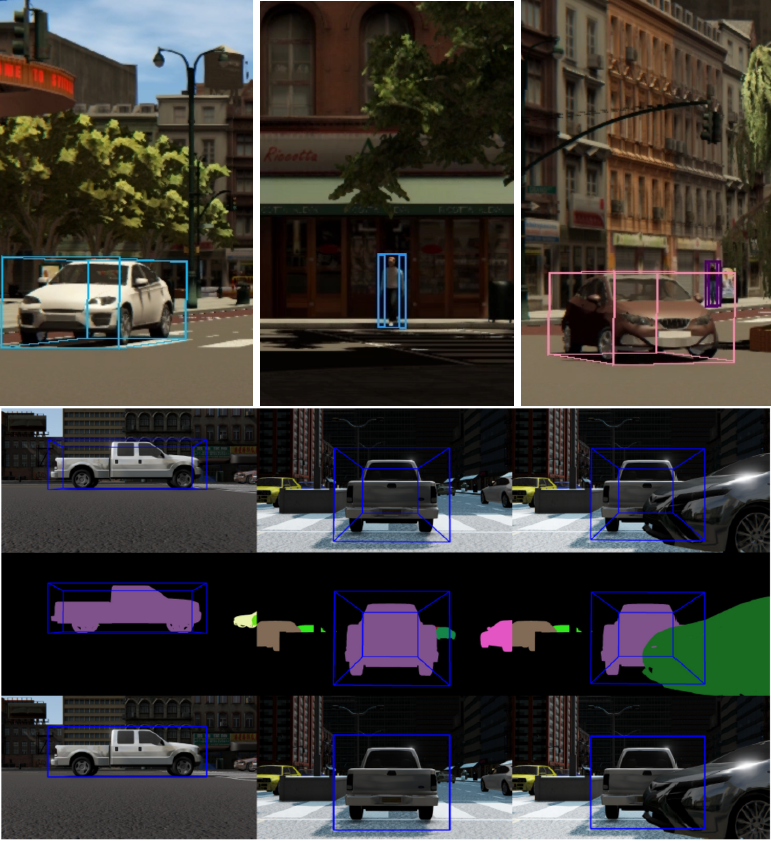}
\caption{Top images: virtual-world patches showing 3D BBs framing vehicles and pedestrians. Bottom image: projecting 3D BBs as 2D BBs for different views of a pickup, with instance segmentation as visual reference.}
\label{fig:virtual-3DBBsIn2DImages}
\end{figure}

As real-world datasets we use KITTI \cite{Geiger:2012} and Waymo \cite{Sun:2020}, in the following denoted as $\Kds, \Wds$, respectively. The former is a well-established standard born in the academia, the latter is a new contribution from the industry. To work with $\Kds$, we follow the splits introduced in \cite{Xiang:2015}, which 
are based on isolated images, and 
guarantee that training and testing images are not highly correlated; {\ie}, training and testing images are not just different frames sampled from the same sequences. On the other hand, $\Wds$ dataset contains video sequences. Following the advice for its use, we have randomly selected part of the sequences for training and the rest for testing. Taking $\Kds$ as reference, we focus on daytime and non-adverse weather conditions in all the cases. In addition, we have accommodated $\Wds$ images to the resolution of $\Kds$ ones, {\ie} $1240\times375$ pixels, by just cropping away their upper part (which mainly shows sky) and keeping the vertically-centered $1240$ columns. Following KITTI benchmark moderate settings, we set the minimum BB height to detect objects to $25$ pixels and, analogously, $50$ pixels for $\Wds$. Moreover, since $\Wds$ contains 3D BBs, as for the virtual-world objects (Figure \ref{fig:virtual-3DBBsIn2DImages}), we obtain 2D BBs from them. In other words, the resulting 2D BBs account for occluded object areas, which is not the case for the 2D BBs directly available with $\Wds$.


\begin{figure*}
\setlength{\tabcolsep}{1pt}
\centering
\begin{tabular}{ccc}
\centeredcell{\includegraphics[width=0.33\textwidth]{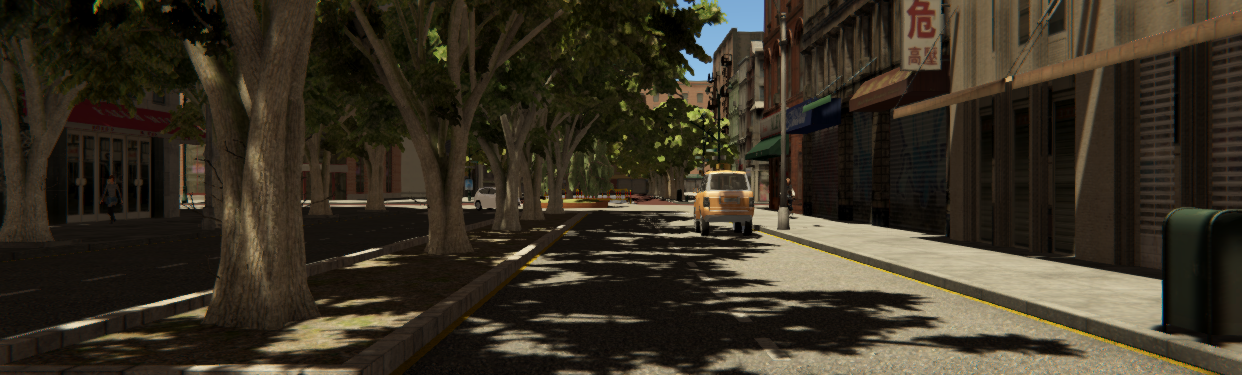}} &
\centeredcell{\includegraphics[width=0.33\textwidth]{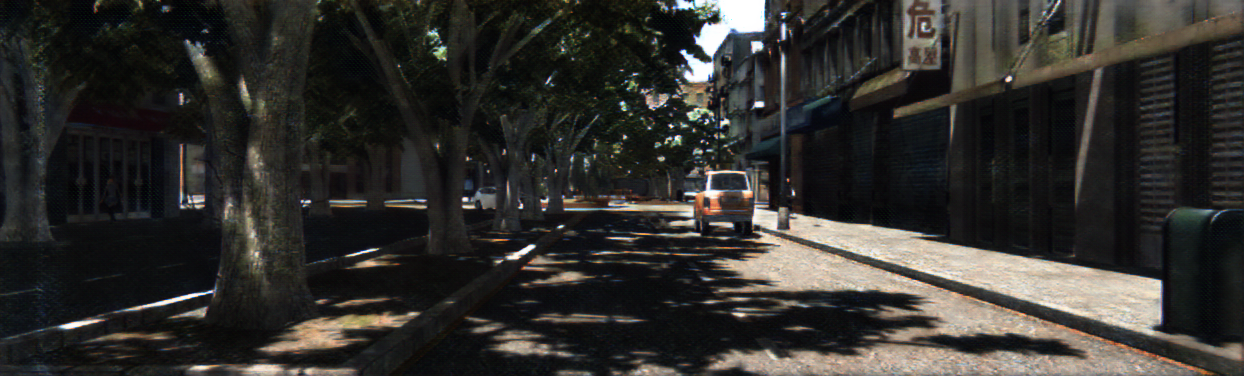}} &
\centeredcell{\includegraphics[width=0.33\textwidth]{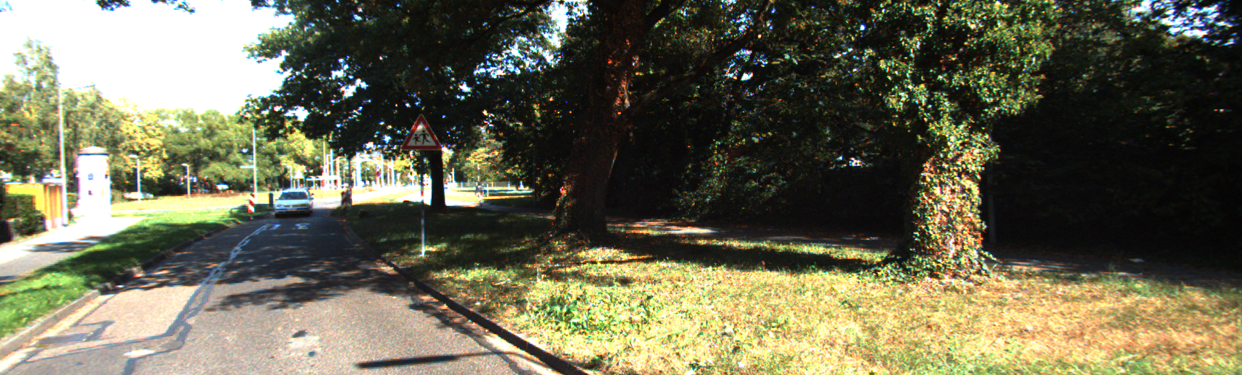}} \\
\centeredcell{\includegraphics[trim=0 0 100 0, clip,width=0.33\textwidth]{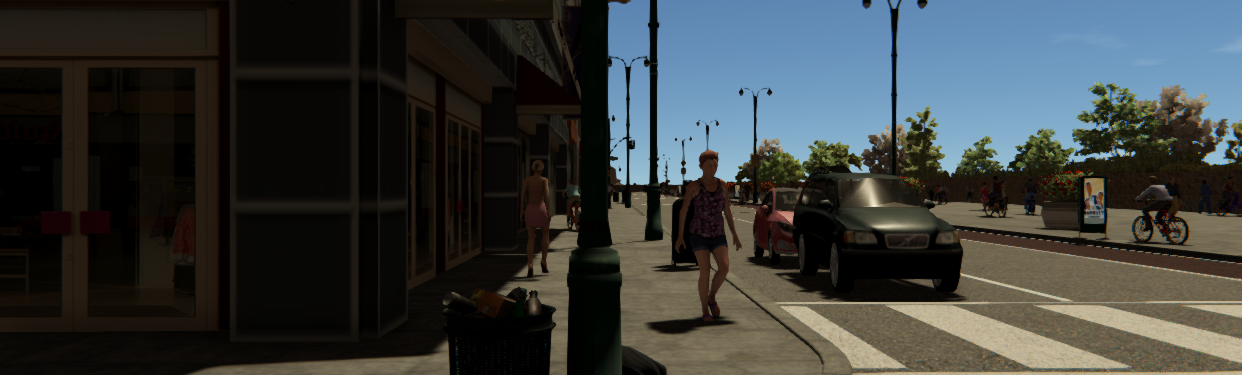}} &
\centeredcell{\includegraphics[trim=0 0 100 0, clip,width=0.33\textwidth]{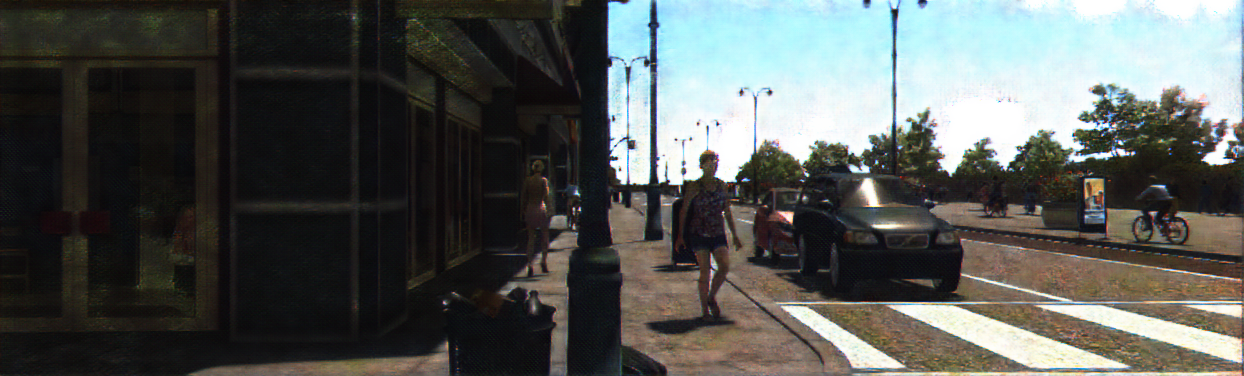}} &
\centeredcell{\includegraphics[width=0.33\textwidth]{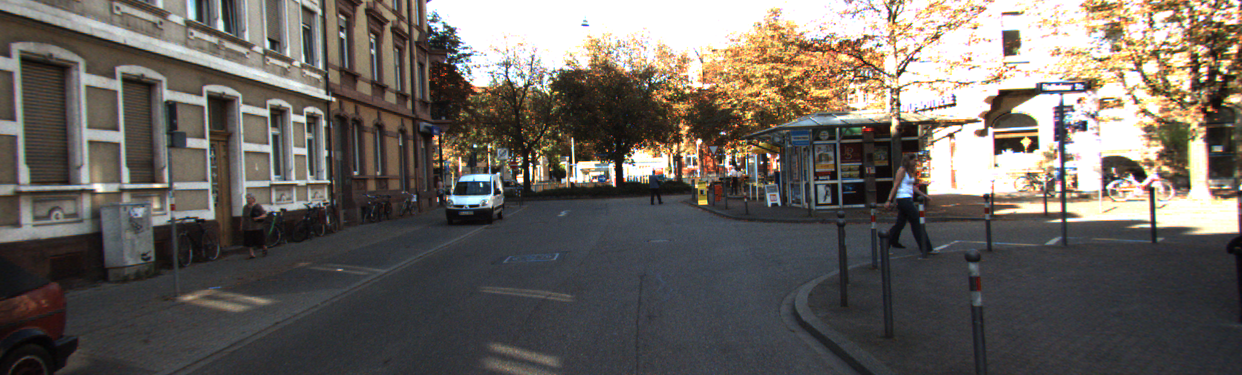}} \\
\centeredcell{\includegraphics[width=0.33\textwidth]{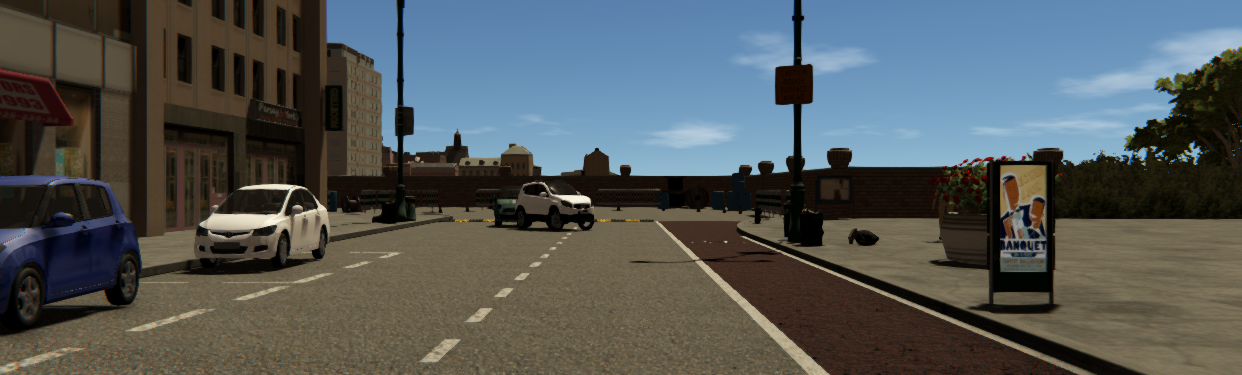}} &
\centeredcell{\includegraphics[width=0.33\textwidth]{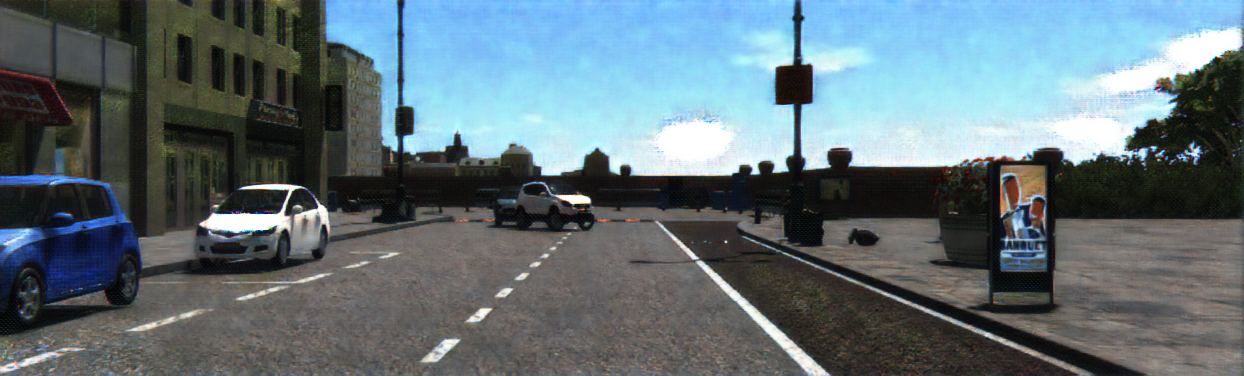}} &
\centeredcell{\includegraphics[width=0.33\textwidth]{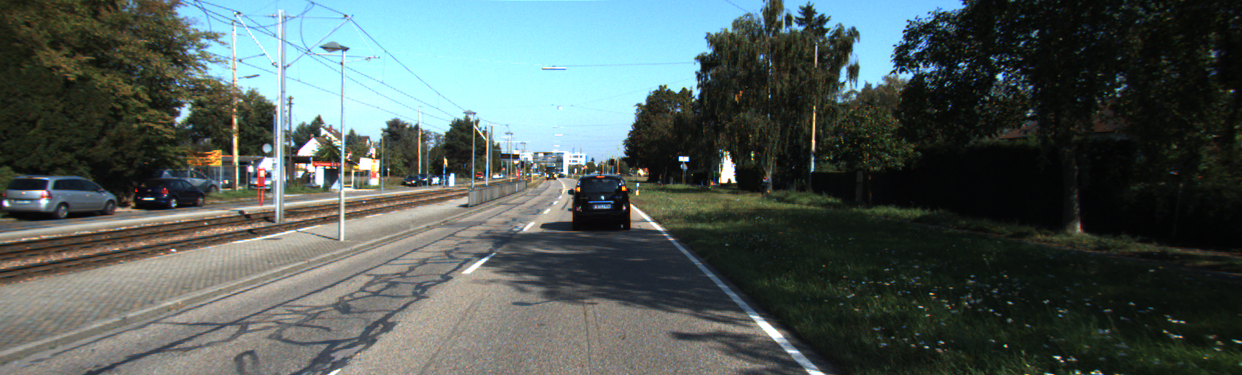}} \\
\centeredcell{\includegraphics[width=0.33\textwidth]{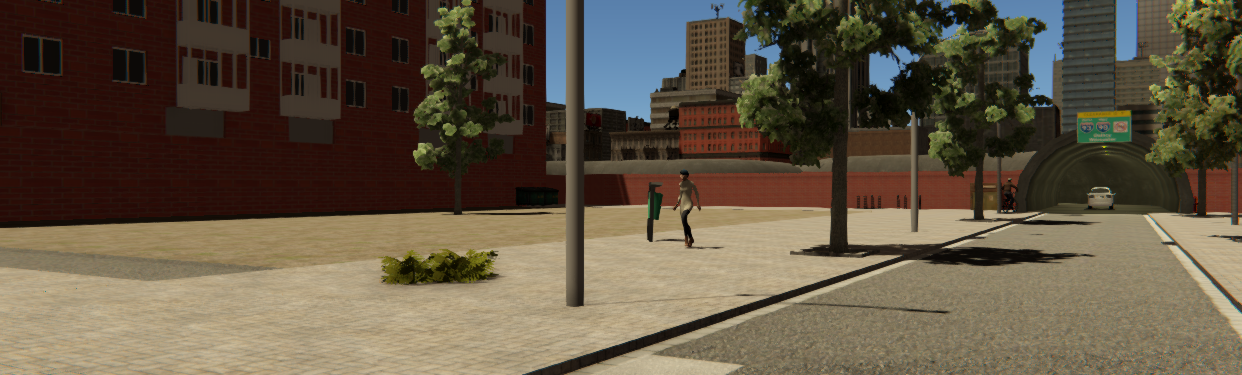}} &
\centeredcell{\includegraphics[width=0.33\textwidth]{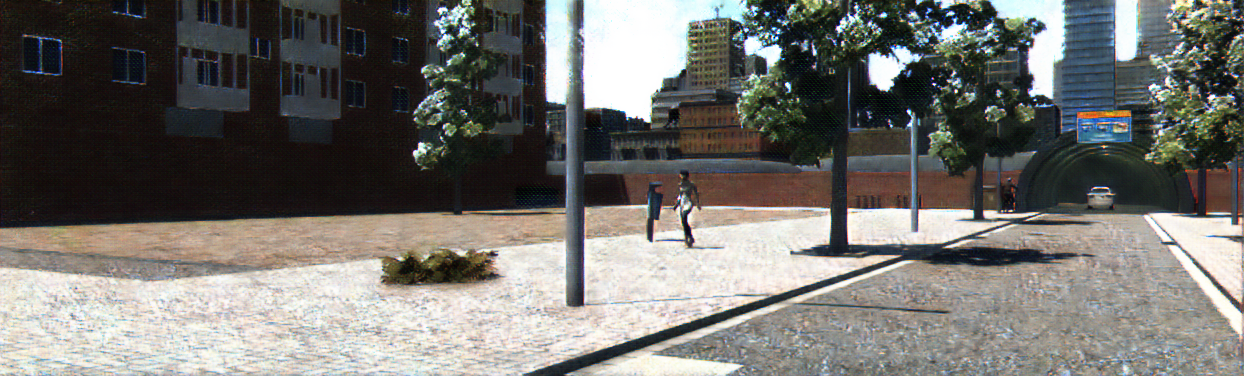}} &
\centeredcell{\includegraphics[width=0.33\textwidth]{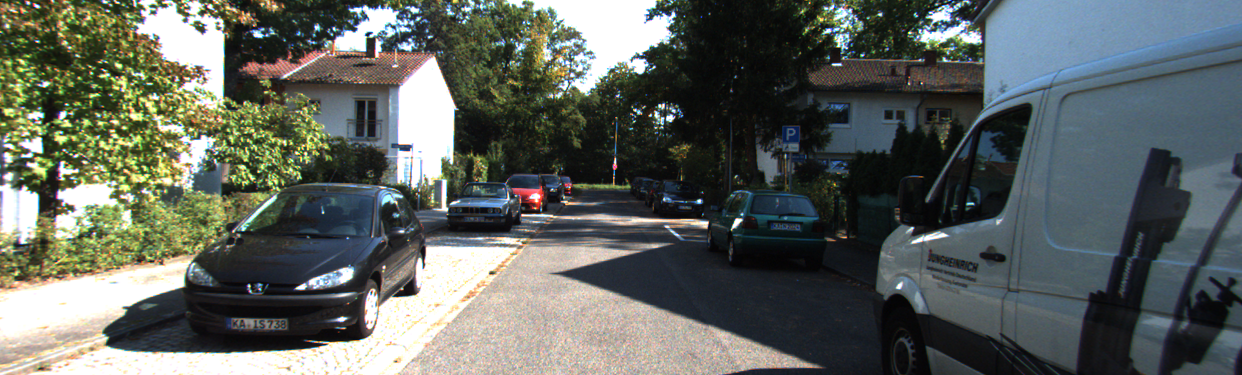}} \\
\centeredcell{\includegraphics[width=0.33\textwidth]{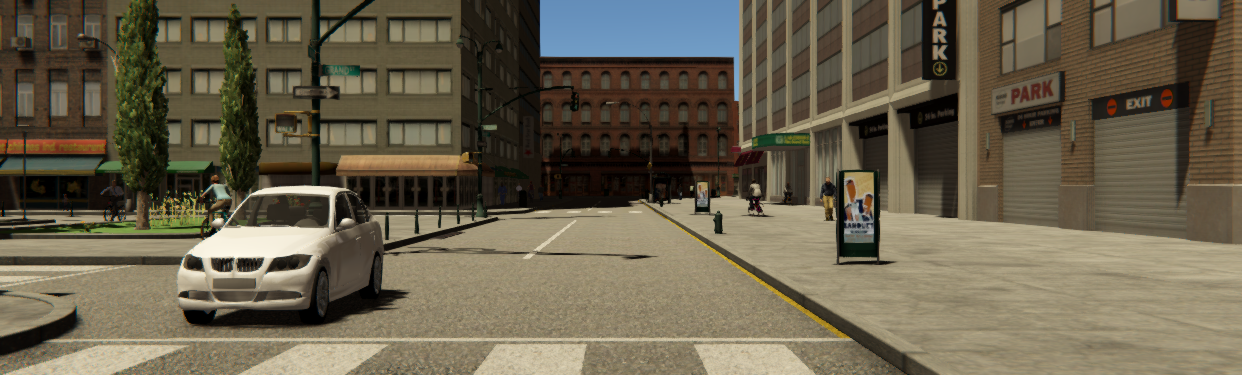}} &
\centeredcell{\includegraphics[width=0.33\textwidth]{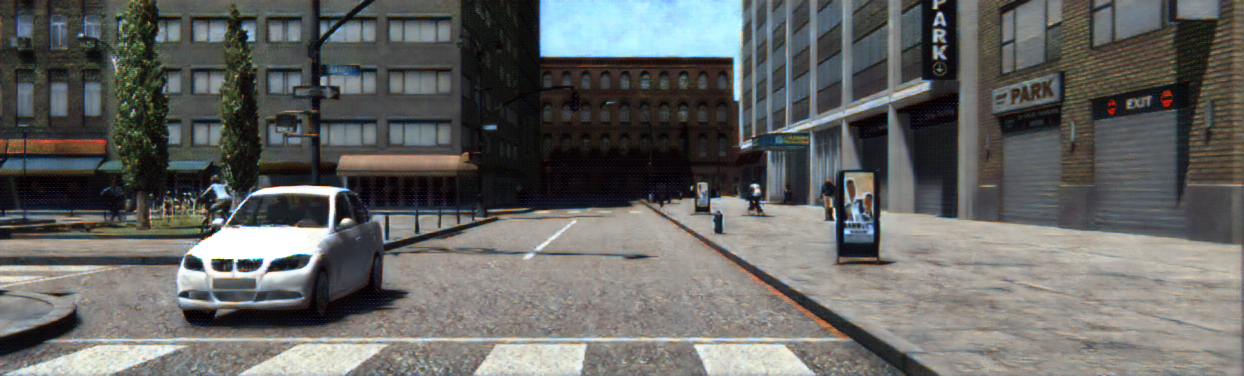}} &
\centeredcell{\includegraphics[width=0.33\textwidth]{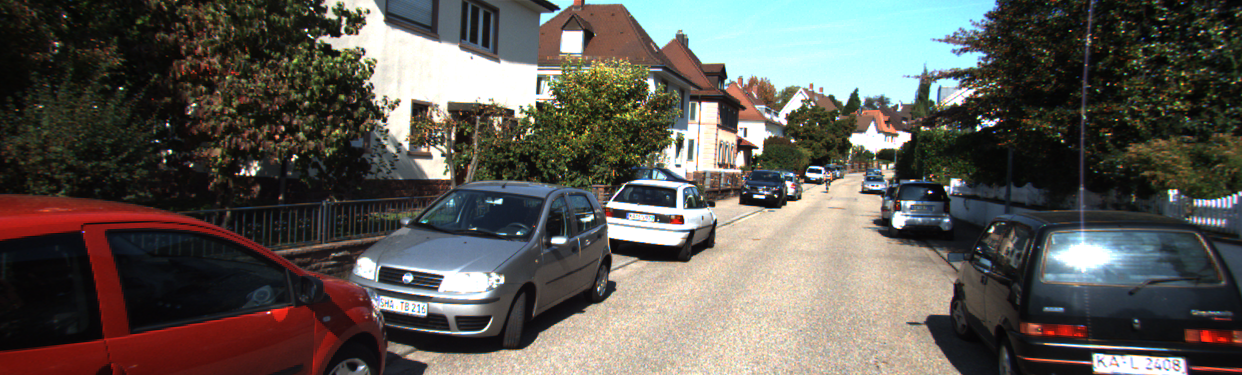}} 
\end{tabular}
\caption{From left to right: images from $\Vds$, corresponding images in $\VGKds$ ({\ie} processed by the $\Vds\rightarrow\Kds$ GAN), and images from $\Kds$. Last column is just a visual reference since, obviously, there is no a one-to-one correspondence between $\Vds$ and $\Kds$.} 
\label{fig:translation_kitti}
\end{figure*}

\begin{figure*}
\setlength{\tabcolsep}{1pt}
\centering
\begin{tabular}{ccc}
\centeredcell{\includegraphics[width=0.33\textwidth]{images/translation/synthia/000003.png}} &
\centeredcell{\includegraphics[width=0.33\textwidth]{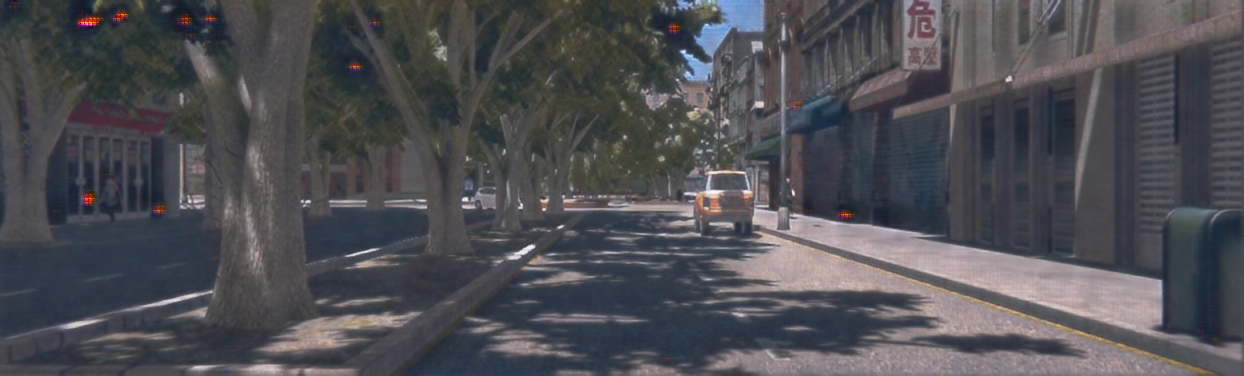}} &
\centeredcell{\includegraphics[width=0.33\textwidth]{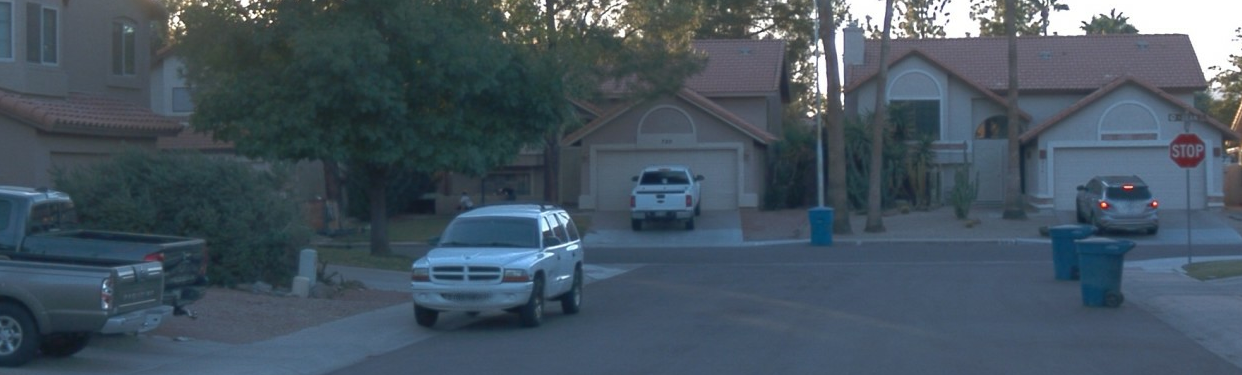}} \\
\centeredcell{\includegraphics[trim=0 0 100 0, clip,width=0.33\textwidth]{images/translation/synthia/000188.png}} &
\centeredcell{\includegraphics[trim=0 0 100 0, clip,width=0.33\textwidth]{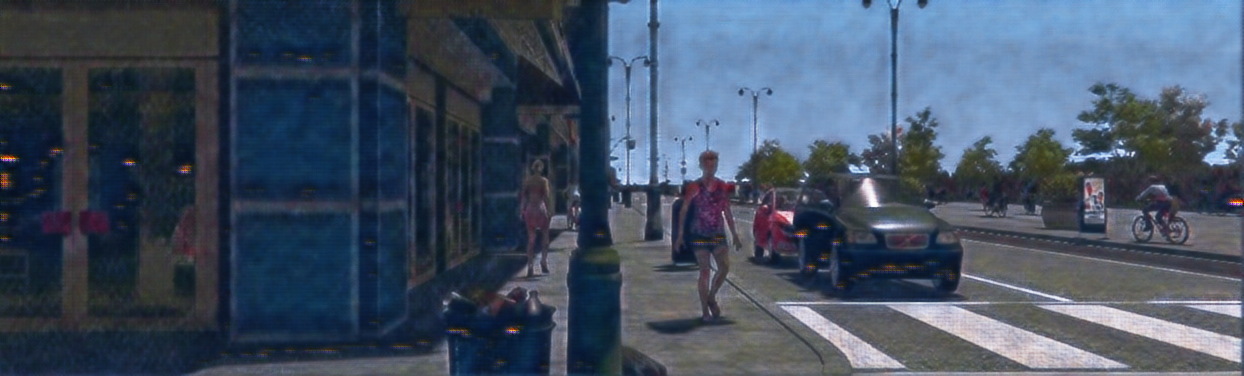}} &
\centeredcell{\includegraphics[width=0.33\textwidth]{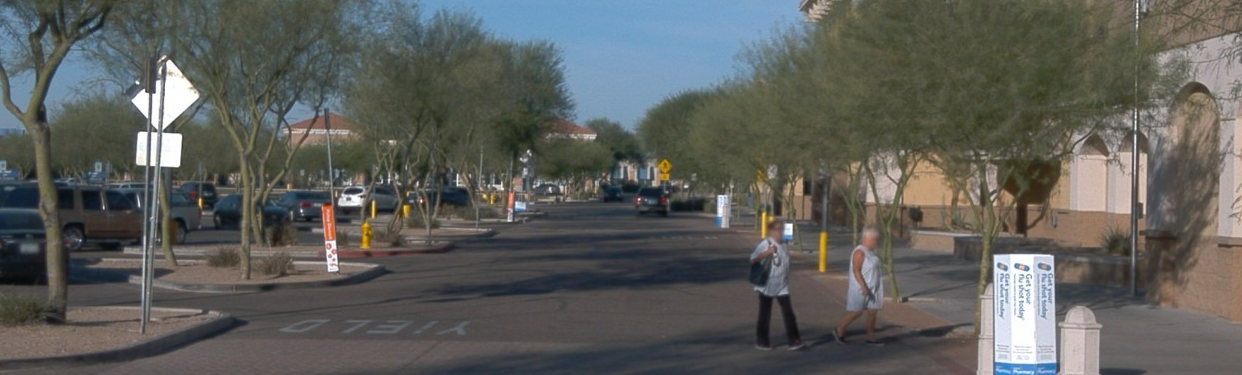}} \\
\centeredcell{\includegraphics[width=0.33\textwidth]{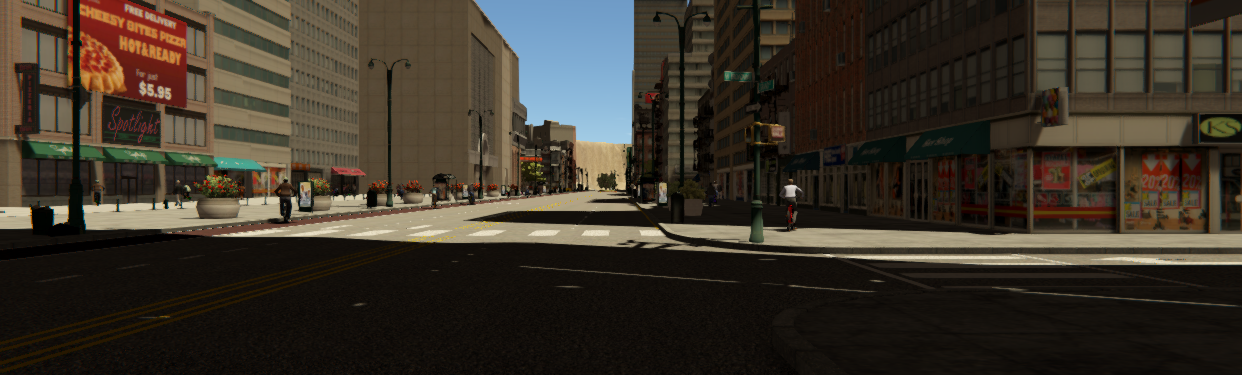}} &
\centeredcell{\includegraphics[width=0.33\textwidth]{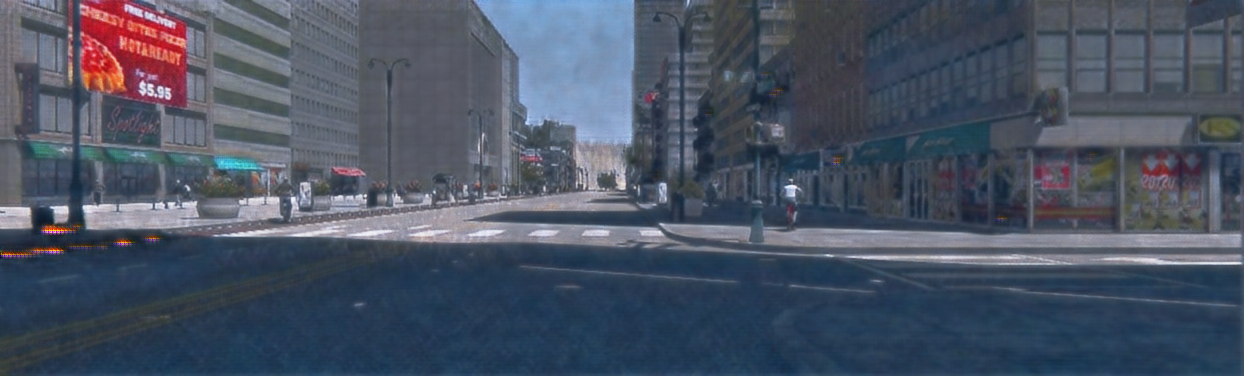}} &
\centeredcell{\includegraphics[width=0.33\textwidth]{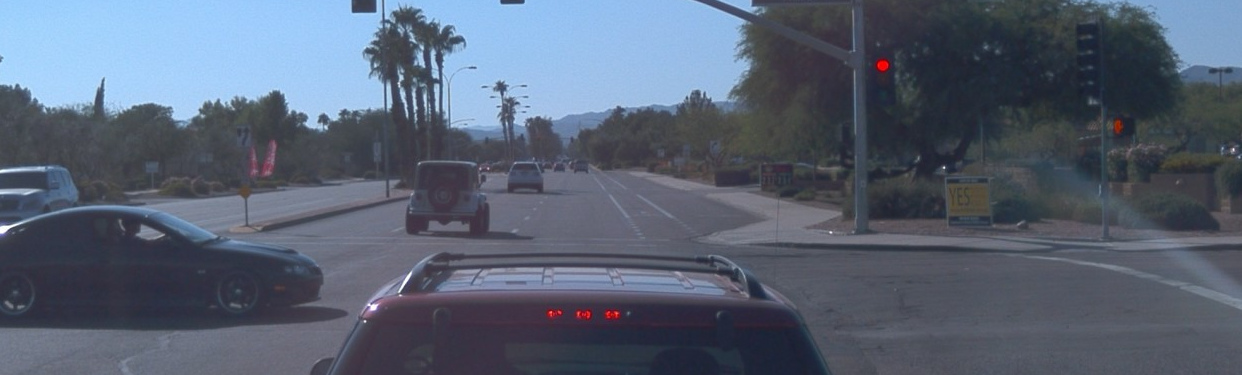}} \\
\centeredcell{\includegraphics[width=0.33\textwidth]{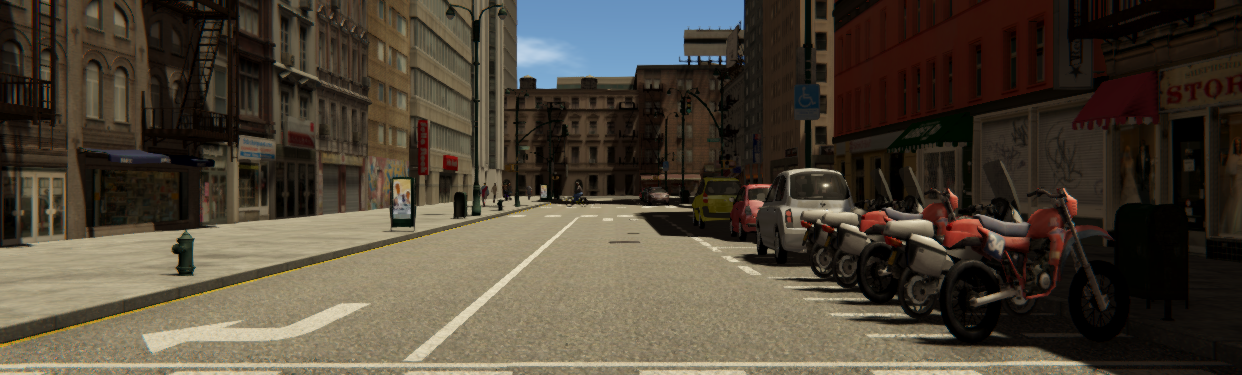}} &
\centeredcell{\includegraphics[width=0.33\textwidth]{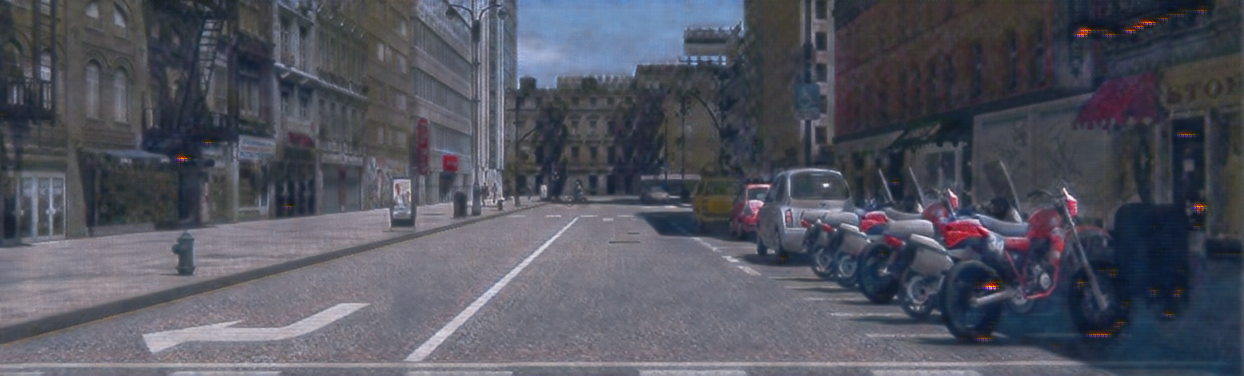}} &
\centeredcell{\includegraphics[width=0.33\textwidth]{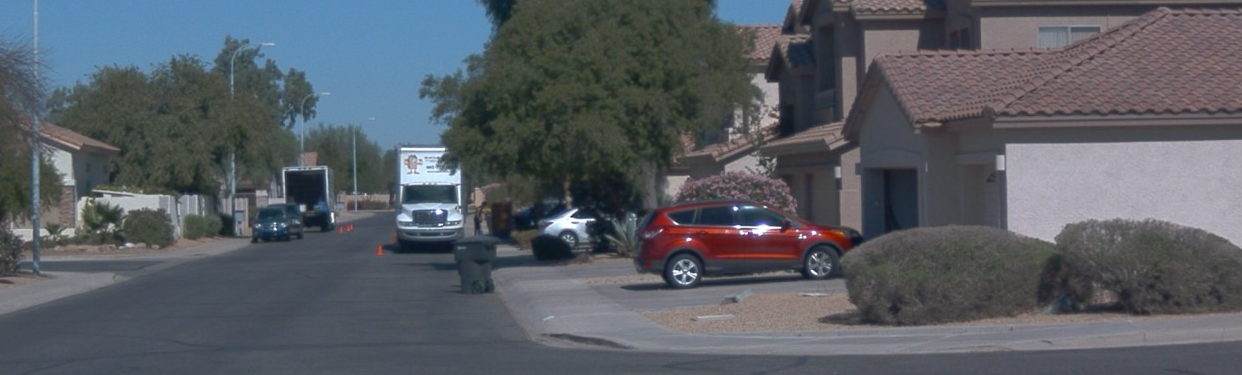}} \\
\centeredcell{\includegraphics[width=0.33\textwidth]{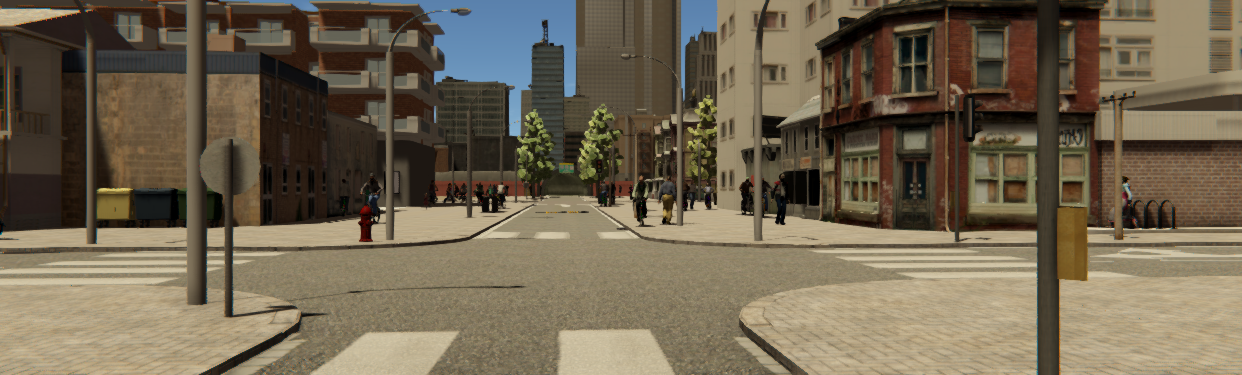}} &
\centeredcell{\includegraphics[width=0.33\textwidth]{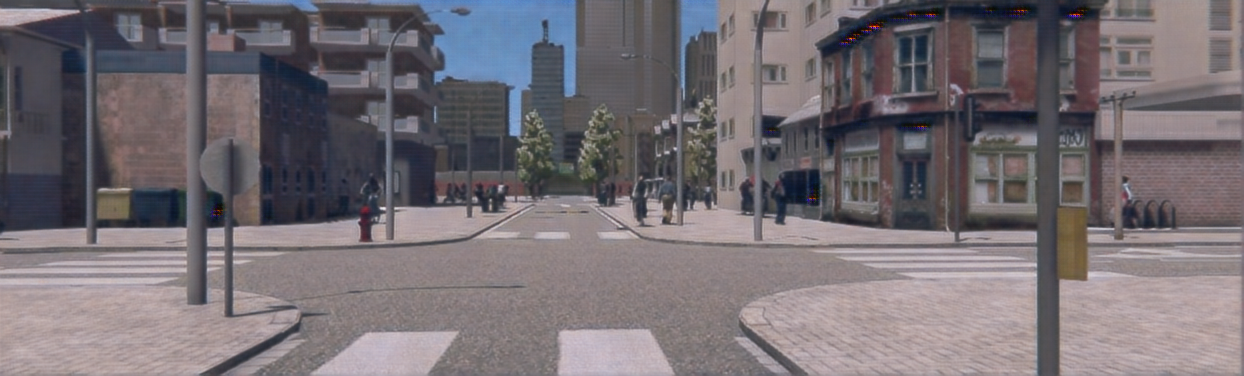}} &
\centeredcell{\includegraphics[width=0.33\textwidth]{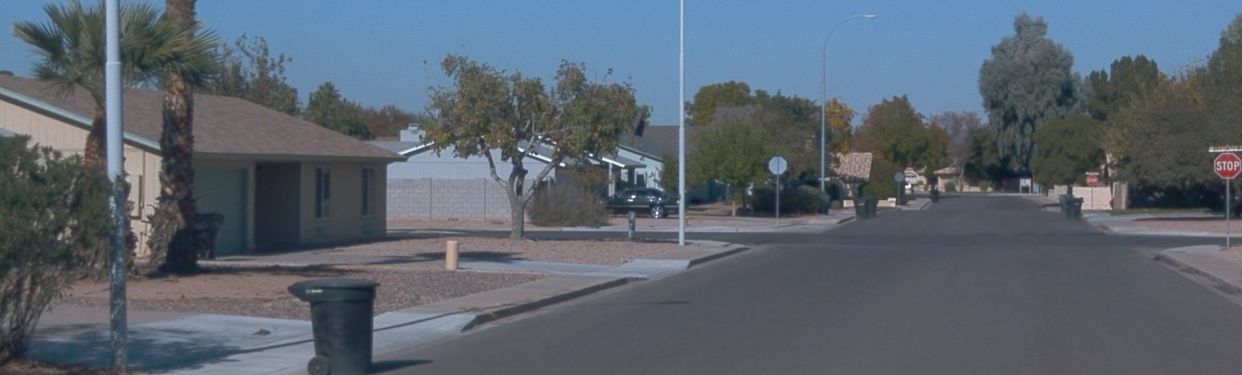}}
\end{tabular}
\caption{From left to right: images from $\Vds$, corresponding images in $\VGWds$ ({\ie} processed by the $\Vds\rightarrow\Wds$ GAN), and images from $\Wds$. For visual comparison, the two top rows of this figure and those in Figure \ref{fig:translation_kitti}, start with the same images in $\Vds$.} 
\label{fig:translation_waymo}
\end{figure*}

In order to perform our experiments, as set of virtual-world images ($\Vds$), we have used a dataset that we generated internally around two years ago as a complement for our former SYNTHIA dataset \cite{Ros:2016}. We did not have the opportunity to release it at that moment, but it will be publicly available to complement this paper. We generated this data following KITTI parameters: same image resolution, daytime and non-adverse weather, and isolated images. As in \cite{Ros:2016}, we did not focus on obtaining photo-realistic images. However, for generating $\Vds$, we included standard visual post-effects such as anti-aliasing, ambient occlusion, depth of field, eye adaptation, blooming and chromatic aberration. 
Later, we will see how the domain shift between $\Vds$ and $\Kds/\Wds$, is similar to the one between $\Kds$ and $\Wds$;
thus, being $\Vds$ a proper virtual-world dataset for our study. Figure \ref{fig:virtual-3DBBsIn2DImages} shows virtual-world images with 3D BBs framing vehicles and pedestrians, illustrating also how the 3D BBs are projected as 2D BBs covering occluded object areas.

For each dataset, Table \ref{tab:datasets} summarizes the number of images/frames and objects (vehicles and pedestrians) used for training and testing our object detectors. 

Since our proposal does not assume a particular object detection architecture, for performing the experiments we have chosen Faster R-CNN \cite{Ren:2017} since it provides a competitive detection accuracy and
is very well known by computer vision practitioners \cite{Liu:2020}. We use the implementation available in the \emph{Detectron} framework \cite{Detectron:2018}, with VGG16 as backbone, always initialized on ImageNet weights since it is a recommended best practice. Each Faster R-CNN training is run for 40,000 iterations using SGD optimizer. The learning rate starts at 0.001 and we have used a decay of 0.1 at iterations 30,000 and 35,000. Each iteration uses a mini-batch of 2 images. In terms of Algorithms \ref{alg:self-training} and \ref{alg:co-training}, these settings correspond to $\HPOD$.

As GAN-based image-to-image translation method we use the CycleGAN implementation in \cite{Zhu:2017}. A GAN training runs for 40 epochs using a weight of 1.0 for the identity mapping loss. In order to be able to use full resolution images, the training of the GAN has been done patch-wise, with patch sizes of $300\times300$ pixels. The rest of parameters have been set as recommended in \cite{Zhu:2017}. We train a $\Vds\rightarrow\Kds$ transforming GAN, $\GANKds$, and another for $\Vds\rightarrow\Wds$, $\GANWds$. Accordingly, $\VGKds=\GANKds(\Vds)$ denotes the set of virtual-world images transformed by $\GANKds$ and, analogously, we have $\VGWds=\GANWds(\Vds)$. We will use the notation $\VGds$ to refer to any of these sets. The ground truth present in $\Vds$ is used as ground truth for $\VGds$.
For qualitative examples of image-to-image transformations, we refer to Figures \ref{fig:translation_kitti} and \ref{fig:translation_waymo}, using $\GANKds$ and $\GANWds$, respectively.

Table \ref{tab:parameters} summarizes the rest of hyper-parameters used to perform our experiments. Note that, regarding our self-labeling approaches, we use the same values for $\Kds$ and $\Wds$, with the only exception that for $\Wds$ we also take into account that it is composed of video sequences. Moreover, we also use the same values for hyper-parameters shared by self-training and co-training. We relied on the meaning of the hyper-parameters to set them with reasonable values. Then, during the experiments of the 5\% with self/co-training (see Table \ref{tab:sslresults}) we did visual inspection of the self-labeled images to ensure the methods were working well. In this process, we noted that, since the confidence thresholds ($T$) were already avoiding too erroneous self-labeled images, for co-training it was better to send all the images self-labeled by the detector $\model_i$ to detector $\model_j$, $i,j\in\{1,2\},i \neq j$, so that $\model_j$ can select the $n$ most difficult for it among them. This is equivalent to set $m=\infty$ in Algorithm \ref{alg:co-training} (so nullifying the parameter).

Algorithms \ref{alg:self-training} and \ref{alg:co-training} return self-labeled images, {\ie} $\XP$. For our experiments, we use $\XP$ together with the input labeled set, $\XL$, to train a final Faster R-CNN object detector out of the self-labeling cycles, but using the same training settings. This detector is the one used for testing. Finally, to measure object detection performance, we follow the KITTI benchmark mean average precision (mAP) metric \cite{Geiger:2012}.


\begin{table}
\centering
\caption{Self-training and co-training hyper-parameters as defined in Algorithms \ref{alg:self-training} and \ref{alg:co-training}. We use the same values for both, as well as to work with KITTI ($\Kds$) and Waymo ($\Wds$) datasets, except for $\HPSeq$ which only applies to $\Wds$. $N$, $n$, $m$, ${\Delta t}_1$, and ${\Delta t}_2$ are set in number-of-images units, $K_{min}$ and ${\Delta K}$ in number-of-cycles, $T_{\Delta_{mAP}}$ runs in [0..100]. We use the same confidence detection threshold for vehicles and pedestrians, which runs in $[0..1]$. $(^\star)$ Only used in co-training, however, for $m=\infty$, it has no effect.}\label{tab:parameters}
\resizebox{1.0\linewidth}{!}{
\begin{tabular}{@{}cccccccccccc@{}}
\toprule
           &       &     &              && \multicolumn{3}{c}{$\HPStop$}        && \multicolumn{2}{c}{$\HPSeq$} \\
$T$ & $N$   & $n$ & $m^{\star}$  && $K_{min}$ & $T_{\Delta_{mAP}}$ & ${\Delta K}$ && ${\Delta t}_1$ & ${\Delta t}_2$        \\
\midrule
\{0.8,0.8\}        & 2,000 & 100 & $\infty$     && 20        & 2.0       & 5            && 5              & 10 \\
\bottomrule
\end{tabular}
}
\end{table}

\subsection{Results}
\label{ssec:results}

We start by assessing the self-labeling algorithms in a pure SSL setting, working only with either the $\Kds$ or $\Wds$ dataset. Table \ref{tab:sslresults} shows the results when using the 100\% of the respective real-world training data, {\ie} $\XL=\Xtrain\in\{\Kdstrain, \Wdstrain\}$, as well as when randomly selecting the 5\% or the 10\% of $\Xtrain$ as $\XL$ set, meaning that these subsets are created once and frozen for all the experiments. Table \ref{tab:self-labeledBBs} shows the number of object instances induced by the random selection in each case. In Table \ref{tab:sslresults}, the 100\% case shows the upper-bound performance, and the 5\% and 10\% act as lower-bounds. 

\begin{table}
\centering
\caption{SSL results for $\Kds$ and $\Wds$. We assess vehicle (V) and pedestrian (P) detection, according to the mAP metric. From $\Xtrain\in\{\Kdstrain, \Wdstrain\}$, we preserve the labeling information for a randomly chosen $p$\% of its images, while it is ignored for the rest. We report results for $p$=100 (all labels are used), $p$=5 and $p$=10. 
Slf-T and Co-T stand for self-training and co-training, resp., which refers to how images
were self-labeled from the respective unlabeled training sets.}\label{tab:sslresults}
\resizebox{1.0\linewidth}{!}{
\begin{tabular}{@{}lcccccc@{}}
\toprule
&\multicolumn{3}{c}{$\Xtest=\Kdstest$} & \multicolumn{3}{c}{$\Xtest=\Wdstest$} \\
Training set & V & P & V\&P & V & P & V\&P  \\
\toprule
100\% Labeled (upper-bound)& 81.02 & 65.40 & 73.21 & 61.71 & 57.74 & 59.73 \\
\midrule
5\% Labeled (lower-bound)  & 59.81                      & 36.49          & 48.15  
                           & 51.69                      & 41.92          & 46.81 \\
5\% Labeled  +      Slf-T  & \underline{68.94}          & 40.99          & 54.97 
                           & \underline{\textbf{54.26}} & 55.38          & 54.82 \\
5\% Labeled  +      Co-T   & \underline{\textbf{68.99}} & \textbf{55.07} & \textbf{62.03} 
                           & \underline{54.00}          & \textbf{56.34} & \textbf{55.17} \\
\midrule
10\% Labeled (lower-bound) & 72.13                      & 49.03          & 60.58 
                           & 49.53                      & 49.83          & 49.68 \\
10\% Labeled +      Slf-T  & \textbf{76.32}             & 56.05          & \textbf{66.19} 
                           & 54.88                      & 57.40          & 56.14 \\
10\% Labeled +      Co-T   & 73.08                      & \textbf{58.53} & 65.81 
                           & \textbf{56.15}             & \textbf{60.20} & \textbf{58.18} \\
\bottomrule
\end{tabular}
}
\end{table}

\begin{table}
\centering
\caption{UDA results for $\Vds \rightarrow \{\Kds,\Wds\}$, {\ie} virtual to real.
ASource (\emph{adapted source}) refers to $\VGds\in\{\VGKds,\VGWds\}$. 
$\XLtrain$ refers to the fully labeled target-domain training set. $\XPtrain$ consists of the same images as $\XLtrain$, but self-labeled by either self-taining (Slf-T) or co-training (Co-T). Just as reference, we also show the domain shift between $\Kds$ and $\Wds$. According to these results, as upper-bound for $\Kds$ we take the detector based on $\XLtrain \& \VGds$, while for $\Wds$ it is the detector based on $\XLtrain$. We refer to the main text for more details.}\label{tab:Synth2Real}
\resizebox{1.0\linewidth}{!}{
\begin{tabular}{@{}lcccccc@{}}
\toprule
&\multicolumn{3}{c}{$\Xtest=\Kdstest$} & \multicolumn{3}{c}{$\Xtest=\Wdstest$} \\
Training set & V & P & V\&P & V & P & V\&P  \\
\toprule
Source ($\Kds$)                        &   -                        &   -            &   -                      & 37.79                      & 49.80                    & 43.80 \\
Source ($\Wds$)                        & 45.57                      & 44.11          & 44.84                    &   -                        &   -                      &   -   \\
\midrule
Source ($\Vds$) (lower-bound)          & 62.27                      & 61.28          & 64.28                    & 38.88                      & 53.37                    & 46.13 \\
ASource ($\VGds$)                      & 75.52                      & 61.94          & 68.73                    & 52.42                      & 53.08                    & 52.75 \\
Target ($\XLtrain$)                    & 81.02                      & 65.40          & 73.21                    & $^\dagger$61.71            & $^\dagger$\textbf{57.74} & $^\dagger$\textbf{59.73} \\
Target + Source ($\XLtrain \& \Vds$)   & 81.68                      & \textbf{68.07} & 74.88                    & 60.65                      & 54.23                    & 57.44 \\
Target + ASource ($\XLtrain \& \VGds$) & $^\dagger$\textbf{84.08}   & $^\dagger$66.00& $^\dagger$\textbf{75.04} & \textbf{62.02}             & 51.55                    & 56.79 \\
\midrule
Slf-T + Source ($\XPtrain \& \Vds$)    & 70.25                      & 67.59          & 68.92                    & \underline{48.50}          & 44.63                    & 46.57 \\
Co-T + Source  ($\XPtrain \& \Vds$)    & \textbf{73.53}             & \textbf{69.50} & \textbf{71.52}           & \underline{\textbf{48.56}} & \textbf{56.33}           & \textbf{52.45} \\
\midrule
Slf-T + ASource ($\XPtrain \& \VGds$)  & \underline{79.62}          & 65.87          & 72.75                    & \underline{59.19}          & 52.48                    & 55.84 \\
Co-T + ASource  ($\XPtrain \& \VGds$)  & \underline{\textbf{79.99}} & \textbf{69.01} & \textbf{74.50}           & \underline{\textbf{59.99}} & \textbf{55.39}           & \textbf{57.69} \\
\midrule
$\Delta_{\mbox{(Co-T + ASource) vs Source}}$        
                                       & +17.72                     & +07.73         & +10.22                   & +21.11                     & +02.02                   & +11.56         \\
$\Delta_{\mbox{(Co-T + ASource) vs ASource}}$      
                                       & +04.47                     & +07.07         & +05.77                   & +07.57                     & +02.31                   & +04.94         \\
$\Delta_{\mbox{(Co-T + ASource) vs upp.-b.}^\dagger}$
                                       & -04.09                     & +03.01         & -00.54                   & -01.72                     & -02.35                   & -02.04         \\
$\Delta_{\mbox{(Co-T + Source) vs Source}}$
                                       & +11.26                     & +08.22         & +07.22                   & +09.68                     & +02.96                   & +06.32         \\
$\Delta_{\mbox{ASource vs Source}}$
                                       & +13.25                     & +00.66         & +04.45                   & +13.54                     & -00.29                   & +06.62         \\
\bottomrule
\end{tabular}
}
\end{table}

We can see that, in all the cases, the self-labeling methods outperform the lower-bounds. For vehicles (V), self-training and co-training perform similarly for the 5\% lower-bound, while for the 10\% self-training performs better than co-training in $\Kds$ but worse in $\Wds$. For pedestrians (P), co-training always performs better. Looking at the vehicle-pedestrian combined (V\&P) performance, we see significant improvements over the lower-bounds. Interestingly, for the 10\% setting, co-training even outperforms the upper-bound for pedestrians in $\Wds$. In fact, in this case, the corresponding V\&P performance is just 1.55 points below the upper-bound. The same setting in $\Kds$, improves 5.3 points over the lower-bound, but is 7.4 points below the upper-bound. These experiments show that our self-labeling algorithms, especially co-training, are performing the task of SSL reasonably well, which encourages to address the UDA challenge with them.   

Table \ref{tab:Synth2Real} shows the UDA results for $\Vds \rightarrow \Kds$ and $\Vds \rightarrow \Wds$, thus, training with $\Vds$ acts as lower-bound. Just as reference, we also show the results of training on $\Kds$ and testing on $\Wds$, and vice versa. The former case shows a similar domain shift as when training on $\Vds$. The latter case shows a significant lower shift from $\Vds$ to $\Kds$, than from $\Wds$ to $\Kds$. Thus, we think that $\Vds$ offers a realistic use case to assess virtual-to-real UDA. The $\Delta_X$~rows at the bottom block of Table \ref{tab:Synth2Real} summarize numerically the main insights.

\begin{figure}
\centering
\includegraphics[trim=0 5 0 7, clip, width=0.8\columnwidth]{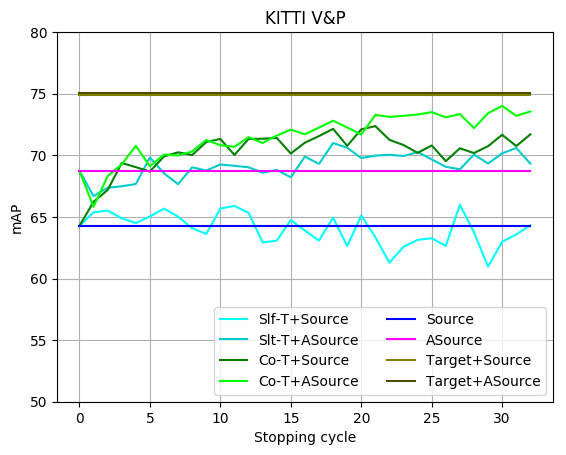}\\
\includegraphics[trim=0 7 0 5, clip, width=0.8\columnwidth]{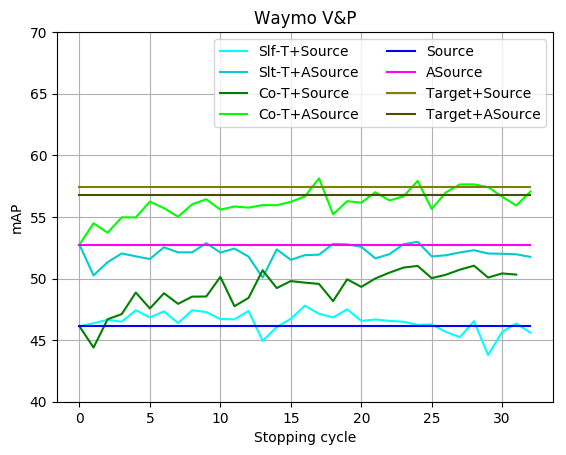}
\caption{Eventual detection performance (mAP) of self-training and co-training as a function of the stopping cycle, in the UDA setting. Upper- and lower-bounds are included as visual reference. We refer to the main text for more details.}
\label{fig:cycleplots}
\end{figure}

\begin{table}
\centering
\caption{Number of self-labeled vehicles and pedestrians applying self-training and co-training, for the SSL  (5\% \& 10\%) and UDA (Source \& ASource) settings, for KITTI ($\Kds$) and Waymo ($\Wds$). In parenthesis we indicate the percentage of false positives. The top block correspond to ground truth labels in the full training sets, and the percentages used for SSL. After removing false positives, in each block of rows, the corresponding $\Delta_{X}$ shows how many more objects are labeled by co-training compared to self-training.}\label{tab:self-labeledBBs}
\resizebox{1.0\linewidth}{!}{
\begin{tabular}{@{}lccccccc@{}}
\toprule
                                      & \multicolumn{2}{c}{$\Kds$}      & \multicolumn{2}{c}{$\Wds$} \\
Training set                          & V              & P              & V              & P \\
\toprule
100 \% Labeled                        & 14,941         & 3,154          & 64,446         &  9,918 \\
5 \% Labeled                          &    647         &    83          & 3,520          &    472 \\
10 \% Labeled                         &  1,590         &   367          & 6,521          &    959 \\
\midrule
5\% Labeled +      Slf-T              &  9,376 (0.8\%) &   837 (0.5\%)  & 52,150 (1.7\%) &  5,554 (3.1\%) \\
5\% Labeled +      Co-T               & 10,960 (1.8\%) & 1,591 (6.2\%)  & 56,102 (4.0\%) &  6,948 (10.9\%) \\
$\Delta_{5\%}$                        & +1,462         &  +660		    & +2,594         & 	 +809 \\
\midrule
10\% Labeled +      Slf-T             & 10,536 (0.7\%) & 1,409 (2.3\%)  & 54,698 (1.6\%) &  6,519 (2.7\%) \\
10\% Labeled +      Co-T              & 11,309 (1.7\%) & 1,552 (3.9\%)  & 56,686 (3.1\%) &  7,384 (6.0\%) \\
$\Delta_{10\%}$                       &   +654		   &  +115   		& +1,106 		 &   +598 \\
\midrule
Slf-T + Source                        & 12,686 (4.6\%) & 1,378 (8.9\%)  & 29,036 (23.8\%) &    837 (5.1\%) \\
Co-T  + Source                        & 14,537 (6.5\%) & 1,964 (14.0\%) & 48,653 (30.4\%) &  3,481 (12.9\%) \\
$\Delta_{Source}$                     & +1,490		   &  +434			& +1,1737		  & +2,238 \\
\midrule
Slf-T + ASource                       & 10,389 (1.7\%) & 1,268 (3.5\%)  & 41,173 (24.7\%) &  1,928 (13.5\%) \\
Co-T  + ASource                       & 12,864 (3.1\%) & 1,532 (5.2\%)  & 53,955 (28.8\%) &  3,553 (16.0\%) \\
$\Delta_{ASource}$                    & +2,253		   &  +229			& +7,413		  & +1,317 \\
\bottomrule
\end{tabular}
}
\end{table}

\begin{figure*}
\setlength{\tabcolsep}{1pt}
\centering
\begin{tabular}{cccccc}
\centeredcell{\includegraphics[trim=750 20 200 80, clip, height=1.65cm]{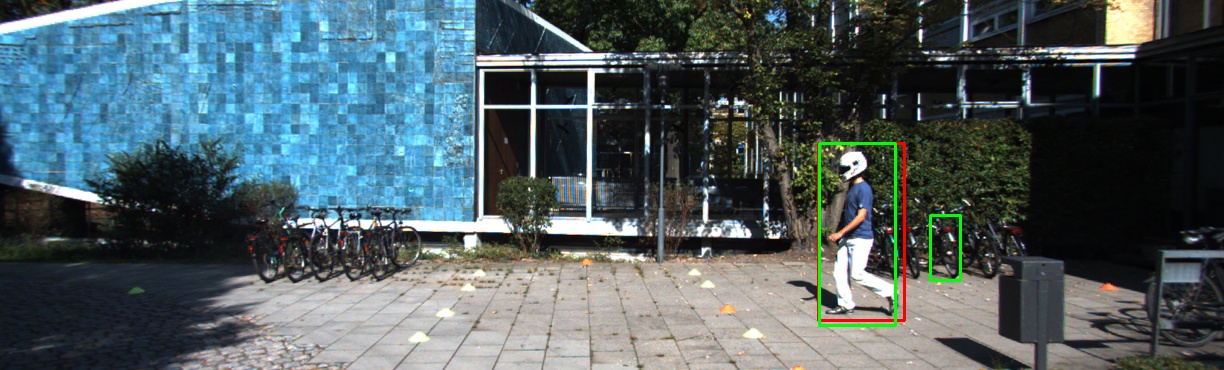}} &
\centeredcell{\includegraphics[trim=750 20 200 80, clip, height=1.65cm]{images/training_kitti/gan/000209.jpeg}} &
\centeredcell{\includegraphics[trim=520 135 600 140, clip, height=1.65cm]{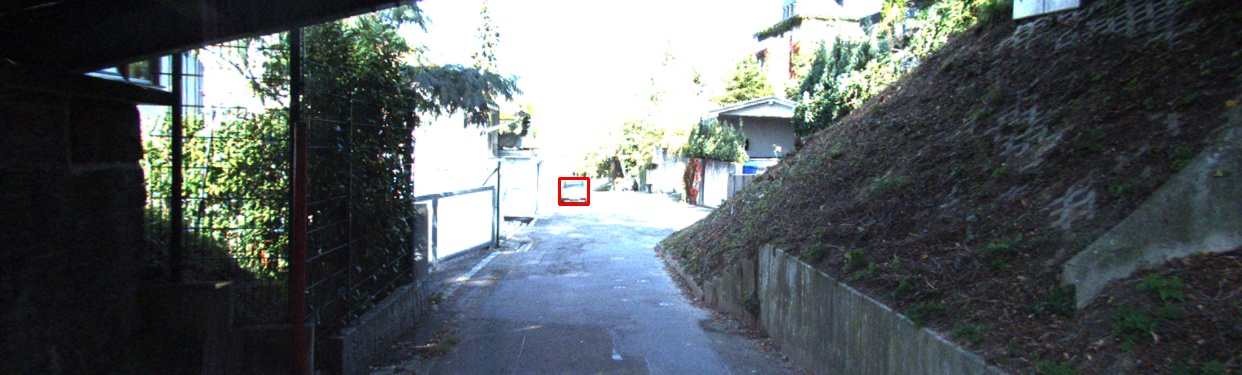}} &
\centeredcell{\includegraphics[trim=520 135 600 140, clip, height=1.65cm]{images/training_kitti/gan/000030.jpeg}} &
\centeredcell{\includegraphics[trim=690 75 0 100, clip, height=1.65cm]{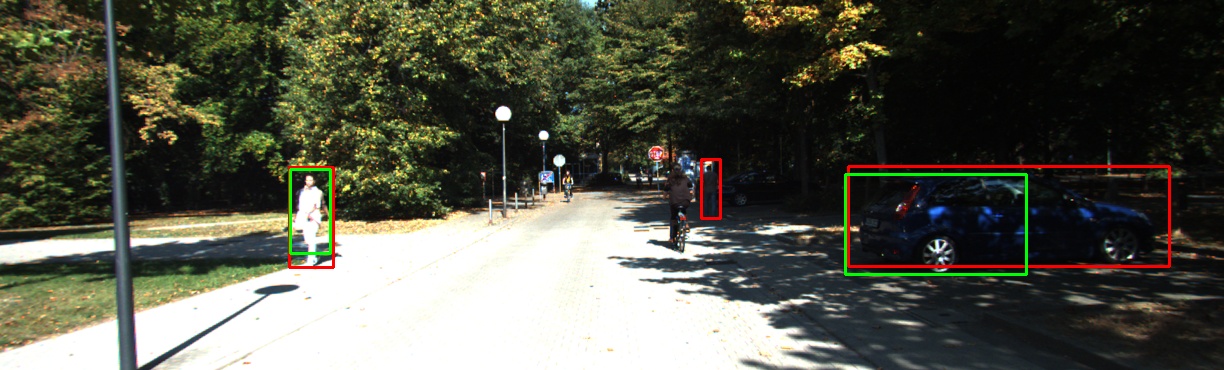}} &
\centeredcell{\includegraphics[trim=690 75 0 100, clip, height=1.65cm]{images/training_kitti/gan/000460.jpeg}} \\
\centeredcell{\includegraphics[trim=750 20 200 80, clip, height=1.65cm]{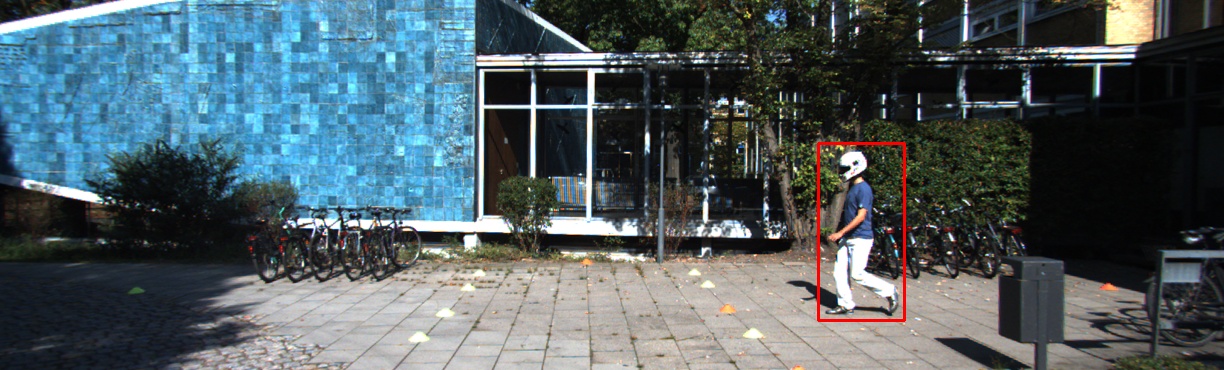}} &
\centeredcell{\includegraphics[trim=750 20 200 80,clip,height=1.65cm]{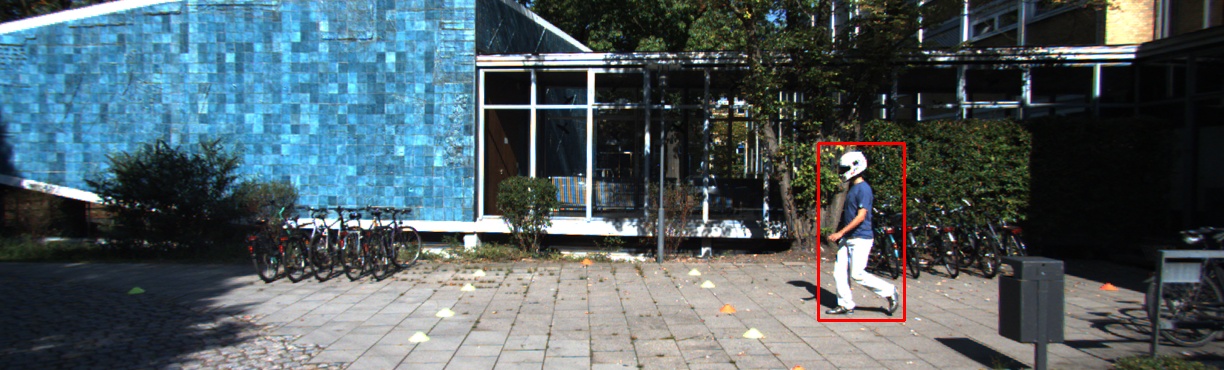}} &
\centeredcell{\includegraphics[trim=520 135 600 140, clip, height=1.65cm]{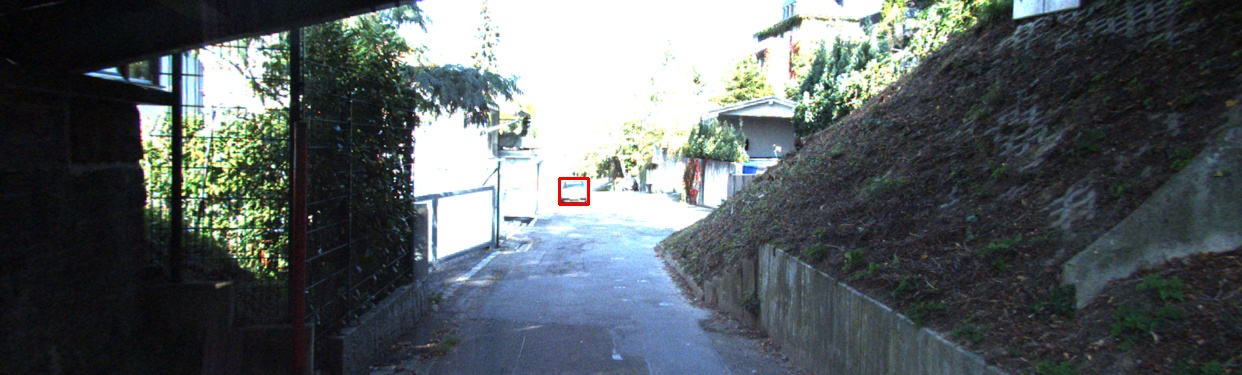}} &
\centeredcell{\includegraphics[trim=520 135 600 140, clip, height=1.65cm]{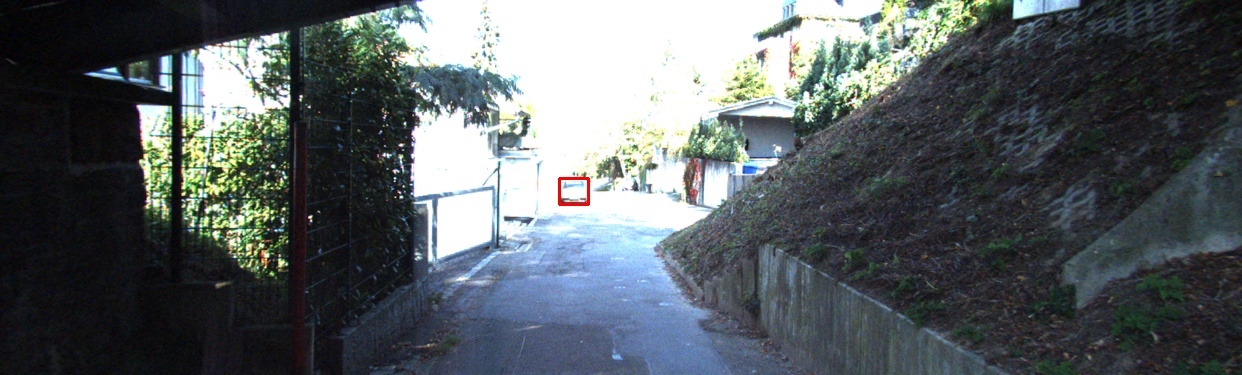}}  &
\centeredcell{\includegraphics[trim=690 75 0 100, clip, height=1.65cm]{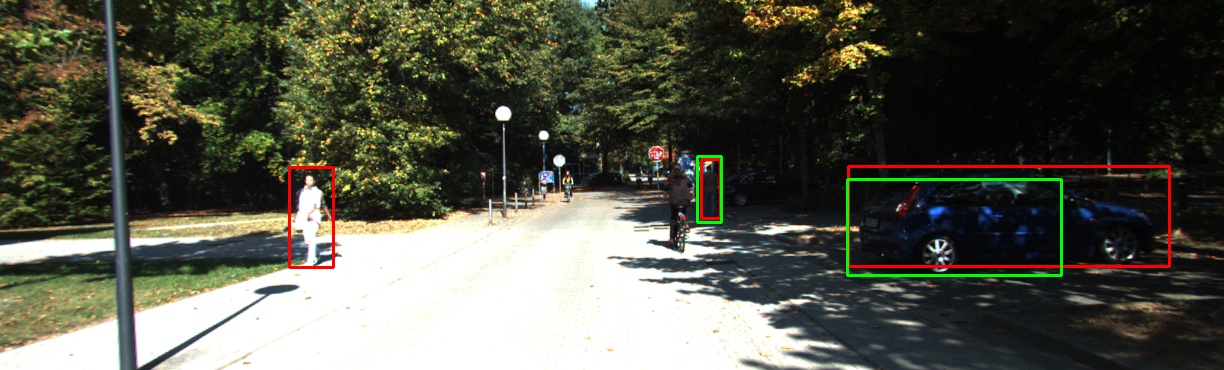}} &
\centeredcell{\includegraphics[trim=690 75 0 100, clip, height=1.65cm]{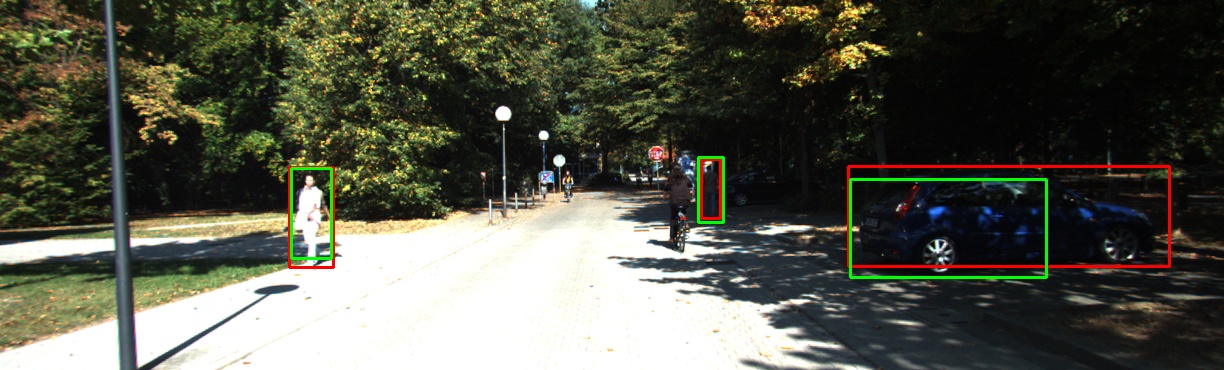}}\\
\centeredcell{\includegraphics[trim=750 20 200 80, clip, height=1.65cm]{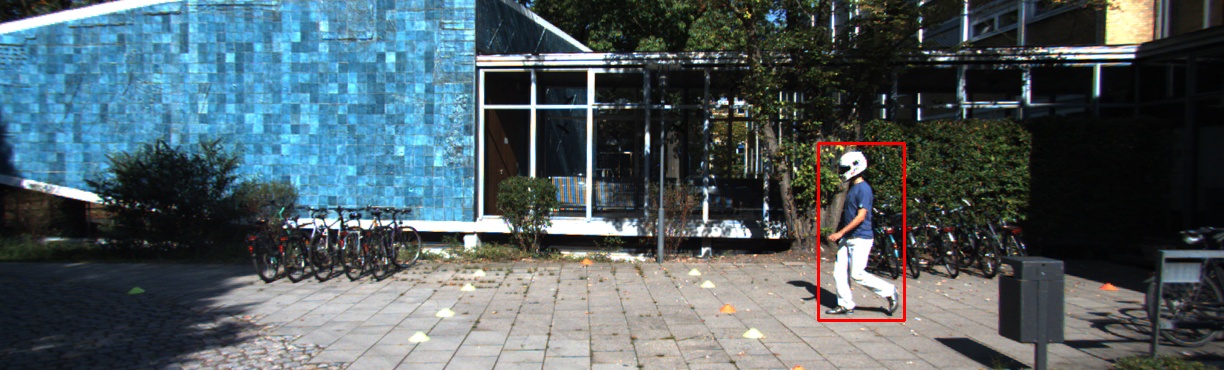}} &
\centeredcell{\includegraphics[trim=750 20 200 80, clip, height=1.65cm]{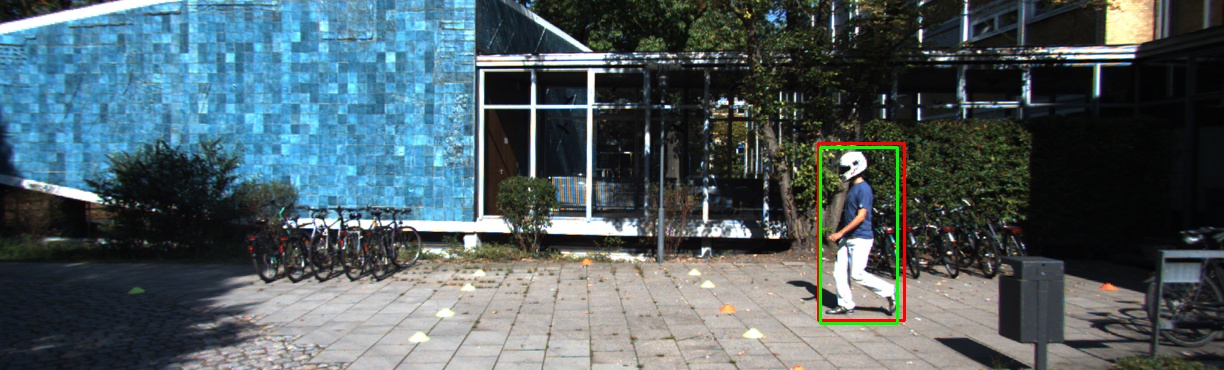}} & 
\centeredcell{\includegraphics[trim=520 135 600 140, clip, height=1.65cm]{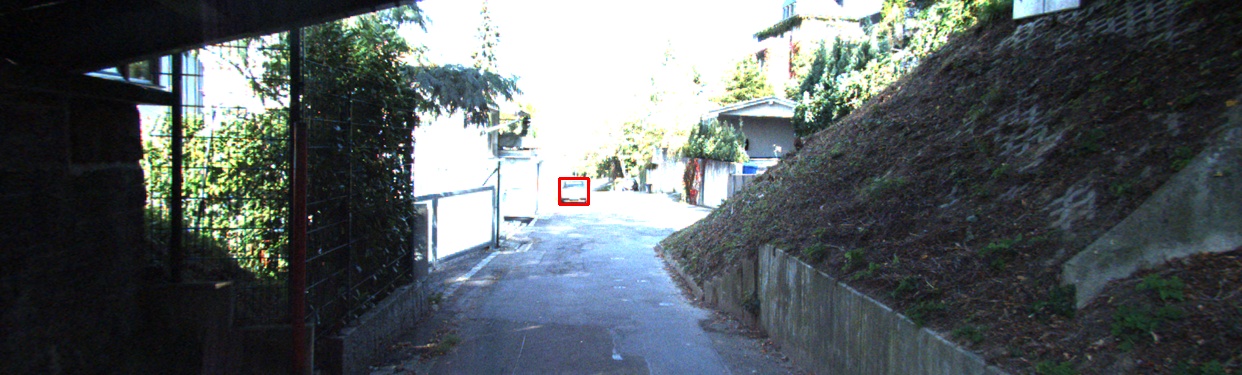}} &
\centeredcell{\includegraphics[trim=520 135 600 140, clip, height=1.65cm]{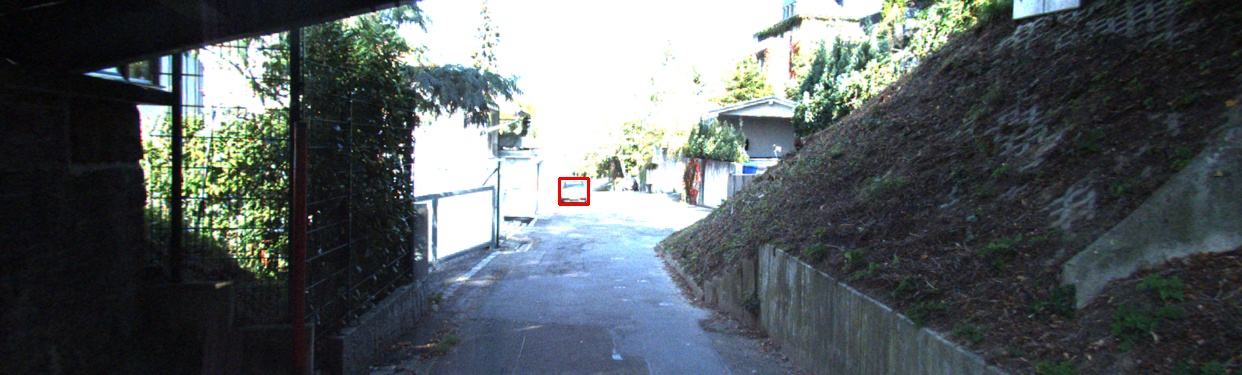}} & 
\centeredcell{\includegraphics[trim=690 75 0 100, clip, height=1.65cm]{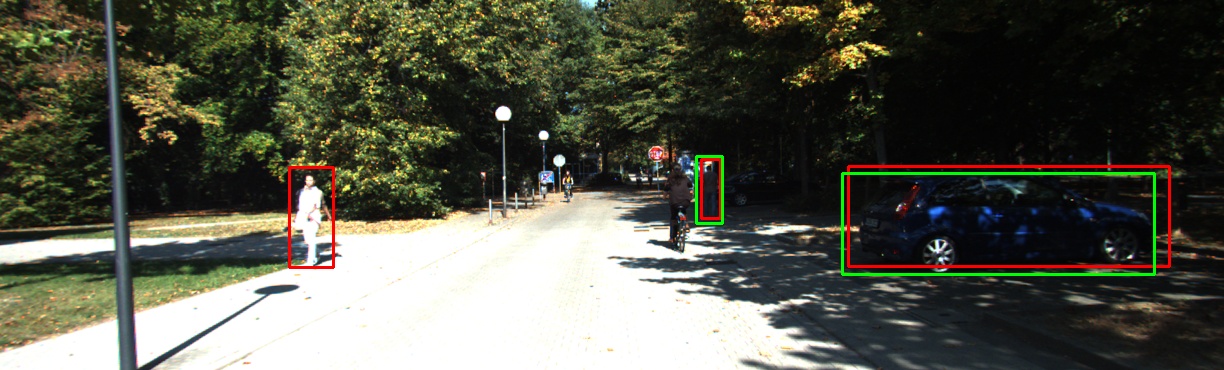}} &
\centeredcell{\includegraphics[trim=690 75 0 100, clip, height=1.65cm]{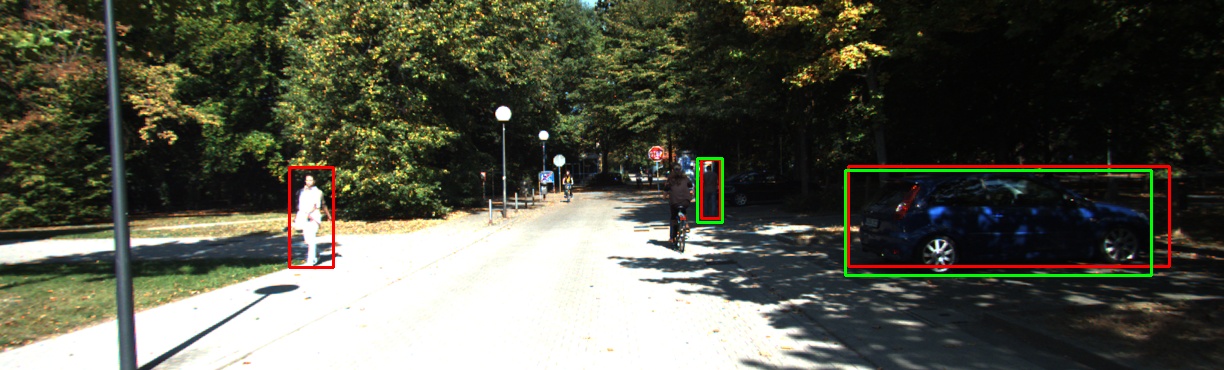}}\\
\centeredcell{\includegraphics[trim=750 20 200 80, clip, height=1.65cm]{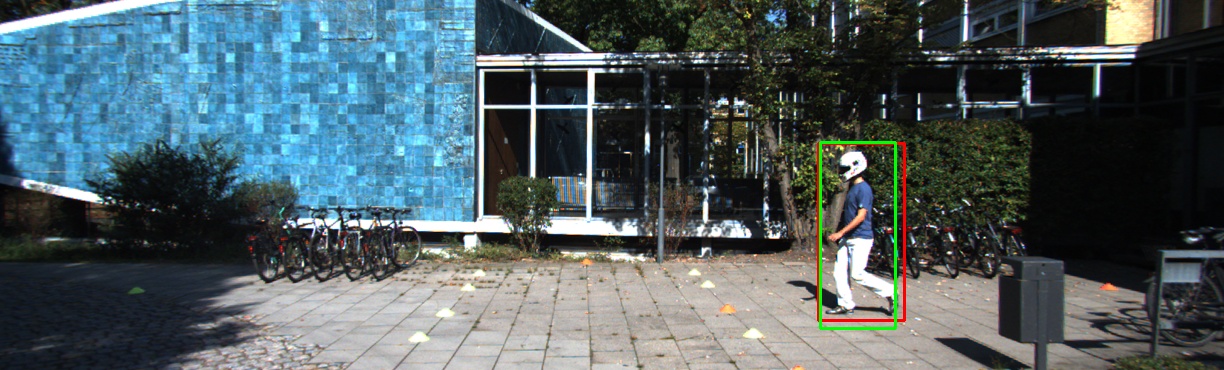}} &
\centeredcell{\includegraphics[trim=750 20 200 80,clip,height=1.65cm]{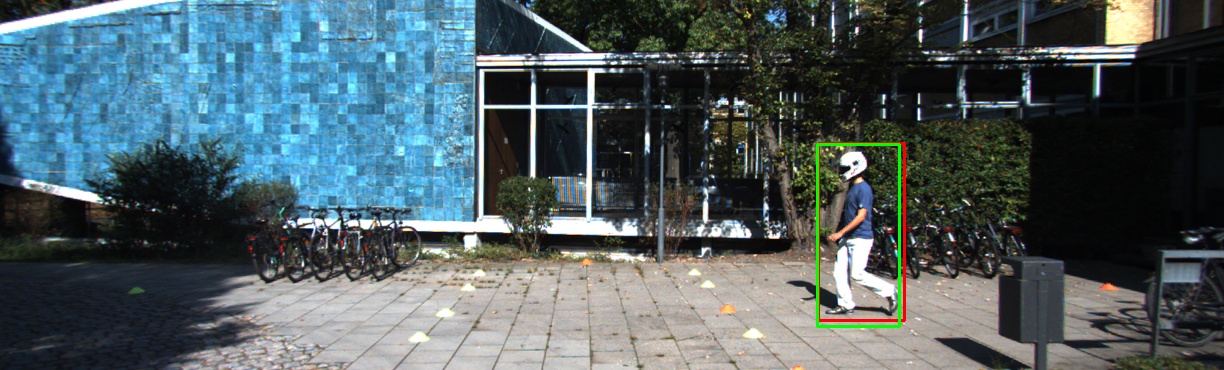}} &
\centeredcell{\includegraphics[trim=520 135 600 140, clip, height=1.65cm]{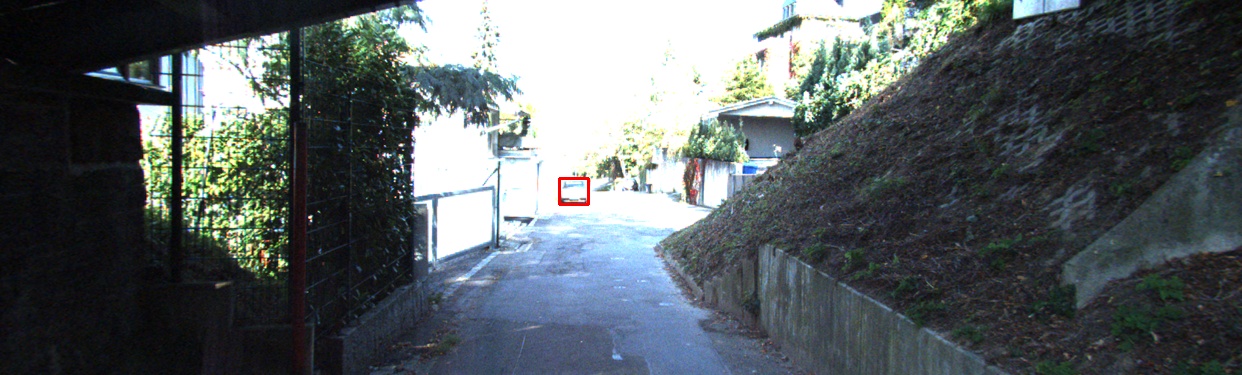}} &
\centeredcell{\includegraphics[trim=520 135 600 140, clip, height=1.65cm]{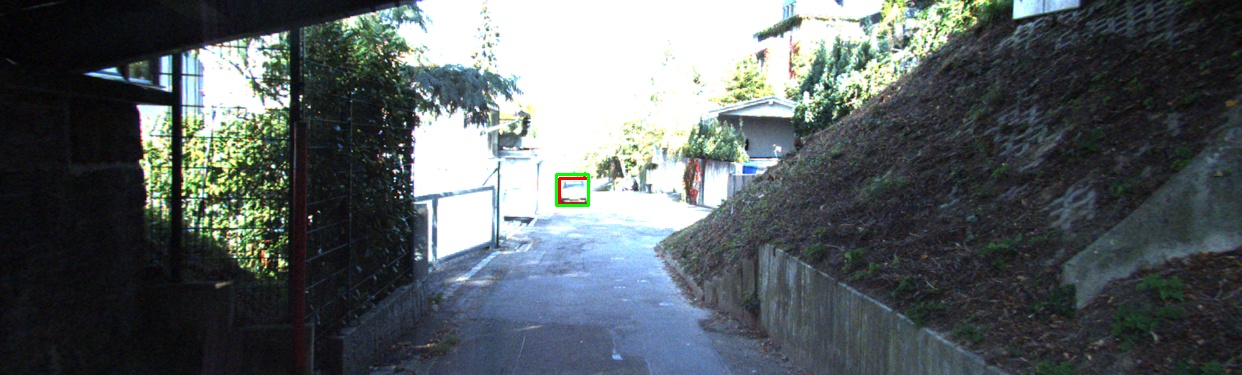}} &
\centeredcell{\includegraphics[trim=690 75 0 100, clip, height=1.65cm]{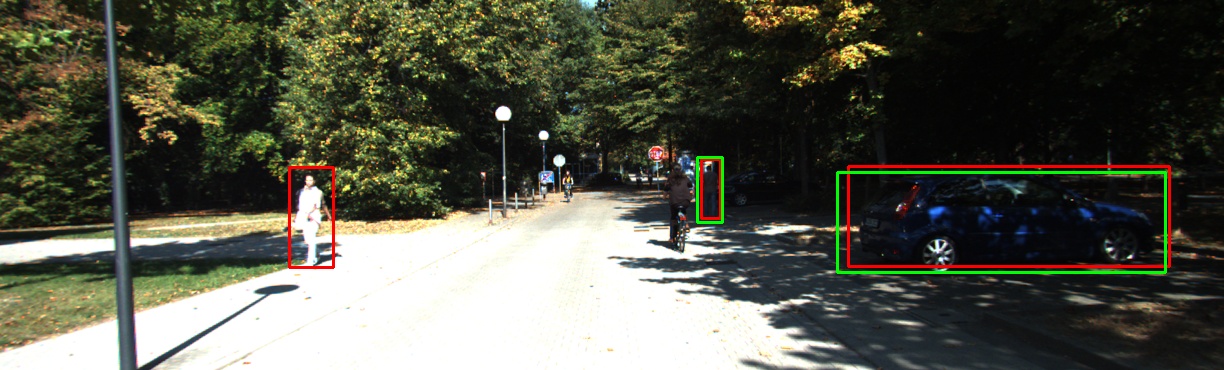}} &
\centeredcell{\includegraphics[trim=690 75 0 100, clip, height=1.65cm]{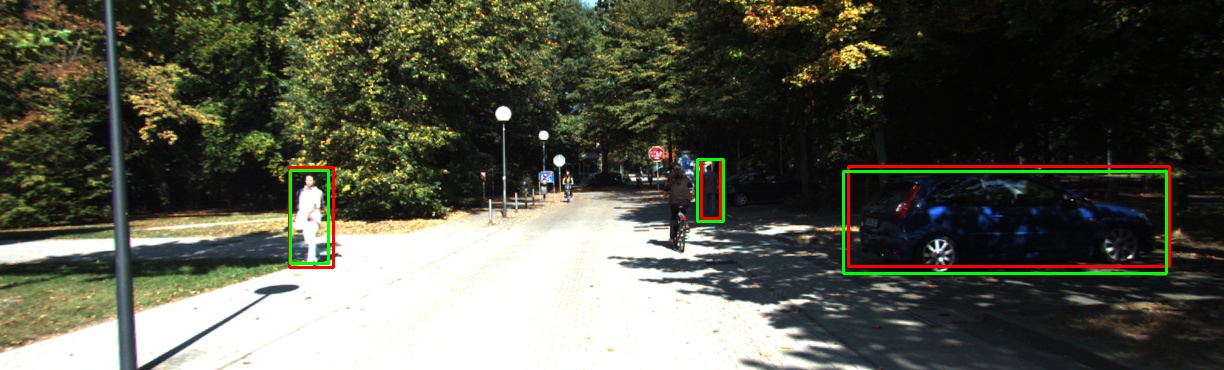}}\\
\end{tabular}
\caption{Examples of Self/Co-training + ASource. Red BBs are from the ground truth of $\Kdstrain$, and green BBs are predicted. Each block of two columns with the same underlying image compares self-training (left column of the block) and co-training (right column of the block). Top row corresponds to detections in Cycle 0, when, in these examples, the only available training data is $\VGKds$ (so it is the same for self-training and co-training). The following rows, top to bottom, correspond to detections from cycles 1, 10, and 20, respectively, when self-labeled images are incrementally added to the training set.} 
\label{fig:training_evolution_kitti}
\end{figure*}

\begin{figure*}
\setlength{\tabcolsep}{1pt}
\centering
\begin{tabular}{ccccccc}
\centeredcell{\includegraphics[trim=560 180 450 80, clip, height=1.65cm]{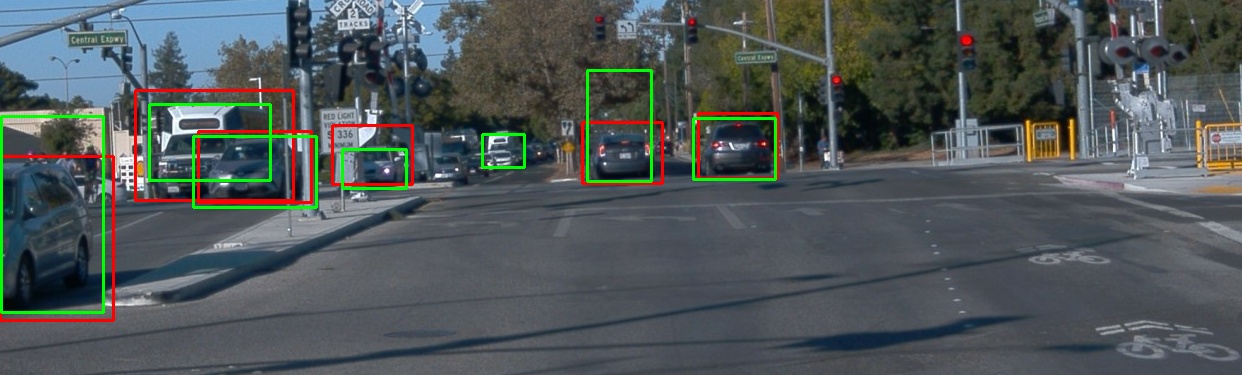}} &
\centeredcell{\includegraphics[trim=560 180 450 80, clip, height=1.65cm]{images/training_waymo/gan/test_segment-6324079979569135086_2372_300_2392_300_with_camera_labels_0000002.jpeg}} &
\centeredcell{\includegraphics[trim=50 0 830 140, clip, height=1.65cm]{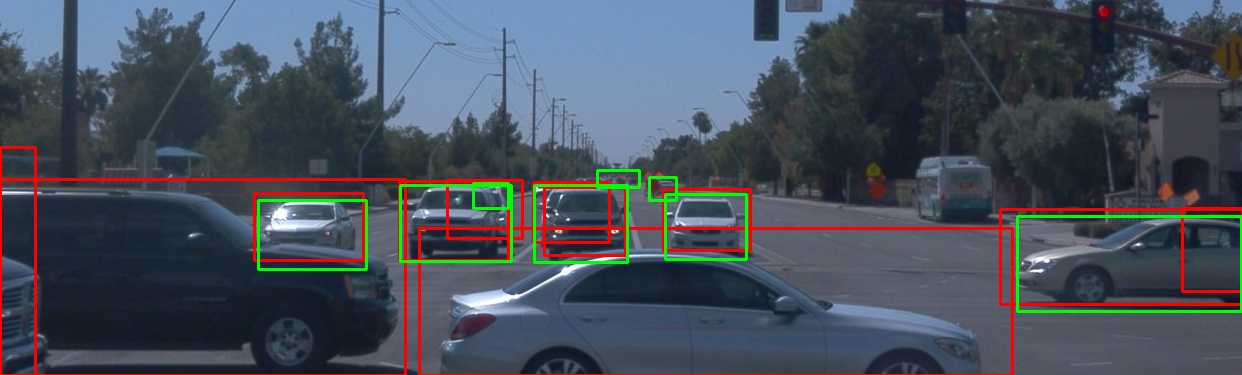}} &
\centeredcell{\includegraphics[trim=50 0 830 140, clip, height=1.65cm]{images/training_waymo/gan/test_segment-2094681306939952000_2972_300_2992_300_with_camera_labels_0000140.jpeg}} &
\centeredcell{\includegraphics[trim=635 110 405 130, clip, height=1.65cm]{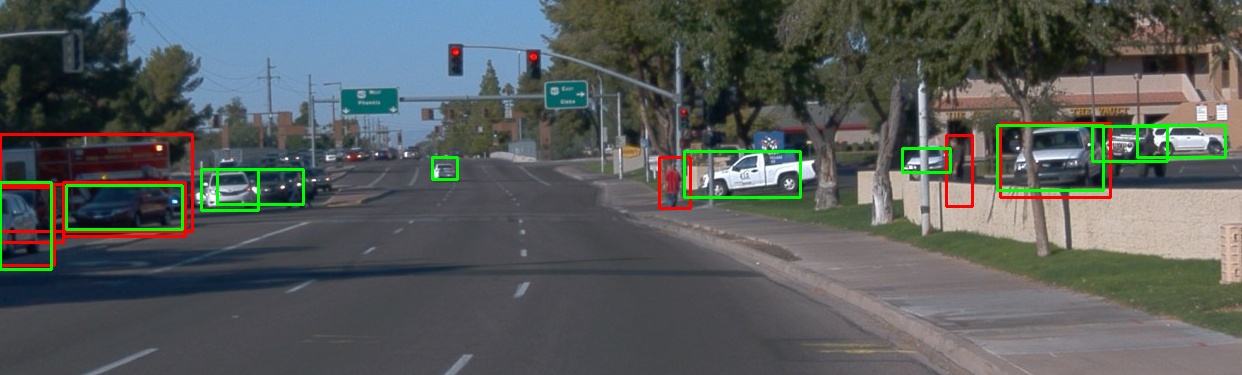}} &
\centeredcell{\includegraphics[trim=635 110 405 130, clip, height=1.65cm]{images/training_waymo/gan/test_segment-4423389401016162461_4235_900_4255_900_with_camera_labels_0000154.jpeg}} \\
\centeredcell{\includegraphics[trim=560 180 450 80, clip, height=1.65cm]{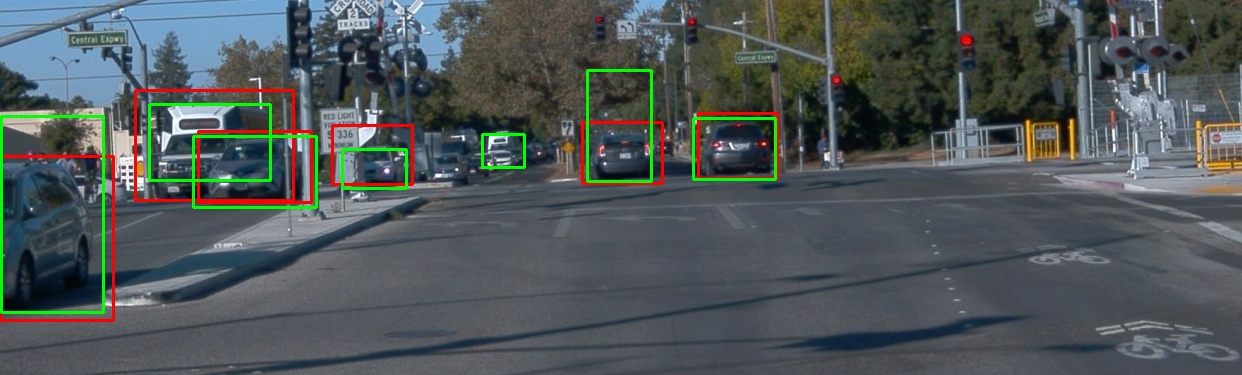}} &
\centeredcell{\includegraphics[trim=560 180 450 80,clip,height=1.65cm]{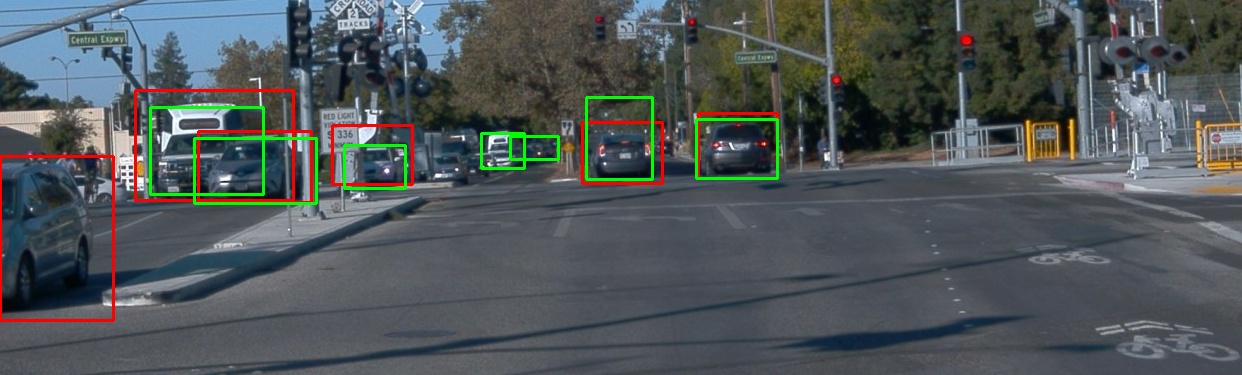}} &
\centeredcell{\includegraphics[trim=50 0 830 140, clip, height=1.65cm]{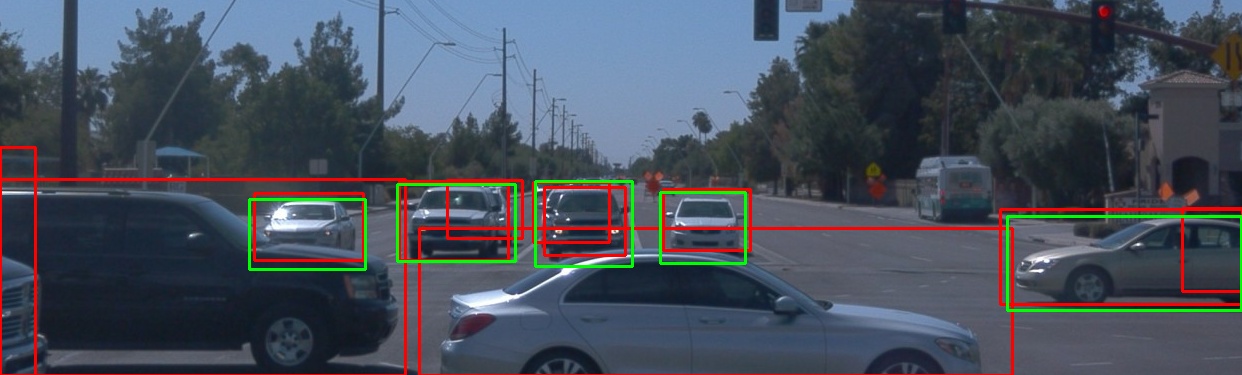}} &
\centeredcell{\includegraphics[trim=50 0 830 140, clip, height=1.65cm]{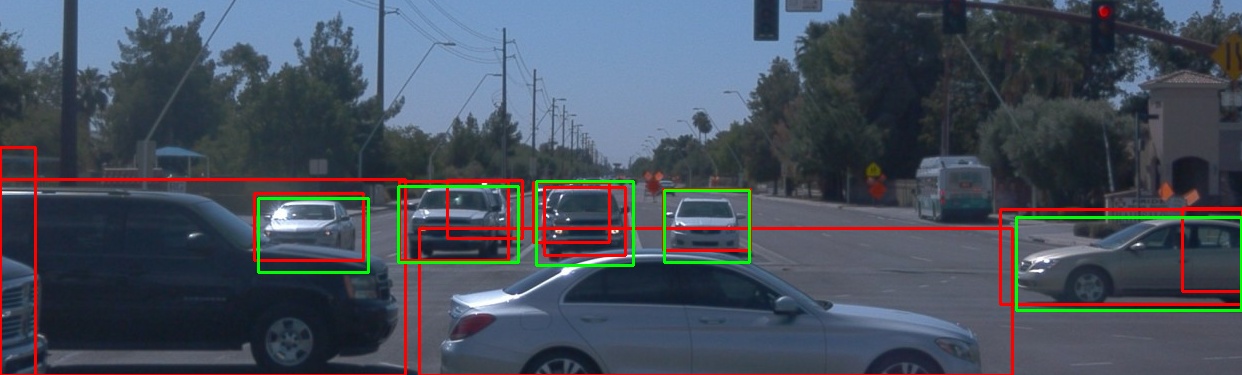}}  &
\centeredcell{\includegraphics[trim=635 110 405 130, clip, height=1.65cm]{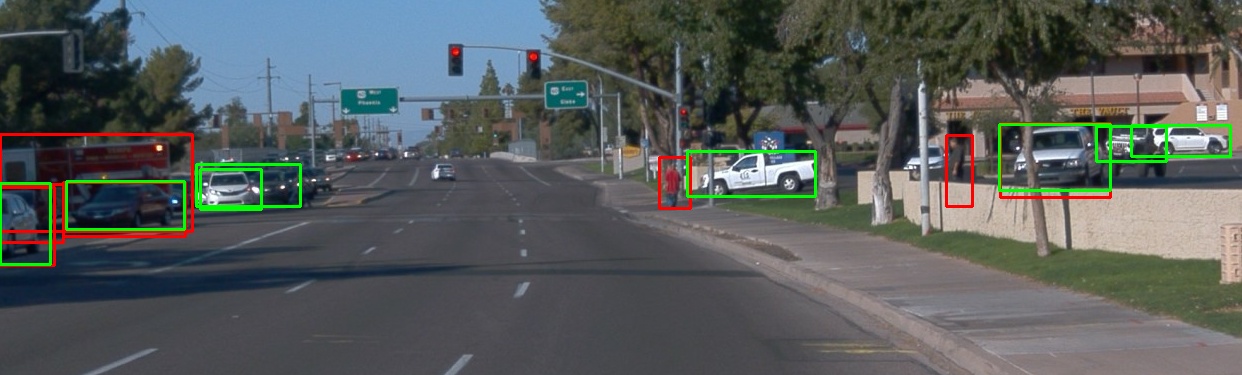}} &
\centeredcell{\includegraphics[trim=635 110 405 130, clip, height=1.65cm]{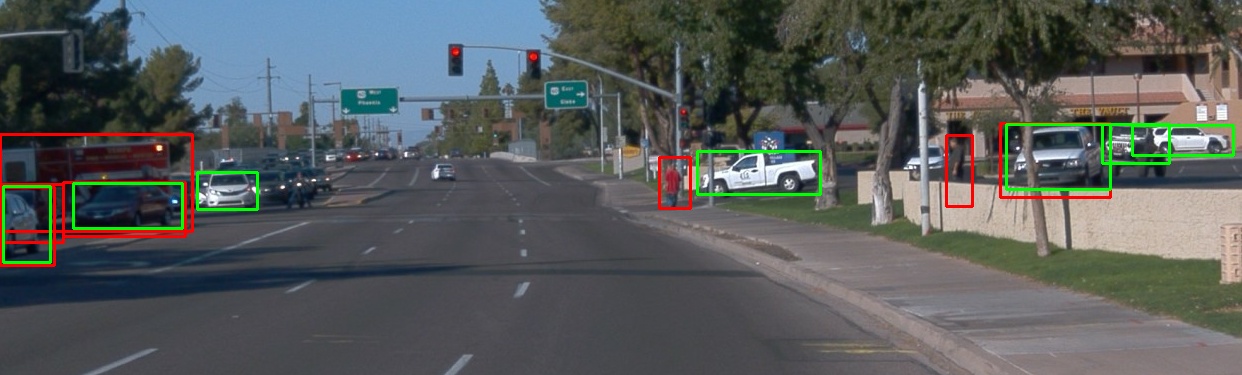}}\\
\centeredcell{\includegraphics[trim=560 180 450 80, clip, height=1.65cm]{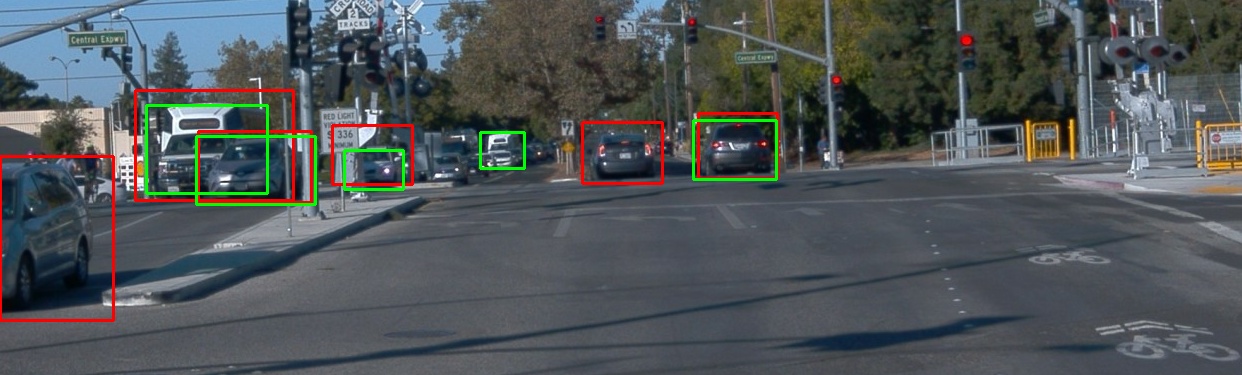}} &
\centeredcell{\includegraphics[trim=560 180 450 80, clip, height=1.65cm]{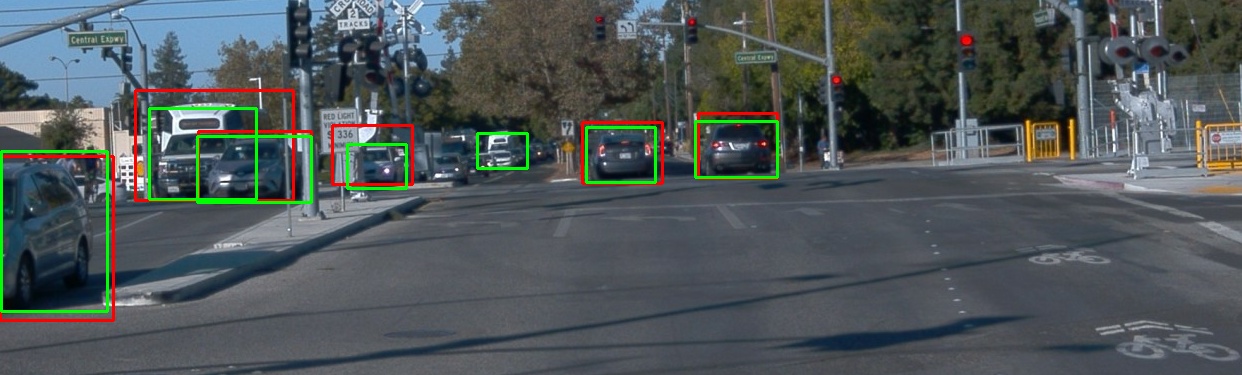}} & 
\centeredcell{\includegraphics[trim=50 0 830 140, clip, height=1.65cm]{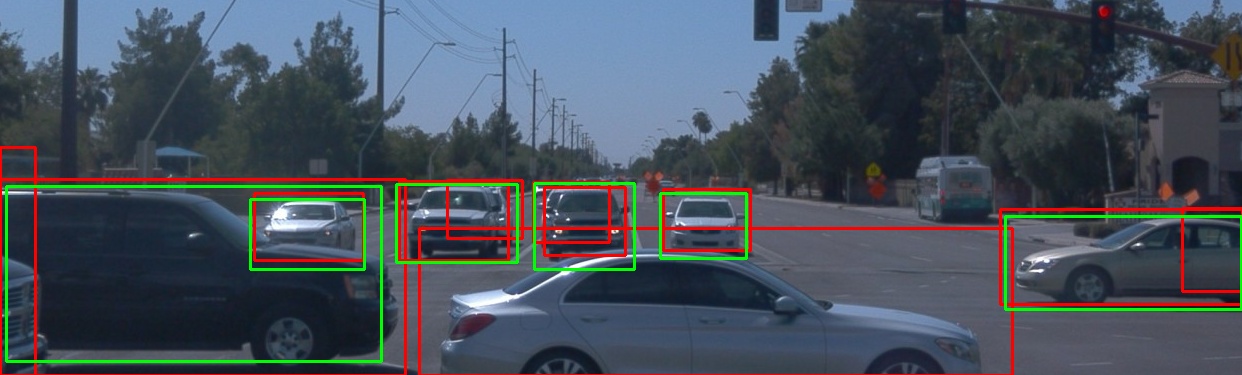}} &
\centeredcell{\includegraphics[trim=50 0 830 140, clip, height=1.65cm]{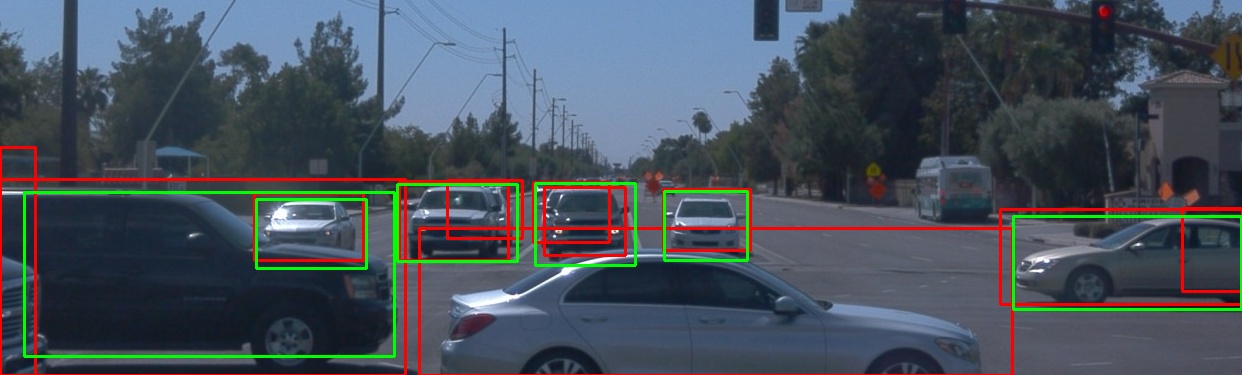}} & 
\centeredcell{\includegraphics[trim=635 110 405 130, clip, height=1.65cm]{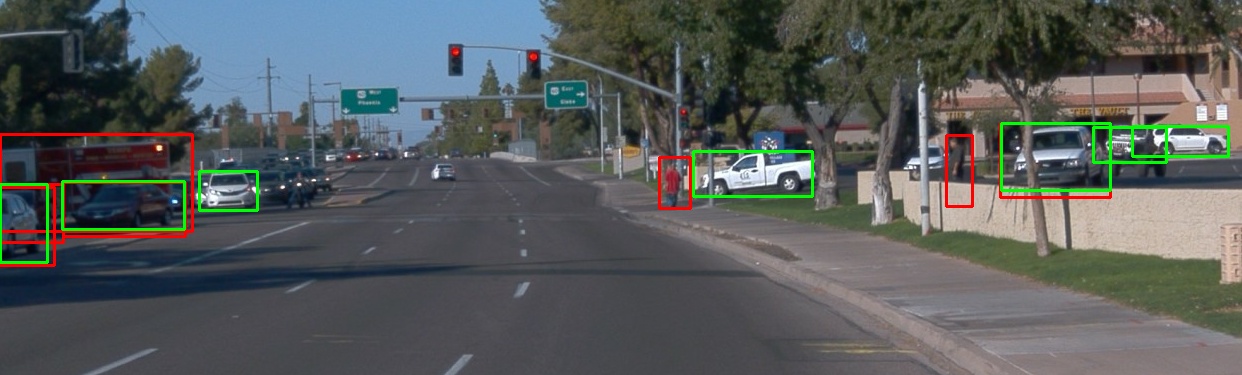}} &
\centeredcell{\includegraphics[trim=635 110 405 130, clip, height=1.65cm]{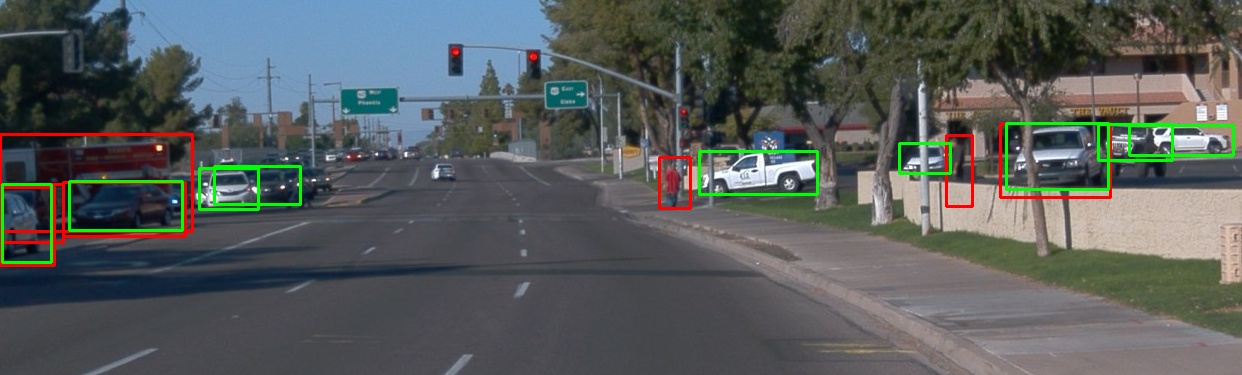}}\\
\centeredcell{\includegraphics[trim=560 180 450 80, clip, height=1.65cm]{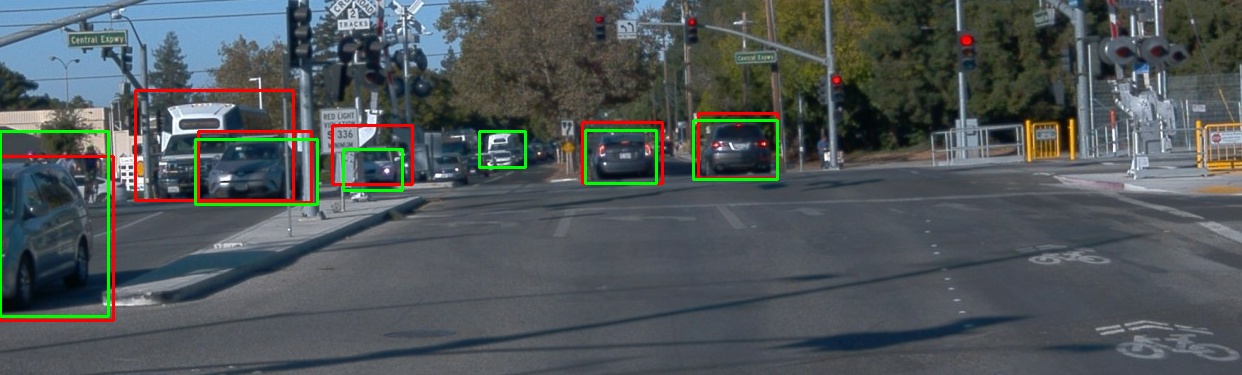}} &
\centeredcell{\includegraphics[trim=560 180 450 80,clip,height=1.65cm]{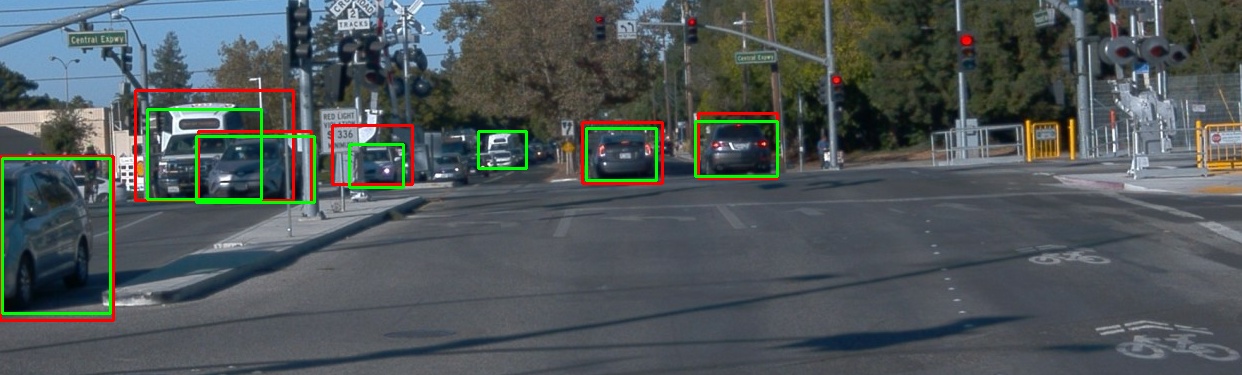}} &
\centeredcell{\includegraphics[trim=50 0 830 140, clip, height=1.65cm]{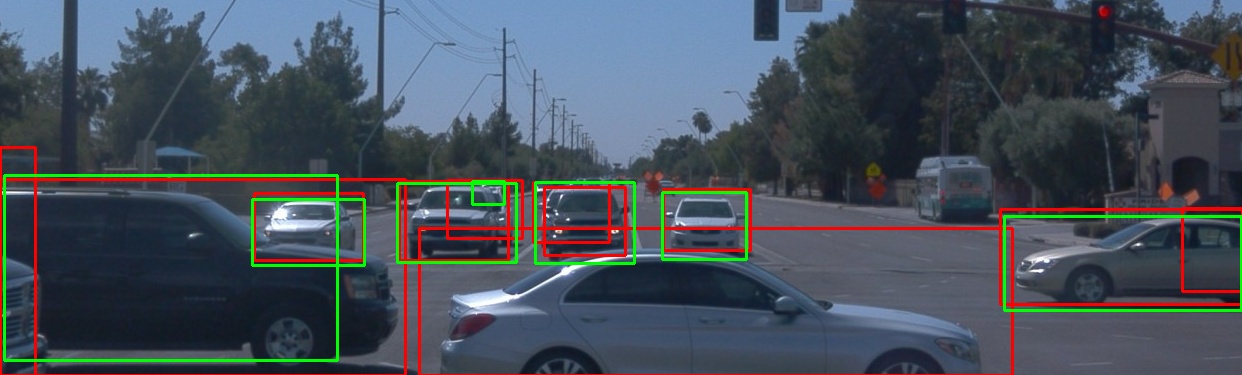}} &
\centeredcell{\includegraphics[trim=50 0 830 140, clip, height=1.65cm]{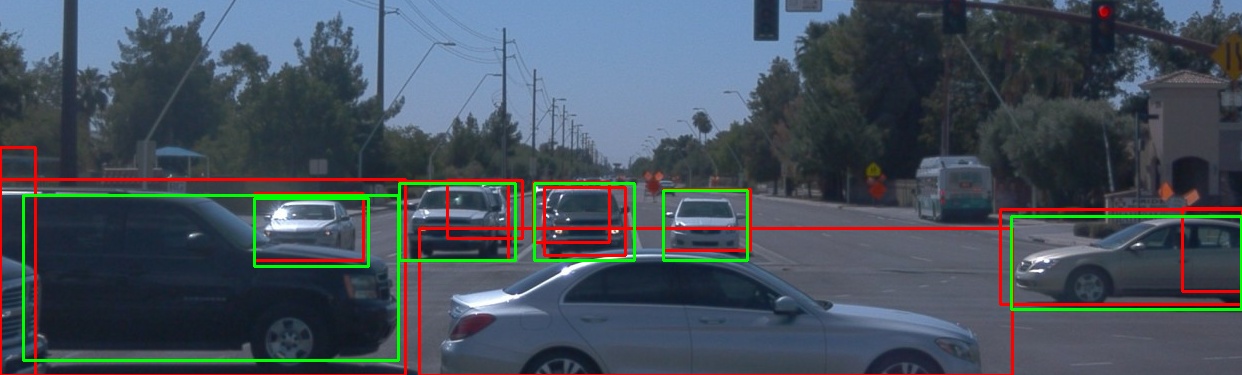}} &
\centeredcell{\includegraphics[trim=635 110 405 130, clip, height=1.65cm]{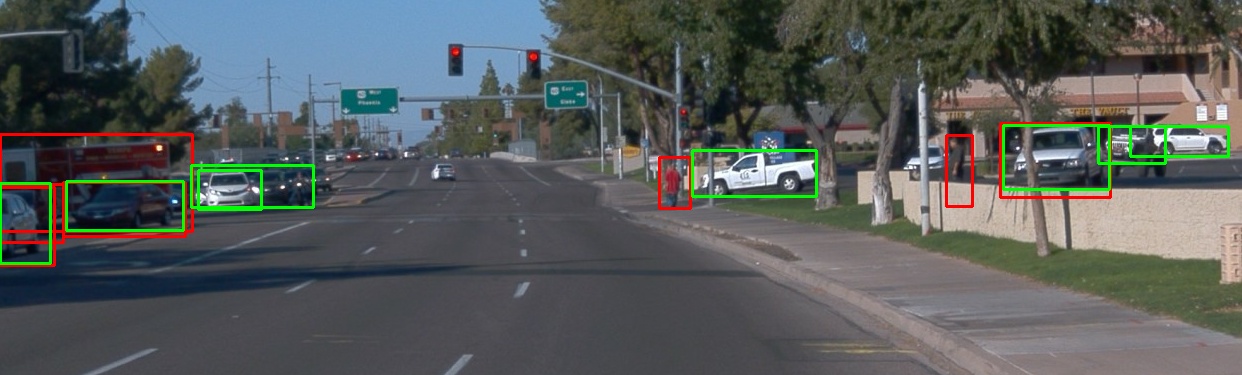}} &
\centeredcell{\includegraphics[trim=635 110 405 130, clip, height=1.65cm]{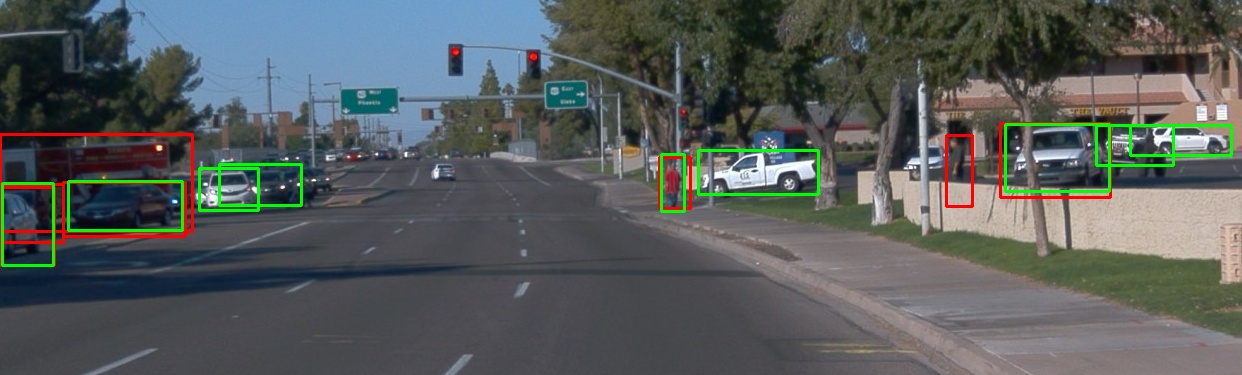}}\\
\end{tabular}
\caption{Analogous to Figure \ref{fig:training_evolution_kitti} for $\Wdstrain$ and $\VGWds$.} 
\label{fig:training_evolution_waymo}
\end{figure*}

A first observation is that, by combining GAN-based image-to-image translation and co-training, we obtain significant performance improvements over the lower-bounds in all cases; in terms of V\&P, 10.22 points for $\Kds$ and 11.56 for $\Wds$. In fact, the gap to reach upper-bound performances, is relatively small compared to the improvement over the lower-bounds; in terms of V\&P, such gap is of 0.54 points for $\Kds$ and of 2.04 for $\Wds$. Note that for $\Kds$, the upper-bound comes from the $\XLtrain \& \VGds$ setting ({\ie} training on the full labeled training set of $\Kds$ plus $\VGKds$); while for $\Wds$, the upper-bound comes from the $\XLtrain$ setting ({\ie} training on the full labeled training set of $\Wds$). Thus, without any manual training data labeling, we are almost reaching upper-bound performance. 

A second observation is that, indeed, co-training brings additional improvements on top of image-to-image translation; in terms of V\&P, 5.77 points for $\Kds$ and 4.94 for $\Wds$. Note that, when applying them separately to the virtual-world images (source domain), both co-training and CycleGAN (ASource) show similar performance improvements for $\Wds$ but co-training outperforms CycleGAN in $\Kds$; in terms of V\&P, co-training improves 7.22 points for $\Kds$ and 6.32 for $\Wds$, while CycleGAN improves 4.45 and 6.62 points, respectively. Interestingly, CycleGAN performs better than co-training for vehicles and it is the opposite for pedestrians. In any case, using both together outperforms them in isolation. 

A third observation is that co-training consistently outperforms self-training. Moreover, in Figure \ref{fig:cycleplots} we can see that this is also consistent along self-labeling cycles (we show the UDA setting). It plots curves illustrating how the self-training and co-training strategies would perform as a function of the stopping cycle. We collect the respective self-labeled images at different cycles ($x$-axis), and train an object detector as these cycles were determined as the stopping ones. Then, we assess the performance of the detector in the corresponding testing set (either from $\Kds$ or $\Wds$), so collecting the corresponding mAP ($y$-axis) per each considered cycle. We see how self-training curves oscillate more around the respective lower-bounds, while co-training ones keep improving performance as we train more cycles. Moreover, in Table \ref{tab:self-labeledBBs}, we can see how co-training systematically self-labels more correct object instances than self-training ($\Delta_X$ rows show the increment for SSL and UDA settings). 

Figures \ref{fig:training_evolution_kitti} and \ref{fig:training_evolution_waymo} show qualitative examples of how self-training and co-training progress in $\Kdstrain$ and $\Wdstrain$ images, respectively; in this case, starting with $\VGKds$ and $\VGWds$ as corresponding initial labeled training sets. Both self-labeling strategies improve over the starting point, since they are able to correct BB localization errors, remove false positives, and recovering from false negatives. In some cases, self-training and co-training show the same final right result, but co-training reaches it in earlier cycles, while in other cases co-training shows better final results.  

As summarized in Table \ref{tab:Synth2RealNoFalsePositives}, with additional experiments, we have further analyzed co-training results. We repeat the training and evaluation of all the detectors developed for both the SSL and UDA settings, considering two variants in the respective training sets. In (/FP) we remove the false positives from the self-labeled data. In (/FP+BB), in addition, for those self-labeled instances that are true positives, we replace the BBs predicted during self-labeling by the corresponding BB ground truth. In this way, we can incrementally analyse the effect of false positives and BB adjustment accuracy. 

Table \ref{tab:Synth2RealNoFalsePositives} shows how the impact of having false positives as training data is not as strong as one may think a priori. For instance, in Table \ref{tab:self-labeledBBs} we can see that co-training with CycleGAN (Co-T + ASource) has output a 28.8\% of false vehicles in $\Wds$ and a 3.1\% in $\Kds$, a large difference; however, Table \ref{tab:Synth2RealNoFalsePositives} indicates that removing them results in a vehicle detection improvement of just 0.71 points in the former case, and 0.17 in the latter. This may be linked with the fact that SGD optimization in deep neural networks (DNNs) tend to prioritize learning patterns instead of noise \cite{Arpit:2017}. 

\begin{table}
\centering
\caption{Results for two new settings: (/FP) assuming we remove the self-labeled false positives; (/FP+BB) assuming that, in addition, for the self-labeled instances, we change the predicted BB by the corresponding one in the ground truth. The $\Delta_X$ rows show differences between these variants and the respective original one ({\ie} neither removing the FP nor adjusting the BBs). The bottom block of rows remarks the differences between the best self-labeling (Co-T+ASource, including /FB and /FB+BB cases) and the upper-bound.}\label{tab:Synth2RealNoFalsePositives}
\resizebox{1.0\linewidth}{!}{
\begin{tabular}{@{}lcccccc@{}}
\toprule
&\multicolumn{3}{c}{$\Xtest=\Kdstest$} & \multicolumn{3}{c}{$\Xtest=\Wdstest$} \\
Training set & V & P & V\&P & V & P & V\&P  \\
\toprule
Upper-bound (UB)              & $^\dagger$84.08 & $^\dagger$66.00 & $^\dagger$75.04 & $^\dagger$61.71 & $^\dagger$57.74 & $^\dagger$59.73 \\
\midrule
5\% Labeled +      Co-T       &  68.99          &  55.07          &  62.03          &  54.00          &  56.34                      &  55.17 \\
5\% Labeled +      Co-T/FP    &  68.37          &  55.34          &  61.86          &  53.94          &  55.23                      &  54.59 \\
5\% Labeled +      Co-T/FP+BB &  \textbf{82.98} &  \textbf{59.24} &  \textbf{71.11} &  \textbf{62.72} &  \textbf{58.54}             &  \textbf{60.63} \\
$\Delta_{\mbox{5\%, FP}}$     &  -0.62          &  +0.27          &  -0.17          &  -0.06          &  -1.11                      &  -0.58 \\
$\Delta_{\mbox{5\%, FP+BB}}$  & +13.99          &  +4.17          &  +9.08          &  +8.72          &  +2.20                      &  +5.46 \\
\midrule
10\% Labeled +      Co-T      &  73.08          &  58.53          &  65.81          &  56.15          &  60.20                      &  58.18 \\
10\% Labeled +      Co-T/FP   &  72.94          &  58.68          &  65.81          &  56.62          &  57.74                      &  57.18 \\
10\% Labeled +      Co-T/FP+BB   &  \textbf{83.16} &  \textbf{61.29} &  \textbf{72.23} &  \textbf{63.17} &  \textbf{60.13}             &  \textbf{61.65} \\
$\Delta_{\mbox{10\%/FP}}$     &  -0.14          &  +0.15          &   0.00          &  +0.47          &  -2.46                      &  -1.00 \\
$\Delta_{\mbox{10\%/FP+BB}}$  & +10.08          &  +2.76          &  +6.42          &  +7.02          &  -0.07                      &  +3.47 \\
\midrule
Co-T + Source                 &  73.53          &  \textbf{69.50} &  71.52          &  48.56          &  \underline{56.33}          &  52.45 \\
Co-T + Source/FP              &  71.64          &  \textbf{69.50} &  70.57          &  51,06          &  \underline{\textbf{56.94}} &  54.00 \\
Co-T + Source/FP+BB           &  \textbf{86.03} &  68.97          &  \textbf{77.50} &  \textbf{59.21} &  \underline{56.59}          &  \textbf{57.90} \\
$\Delta_{\mbox{Co-T + Source/FP}}$   
                              &  -1.89          &  0.00           &  -0.95          &  +2.50          &  +0.61                      &  +1.55 \\
$\Delta_{\mbox{Co-T + Source/FP+BB}}$   
                              & +12.50          & -0.53           &  +5.98          & +10.65          &  +0.26                      &  +5.45\\
\midrule
Co-T + ASource                &  79.99          &  \textbf{69.01} &  74.50          &  59.99          &  55.39                      &  57.69 \\
Co-T + ASource/FP             &  80,16          &  67.79          &  73.98          &  60.70          &  56,87                      &  58,79 \\
Co-T + ASource/FP+BB          &  \textbf{86.26} &  66.69          &  \textbf{76.48} &  \textbf{63.98} &  \textbf{58.55}             &  \textbf{61.27} \\
$\Delta_{\mbox{Co-T + ASource/FP}}$  
                              &  +0.17          &  -1.22          &  -0.52          &  +0.71          &  +1.48                      &  +1.10 \\
$\Delta_{\mbox{FP+FP+BB}}$  
                              &  +6.27          &  -2.32          &  +1.98          &  +3.99          &  +3.16                      &  +3.58 \\
\midrule
$\Delta_{\mbox{(Co-T + ASource) vs UB}}$  
                              &  -4,09	        &  +3,01	      &  -0,54	        &  -1,72	      &  -2,35	                    &  -2,04 \\
$\Delta_{\mbox{(Co-T + ASource/FP) vs UB}}$  
                              &  -3.92	        &  +1.79	      &  -1.06	        &  -1.01	      &  -0.87	                    &  -0.94 \\
$\Delta_{\mbox{(Co-T + ASource/FP+BB) vs UB}}$  
                              &  +2.18          &  -0.69          &  +1.44          &  +2.27          &  +0.81                      &  +1.54 \\
\bottomrule
\end{tabular}
}
\end{table}

\begin{figure}[!t]
\centering
\includegraphics[trim=845 100 100 115, clip, width=0.49\columnwidth]{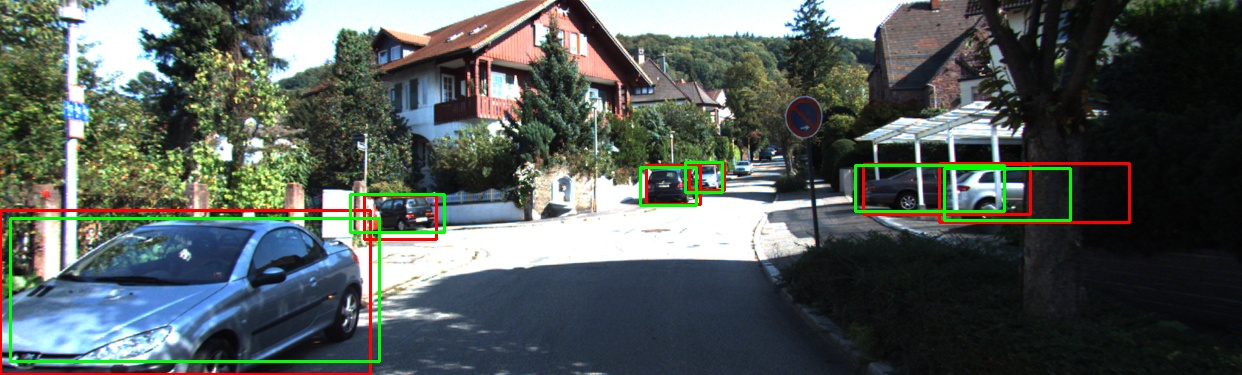}
\includegraphics[trim=355 130 710 150, clip, width=0.49\columnwidth]{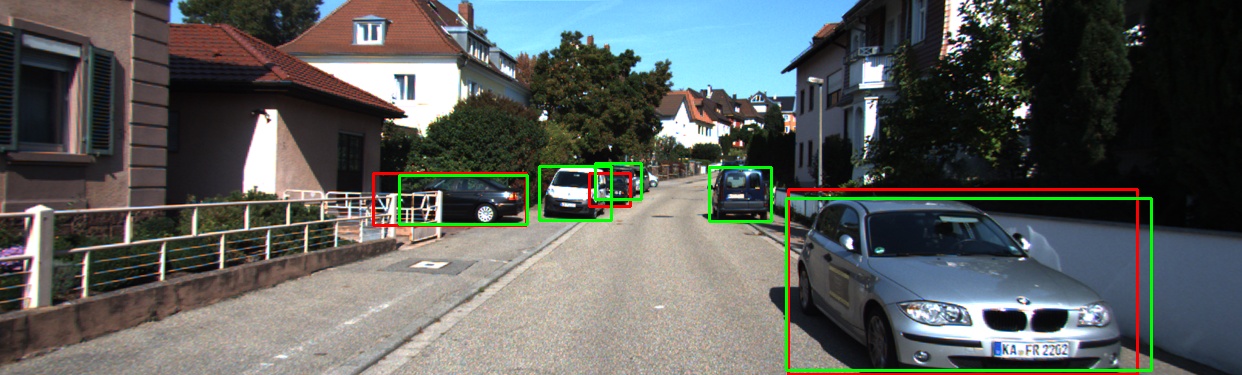}\\
\hspace{0.05mm}\includegraphics[trim=770 80 40 120, clip, width=0.99\columnwidth]{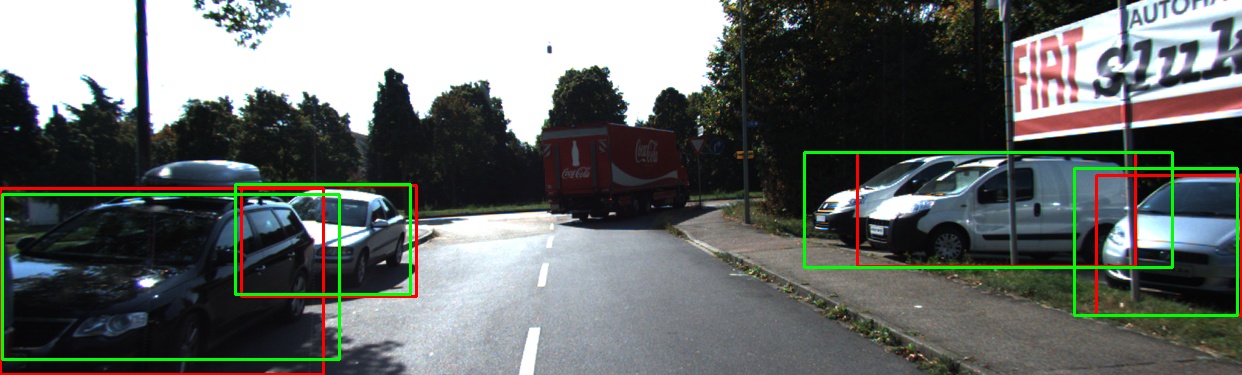}
\caption{Examples of misalignment between ground truth BBs (red) and self-labeled ones (green). Occlusion is the underlying problem, giving rise to shorter BBs (top) or BBs fusing several instances in one (bottom).} 
\label{fig:occlusions}
\end{figure}

In general, as Table \ref{tab:Synth2RealNoFalsePositives} shows, a better BB adjustment has a higher impact than removing false positives, especially for vehicles. In fact, for the higher performing detector, {\ie} co-training with CycleGAN, this adjustment would allow to even outperform the upper-bound. This improvement is mainly coming from the detection of vehicles. For instance, Figure \ref{fig:occlusions} shows examples of the typical BB misalignment we have found, mainly due to occluded vehicle instances. 
Note also that the (/FP+BB) results indicate that non-self-labeled objects (false negatives) do not cause any loss of performance, unless they would result in better BB adjustments. 

\begin{figure*}
\setlength{\tabcolsep}{1pt}
\centering
\begin{tabular}{cccc}
\centeredcell{\includegraphics[trim=220 50 580 120,clip,height=2.2cm]{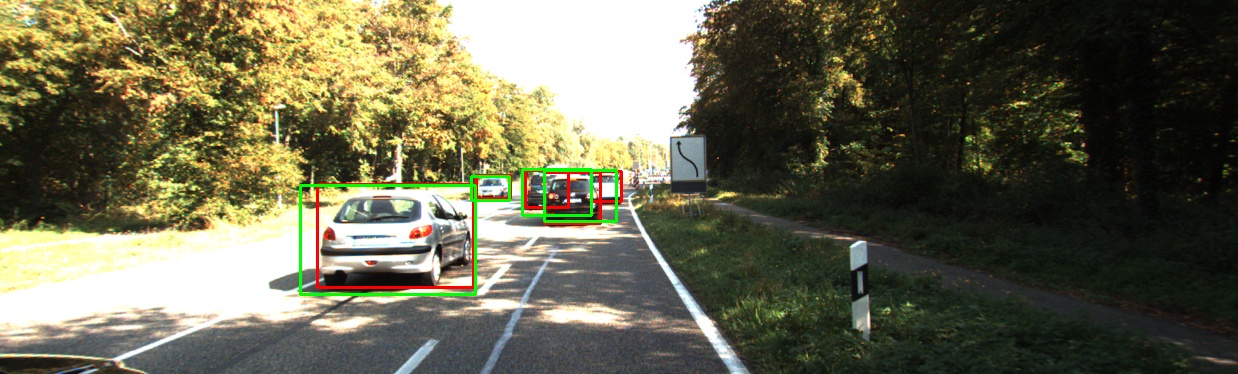}} &
\centeredcell{\includegraphics[trim=250 0 250 100,clip,height=2.2cm]{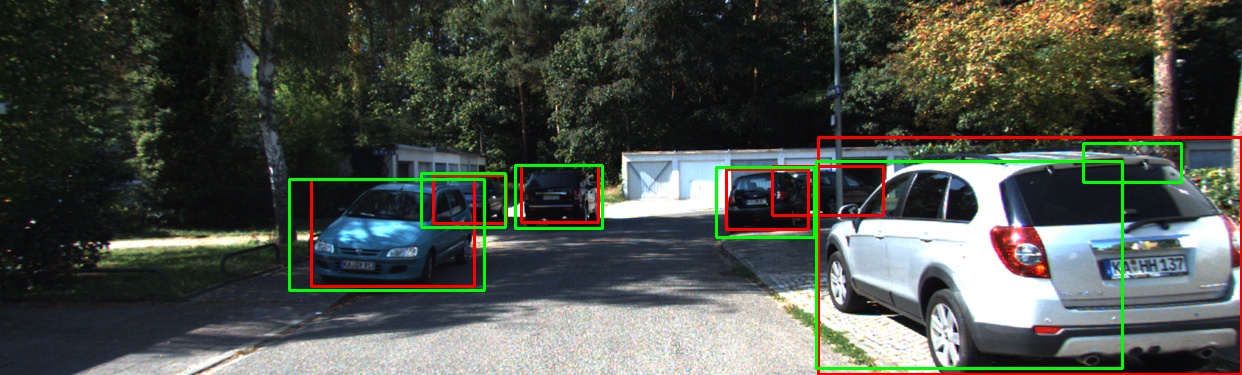}} &
\centeredcell{\includegraphics[trim=0 0 800 130,clip,height=2.2cm]{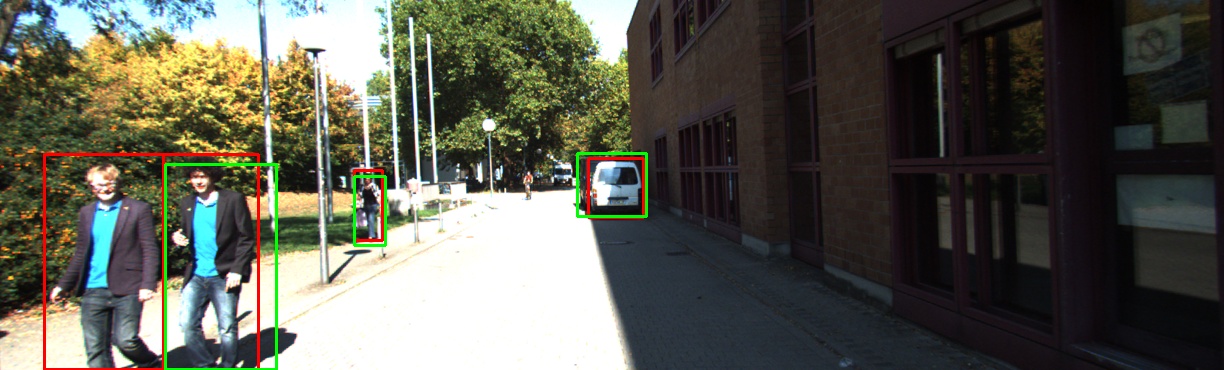}} &
\centeredcell{\includegraphics[trim=150 0 800 130,clip,height=2.2cm]{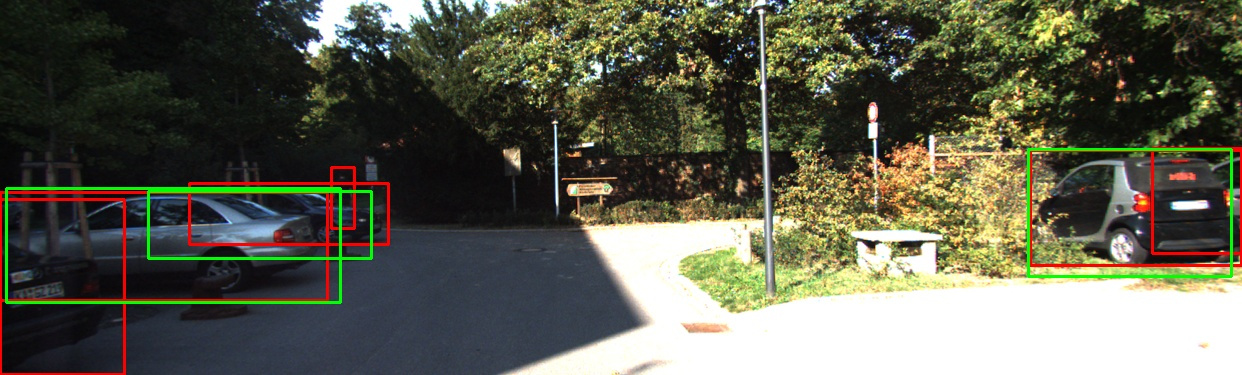}} \\
\centeredcell{\includegraphics[trim=220 50 580 120,clip,height=2.2cm]{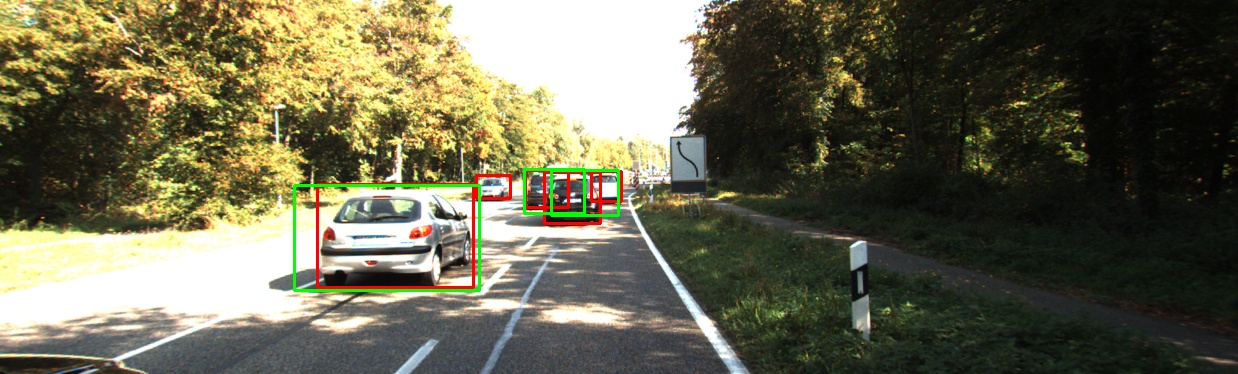}} &
\centeredcell{\includegraphics[trim=250 0 250 100,clip,height=2.2cm]{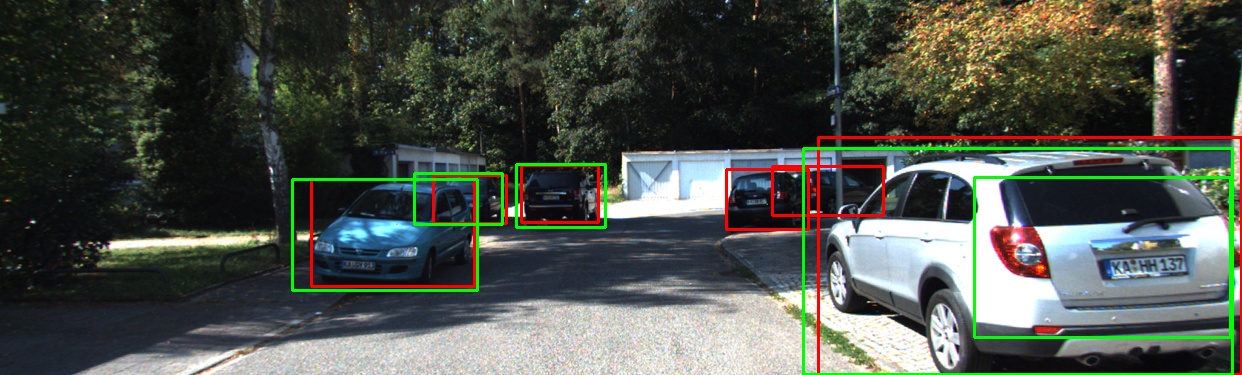}} &
\centeredcell{\includegraphics[trim=0 0 800 130,clip,height=2.2cm]{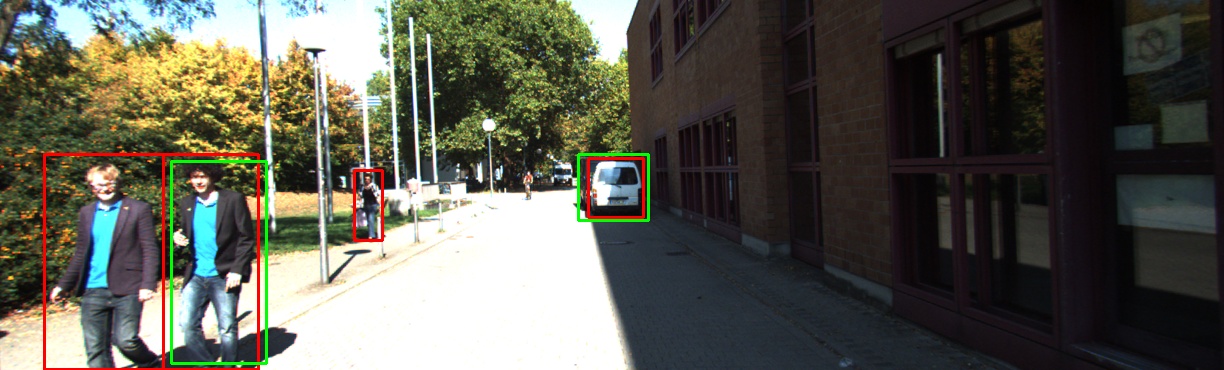}} &
\centeredcell{\includegraphics[trim=150 0 800 130,clip,height=2.2cm]{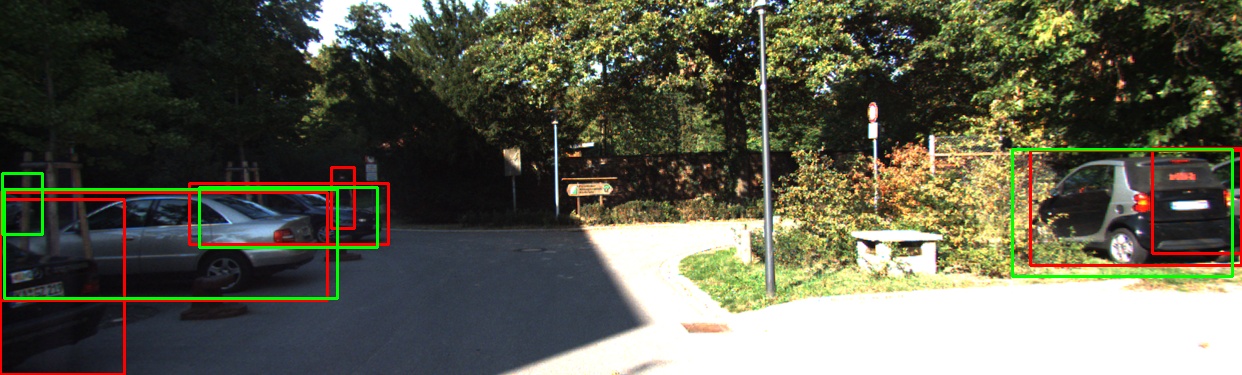}} \\
\centeredcell{\includegraphics[trim=220 50 580 120,clip,height=2.2cm]{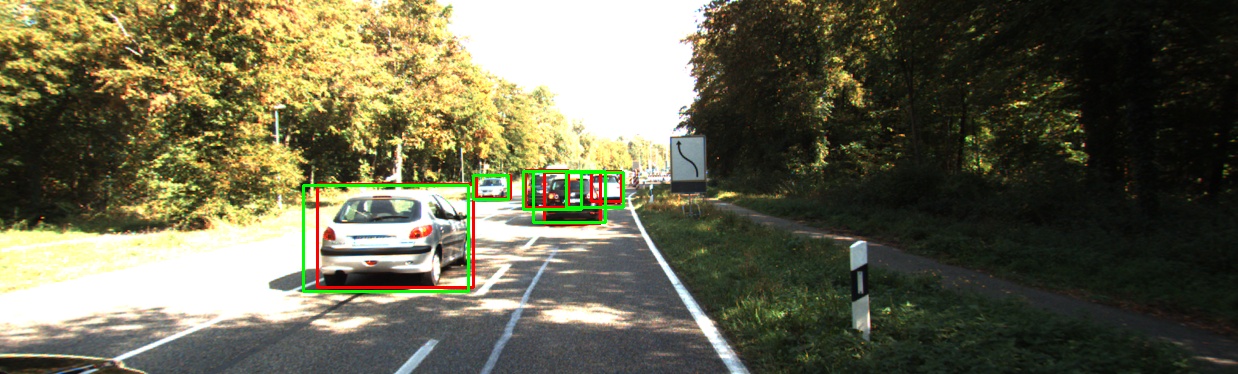}} &
\centeredcell{\includegraphics[trim=250 0 250 100,clip,height=2.2cm]{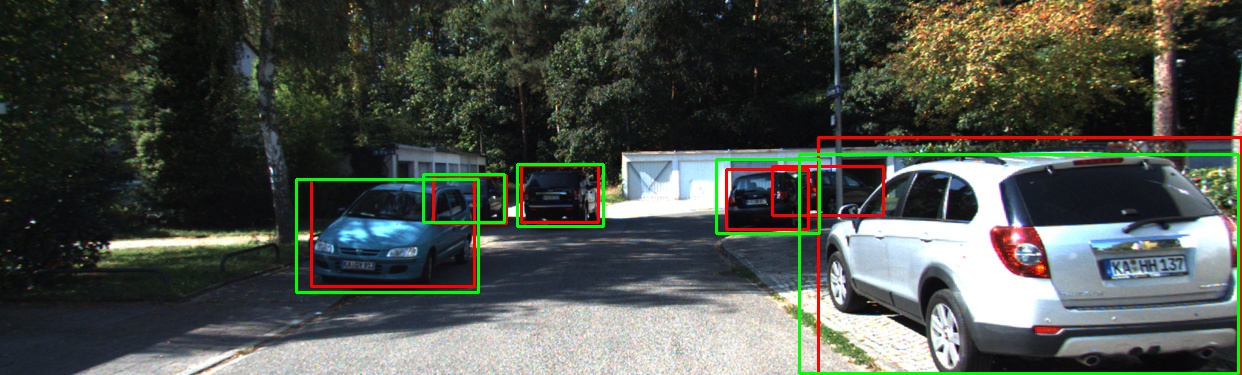}} &
\centeredcell{\includegraphics[trim=0 0 800 130,clip,height=2.2cm]{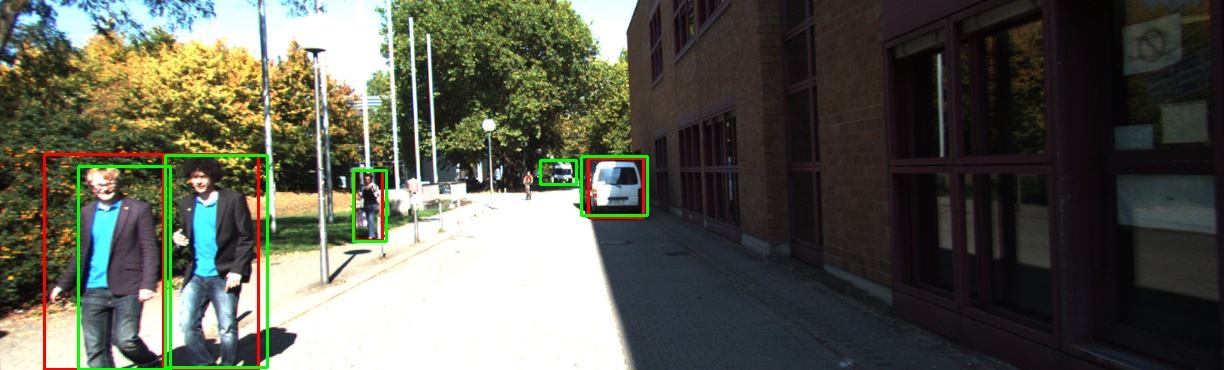}} &
\centeredcell{\includegraphics[trim=150 0 800 130,clip,height=2.2cm]{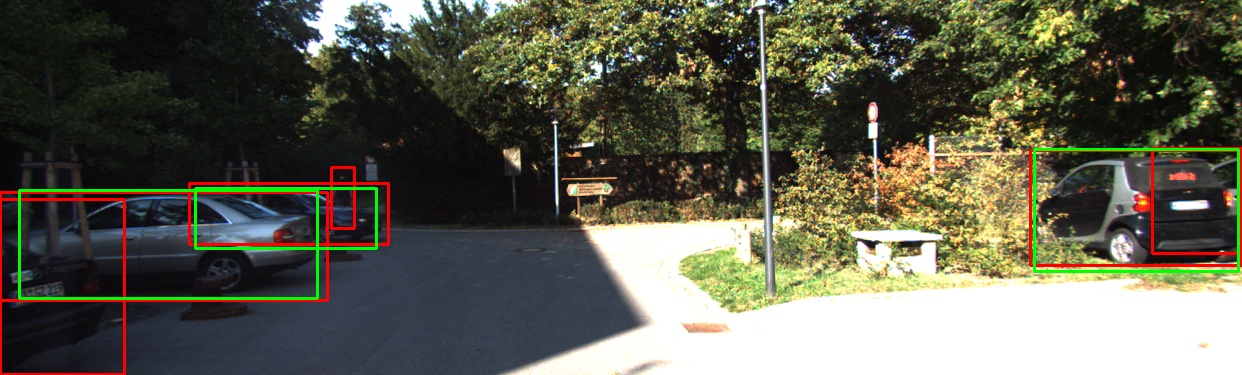}} \\
\centeredcell{\includegraphics[trim=220 50 580 120,clip,height=2.2cm]{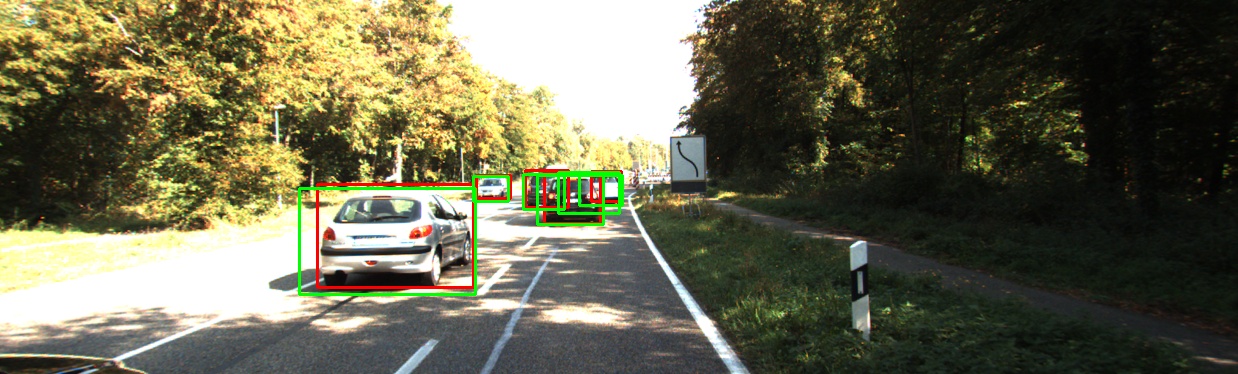}} &
\centeredcell{\includegraphics[trim=250 0 250 100,clip,height=2.2cm]{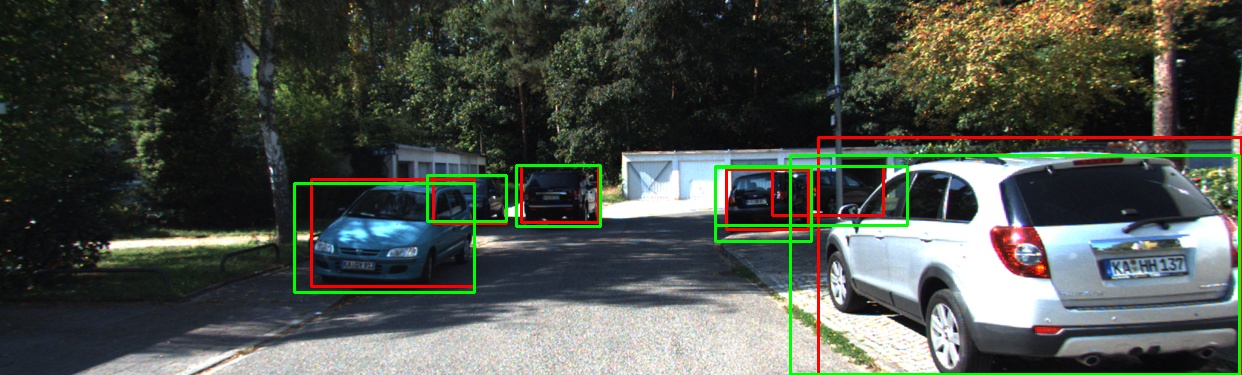}} &
\centeredcell{\includegraphics[trim=0 0 800 130,clip,height=2.2cm]{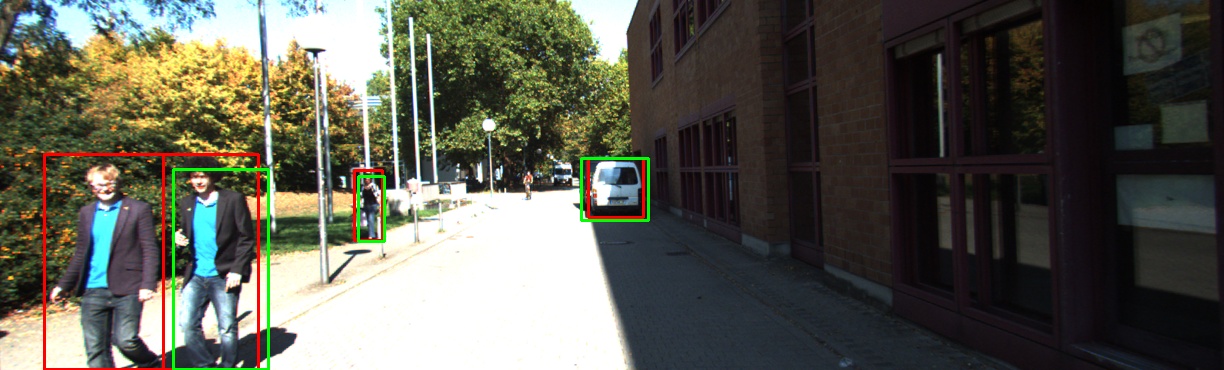}} &
\centeredcell{\includegraphics[trim=150 0 800 130,clip,height=2.2cm]{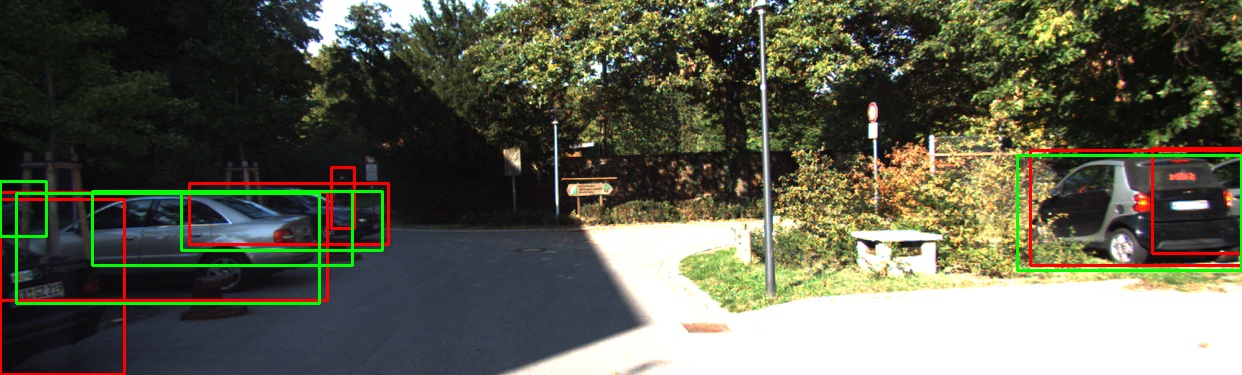}} \\
\centeredcell{\includegraphics[trim=220 50 580 120,clip,height=2.2cm]{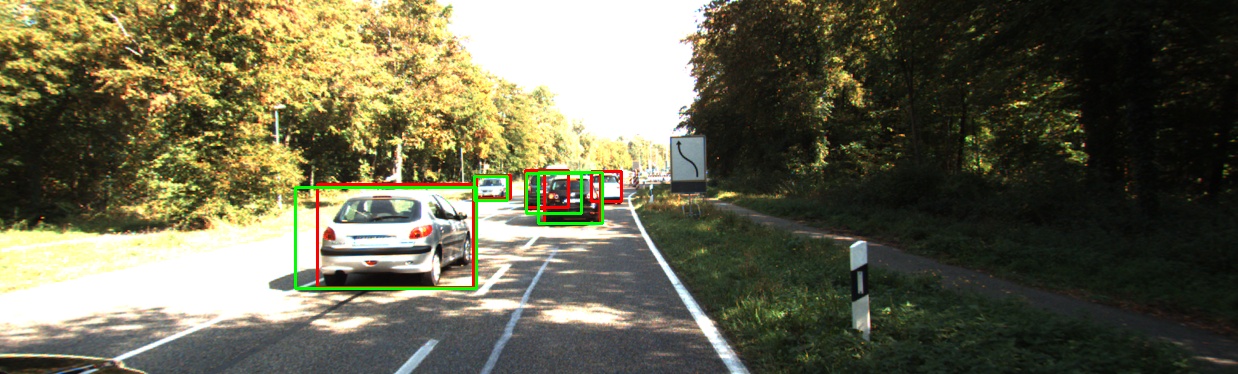}} &
\centeredcell{\includegraphics[trim=250 0 250 100,clip,height=2.2cm]{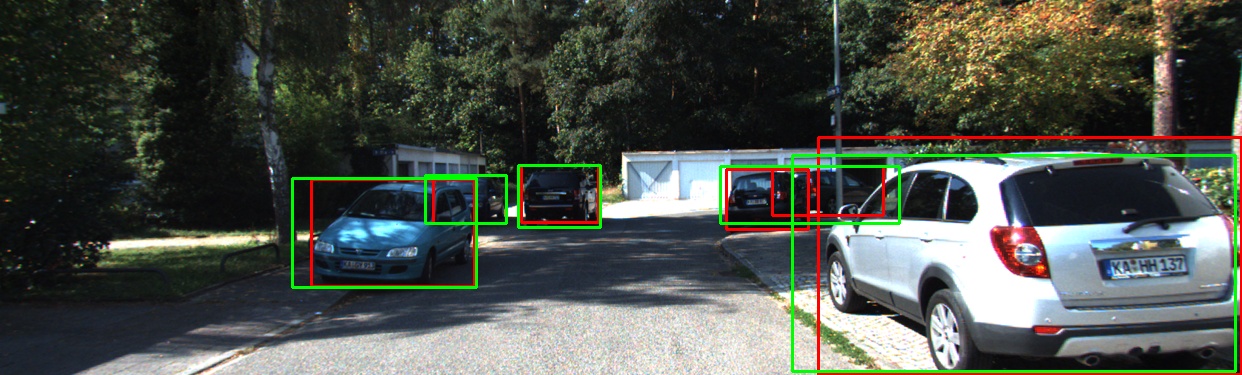}} &
\centeredcell{\includegraphics[trim=0 0 800 130,clip,height=2.2cm]{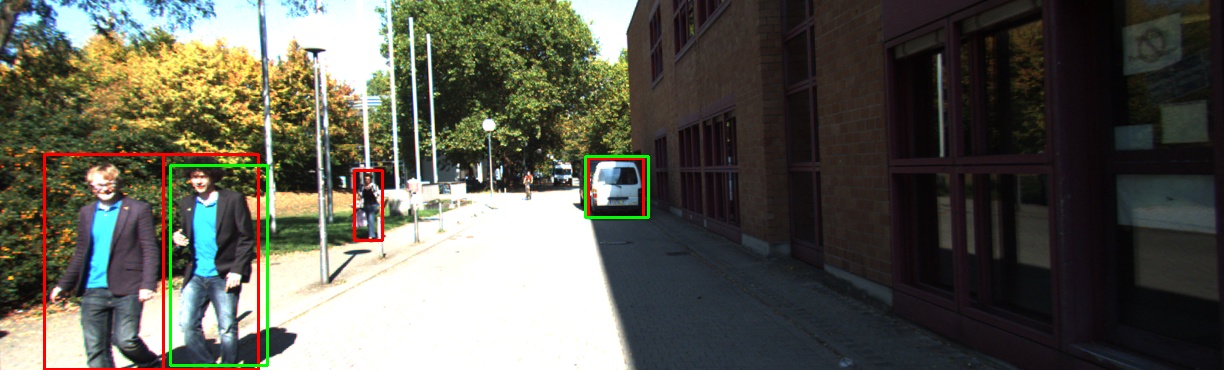}} &
\centeredcell{\includegraphics[trim=150 0 800 130,clip,height=2.2cm]{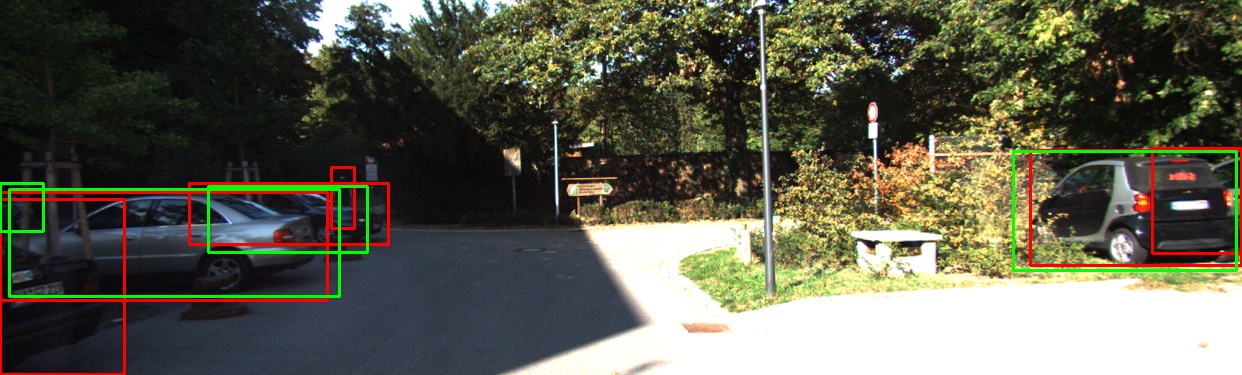}} \\
\centeredcell{\includegraphics[trim=220 50 580 120,clip,height=2.2cm]{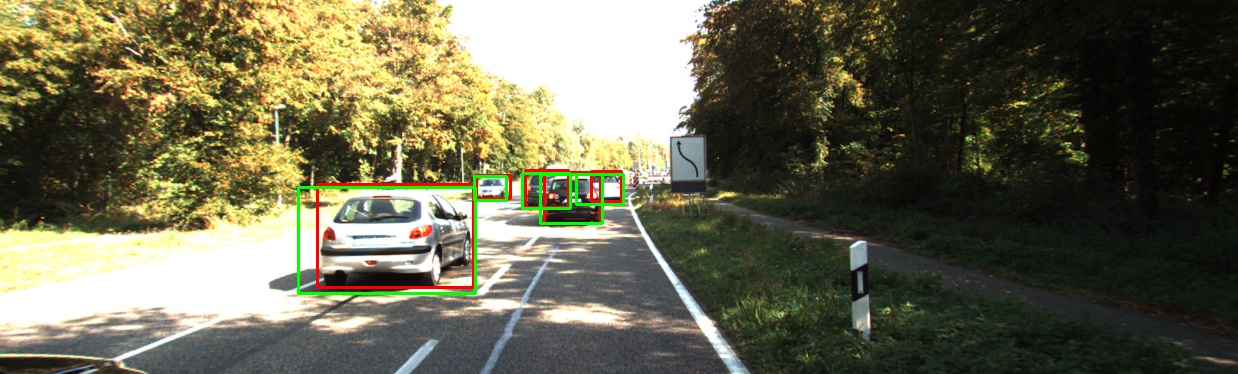}} &
\centeredcell{\includegraphics[trim=250 0 250 100,clip,height=2.2cm]{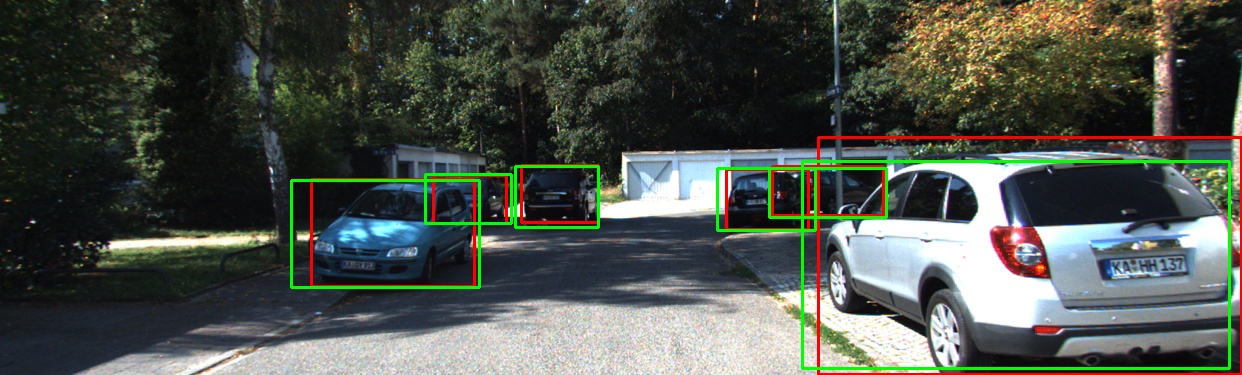}} &
\centeredcell{\includegraphics[trim=0 0 800 130,clip,height=2.2cm]{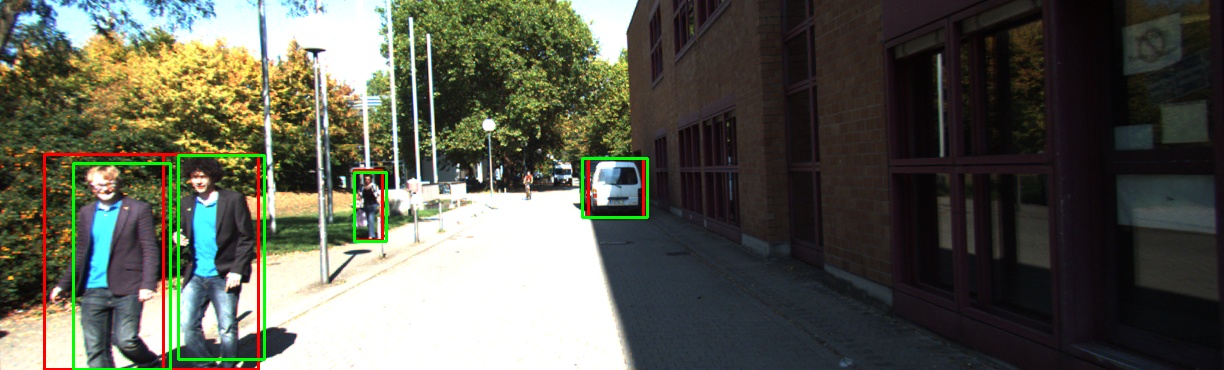}} &
\centeredcell{\includegraphics[trim=150 0 800 130,clip,height=2.2cm]{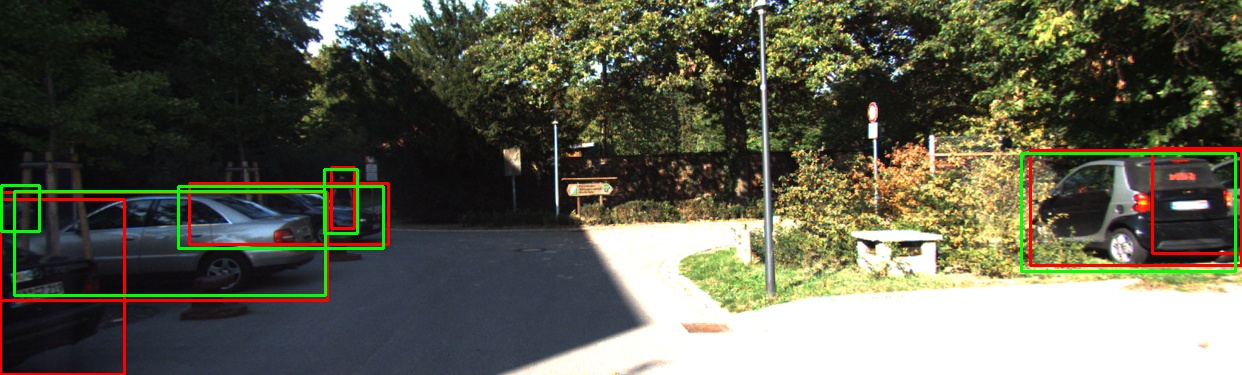}}
\end{tabular}
\caption{Object detection examples in $\Kdstest$. Red BBs correspond to the ground truth, green ones to detections according to different trained models. From top to bottom, training based on: Source ($\Vds$), Slf-T + Source, Co-T + Source, ASource ($\VGKds$), Slf-T + ASource, Co-T + ASource.} 
\label{fig:final_detection_kitti}
\end{figure*}

\begin{figure*}
\setlength{\tabcolsep}{1pt}
\centering
\begin{tabular}{cccc}
\centeredcell{\includegraphics[trim=100 0 520 70,clip,height=2.2cm]{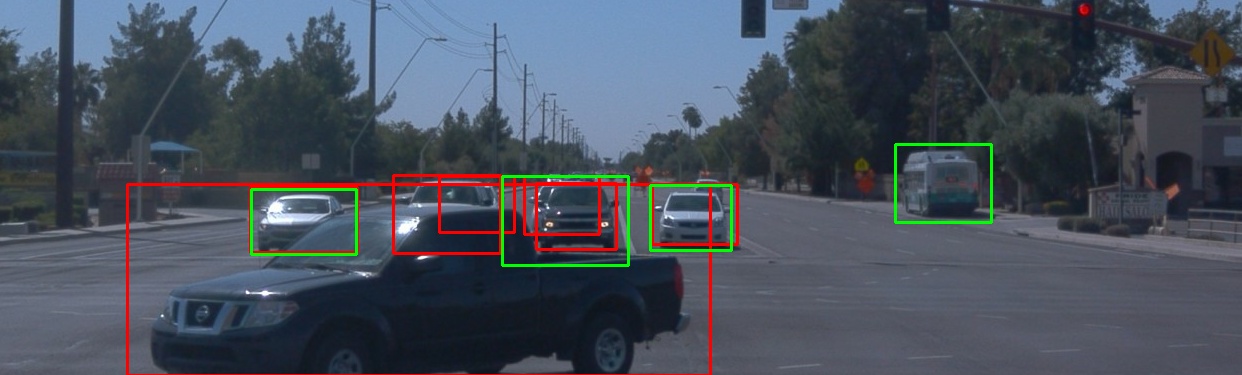}} &
\centeredcell{\includegraphics[trim=0 0 700 0,clip,height=2.2cm]{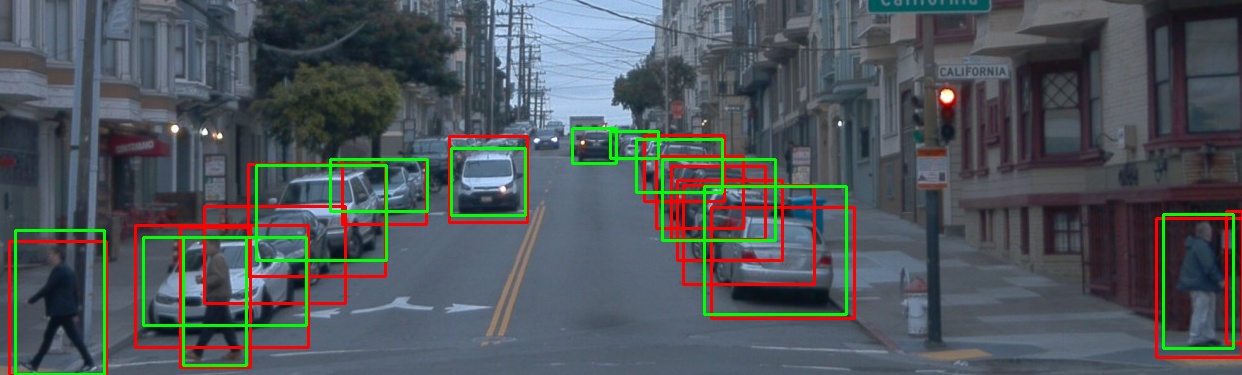}} &
\centeredcell{\includegraphics[trim=0 0 700 50,clip,height=2.2cm]{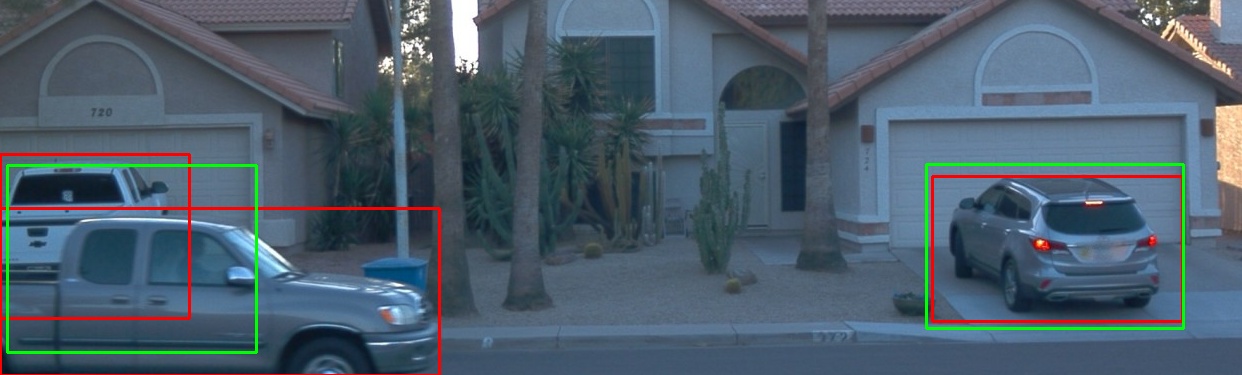}} &
\centeredcell{\includegraphics[trim=350 0 0 0,clip,height=2.2cm]{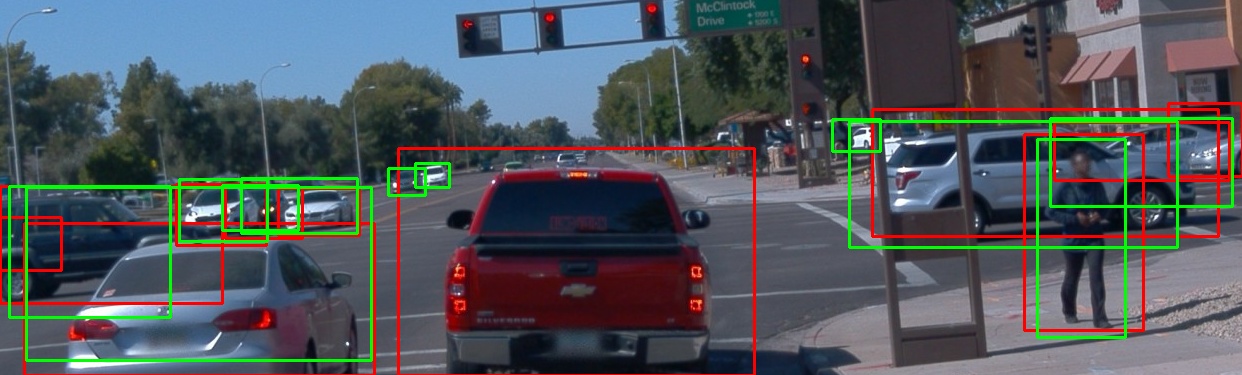}} \\
\centeredcell{\includegraphics[trim=100 0 520 70,clip,height=2.2cm]{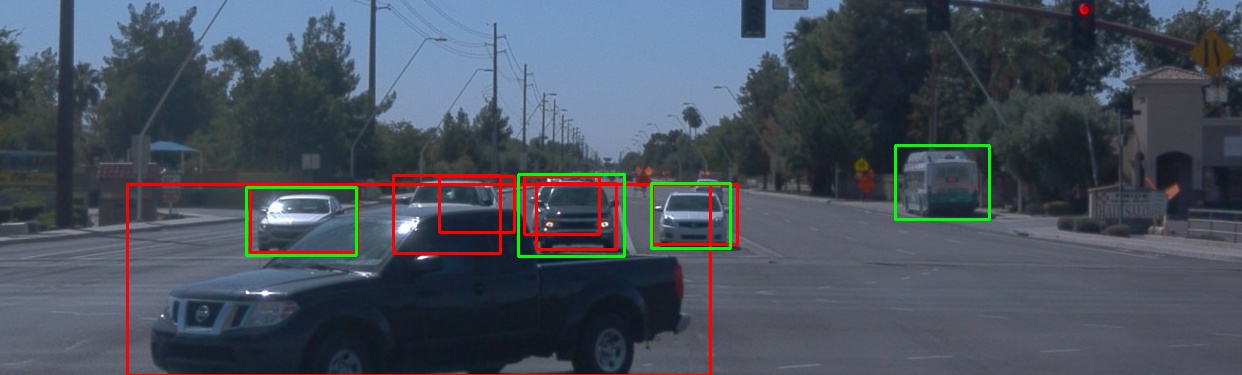}} &
\centeredcell{\includegraphics[trim=0 0 700 0,clip,height=2.2cm]{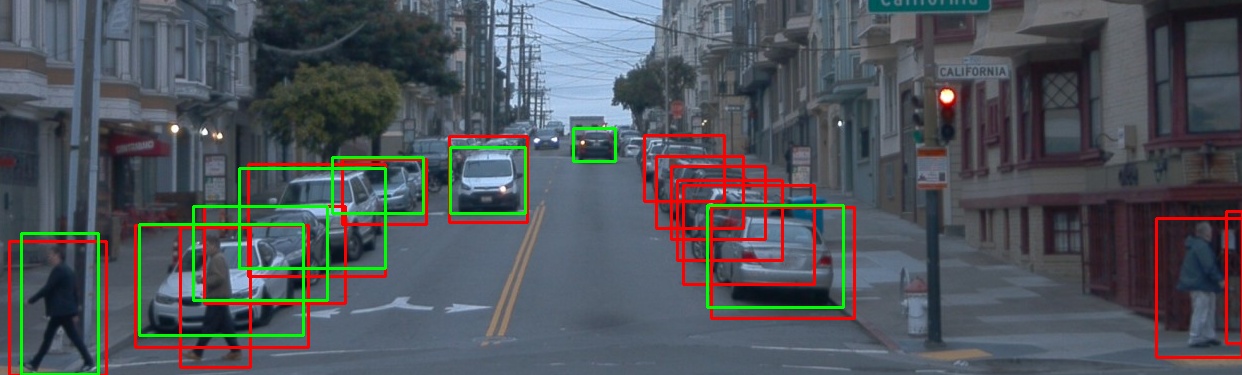}} &
\centeredcell{\includegraphics[trim=0 0 700 50,clip,height=2.2cm]{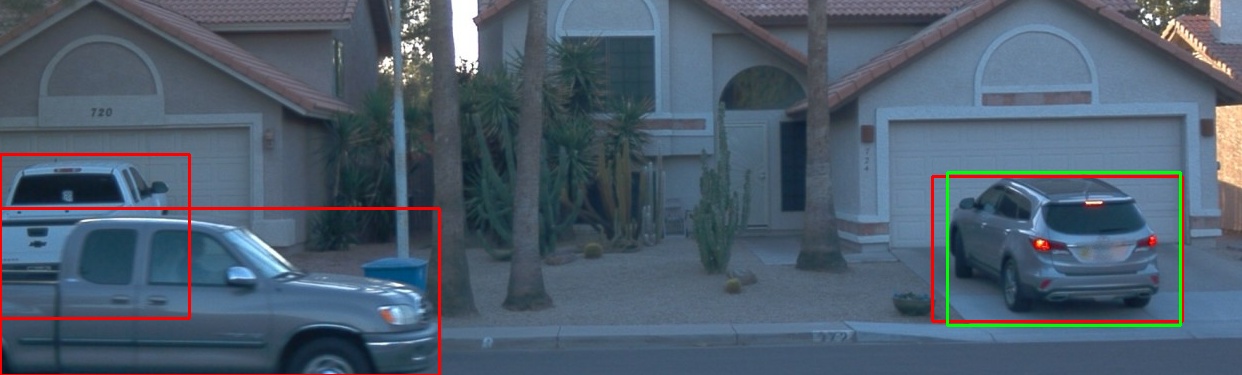}} &
\centeredcell{\includegraphics[trim=350 0 0 0,clip,height=2.2cm]{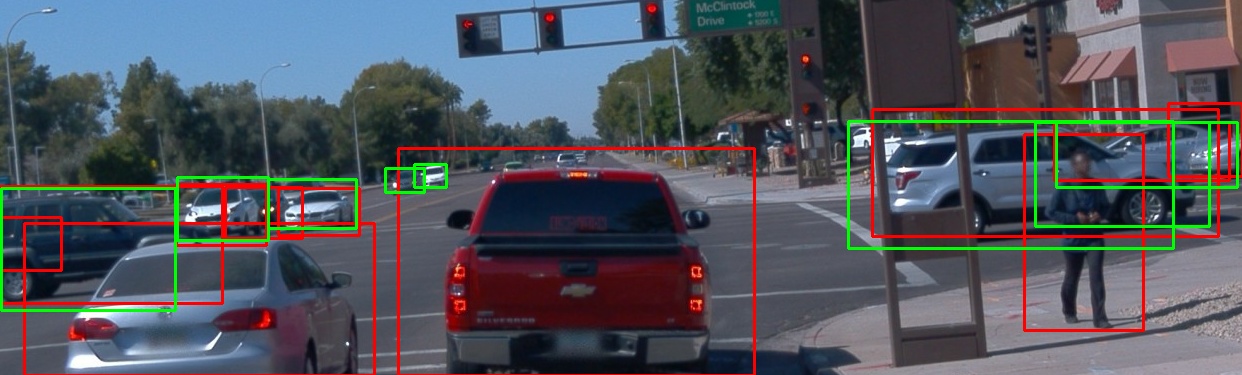}} \\
\centeredcell{\includegraphics[trim=100 0 520 70,clip,height=2.2cm]{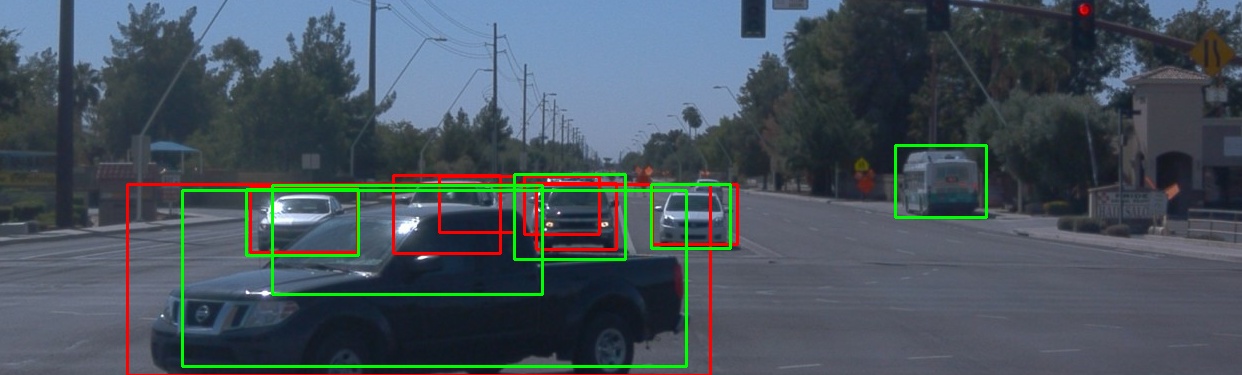}} &
\centeredcell{\includegraphics[trim=0 0 700 0,clip,height=2.2cm]{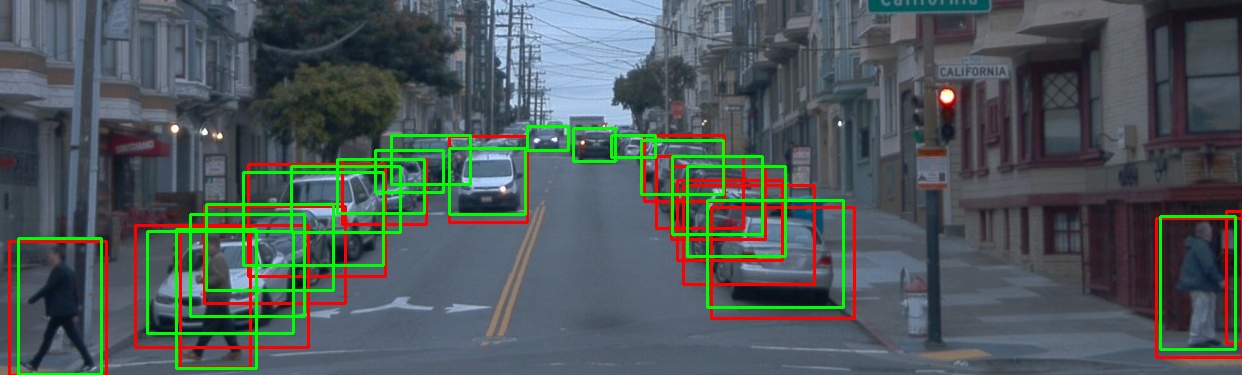}} &
\centeredcell{\includegraphics[trim=0 0 700 50,clip,height=2.2cm]{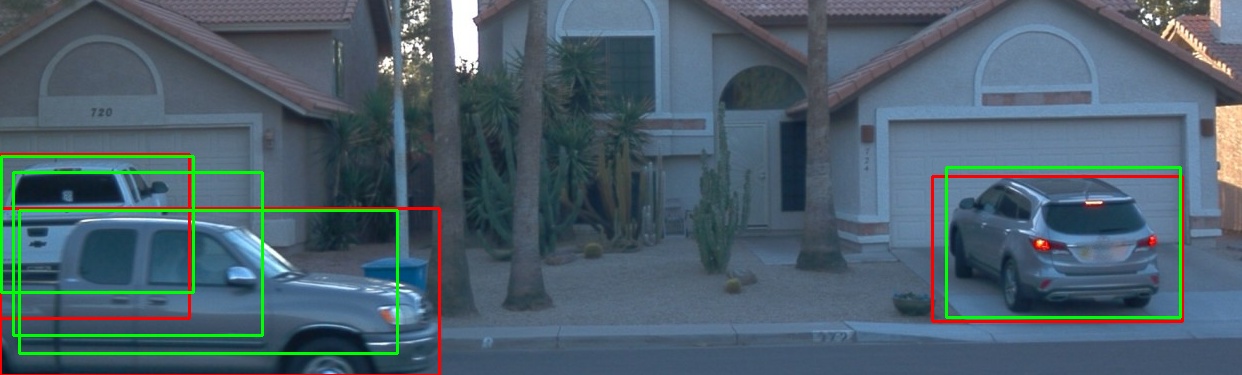}} &
\centeredcell{\includegraphics[trim=350 0 0 0,clip,height=2.2cm]{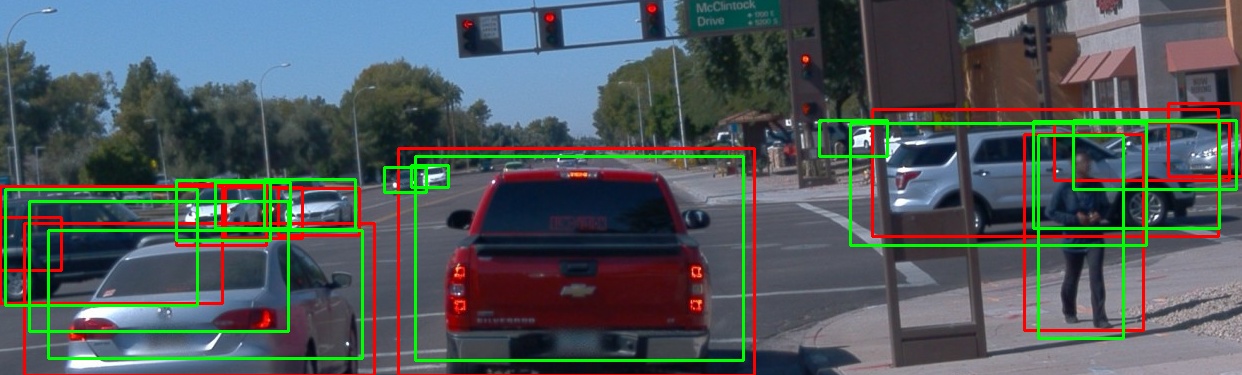}} \\
\centeredcell{\includegraphics[trim=100 0 520 70,clip,height=2.2cm]{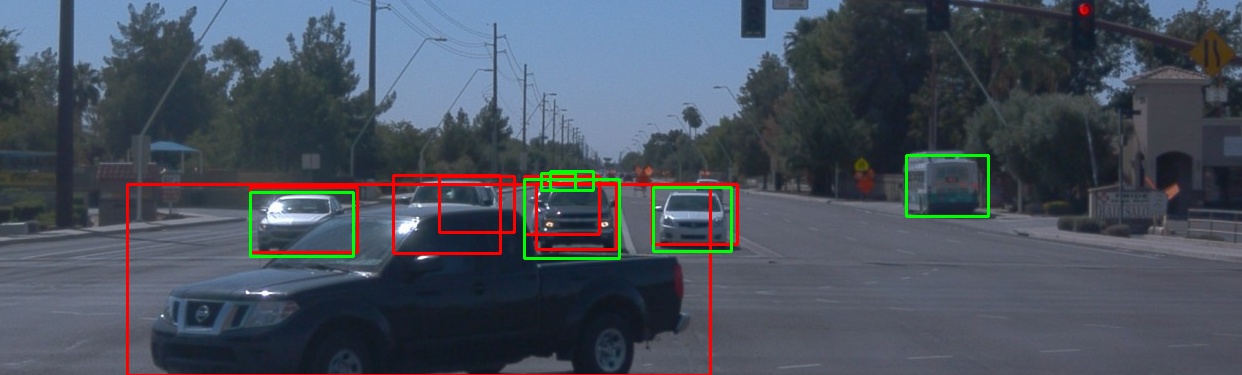}} &
\centeredcell{\includegraphics[trim=0 0 700 0,clip,height=2.2cm]{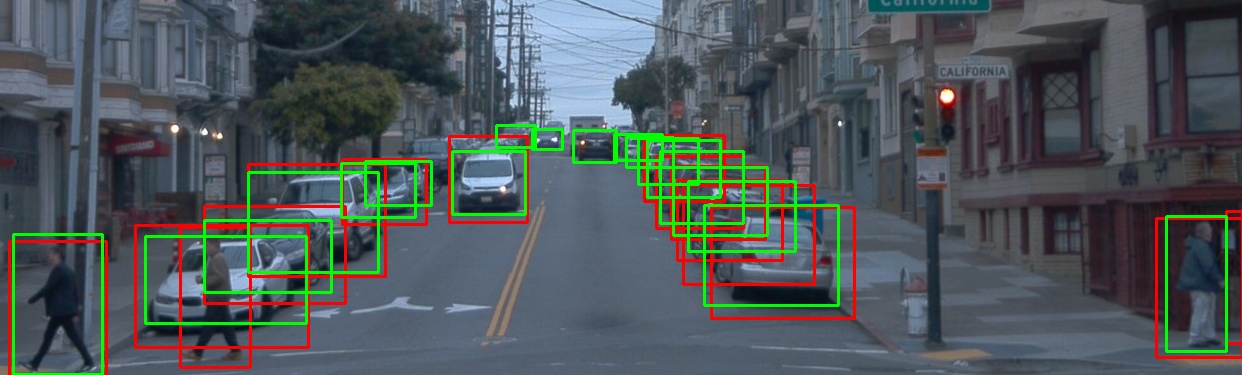}} &
\centeredcell{\includegraphics[trim=0 0 700 50,clip,height=2.2cm]{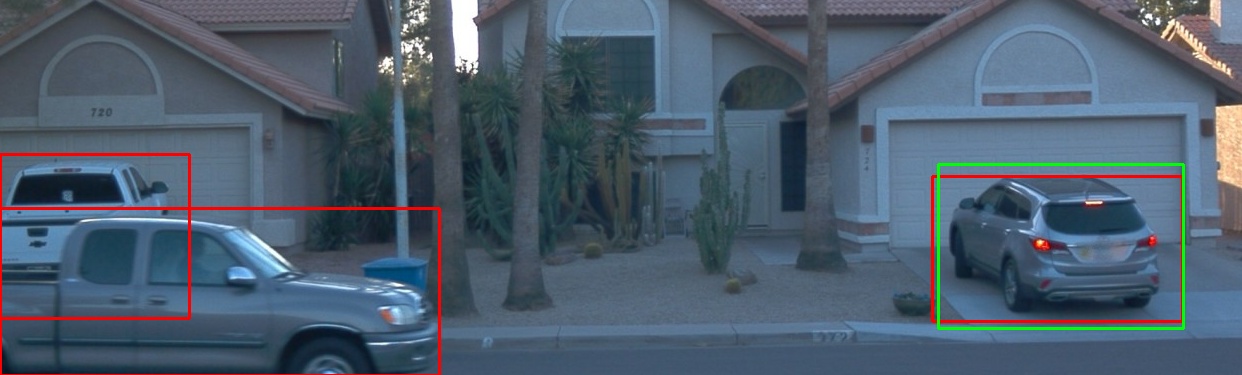}} &
\centeredcell{\includegraphics[trim=350 0 0 0,clip,height=2.2cm]{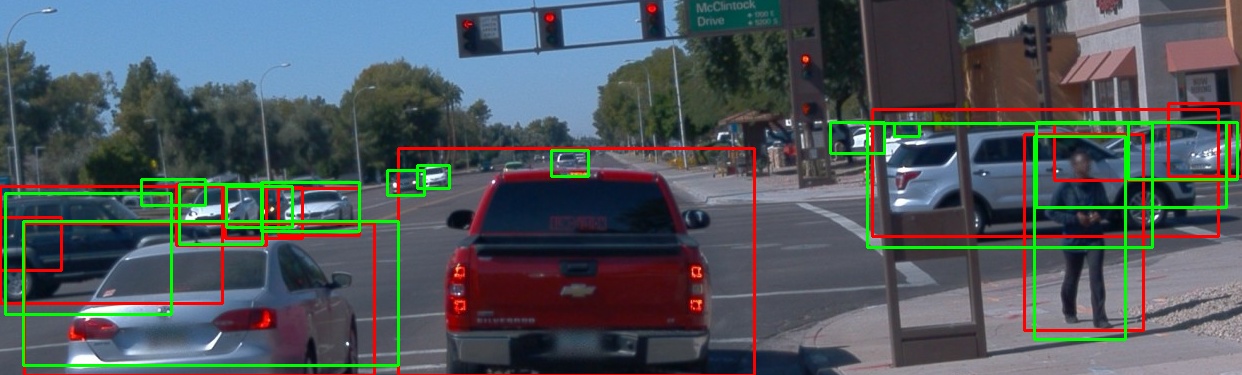}} \\
\centeredcell{\includegraphics[trim=100 0 520 70,clip,height=2.2cm]{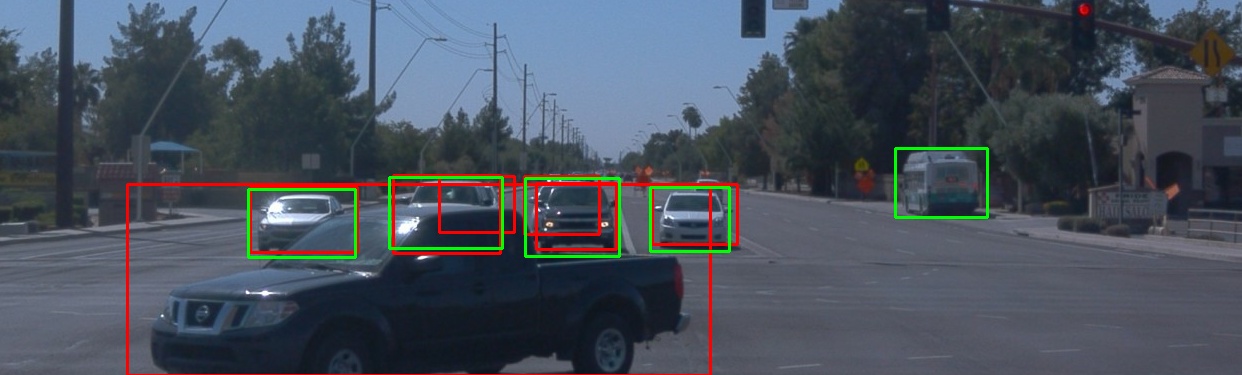}} &
\centeredcell{\includegraphics[trim=0 0 700 0,clip,height=2.2cm]{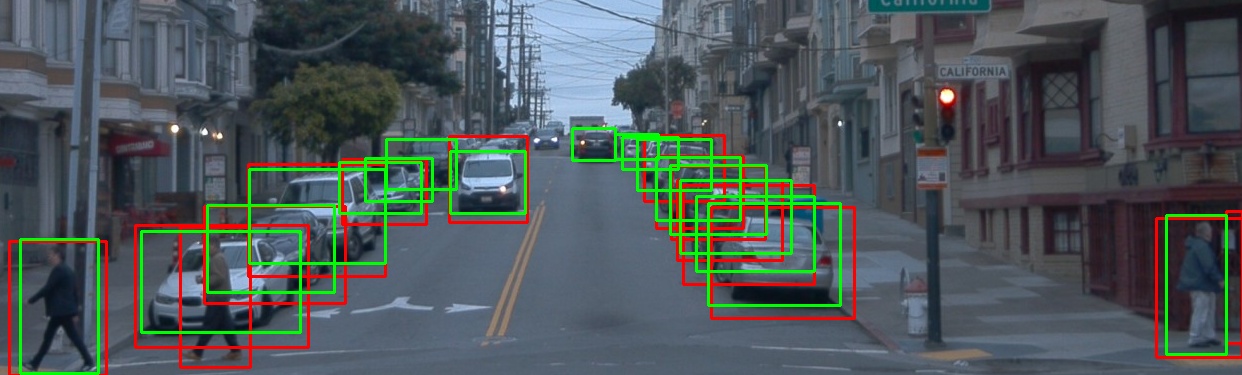}} &
\centeredcell{\includegraphics[trim=0 0 700 50,clip,height=2.2cm]{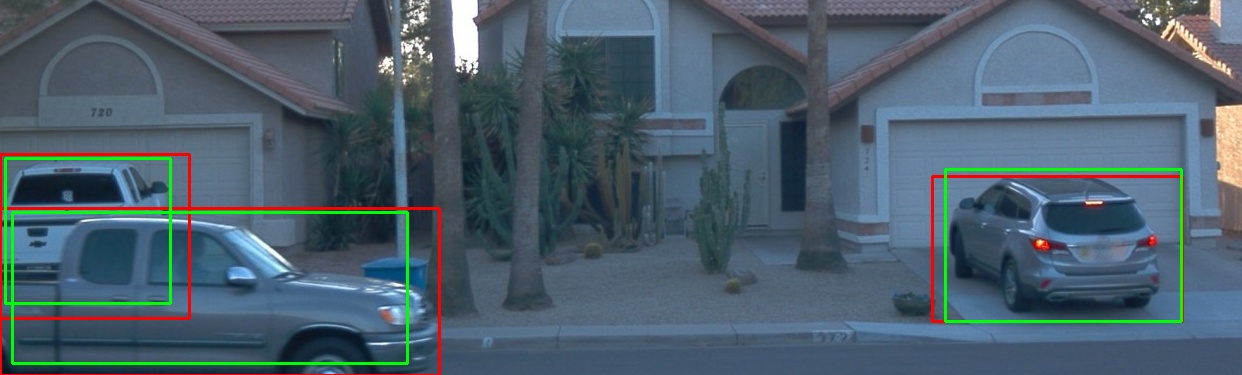}} &
\centeredcell{\includegraphics[trim=350 0 0 0,clip,height=2.2cm]{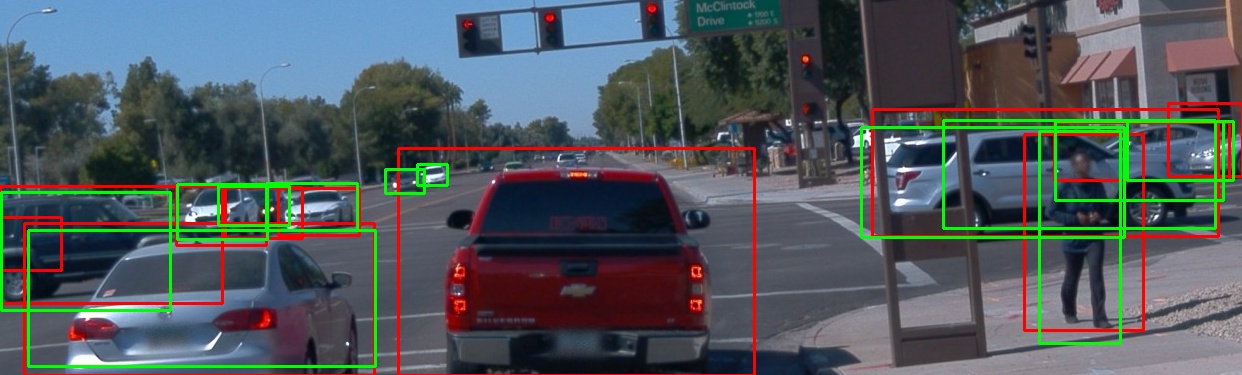}} \\
\centeredcell{\includegraphics[trim=100 0 520 70,clip,height=2.2cm]{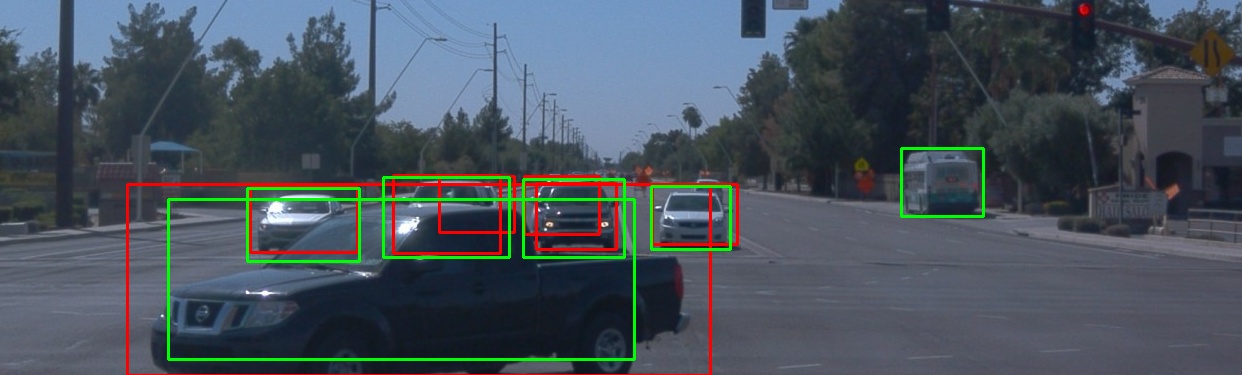}} &
\centeredcell{\includegraphics[trim=0 0 700 0,clip,height=2.2cm]{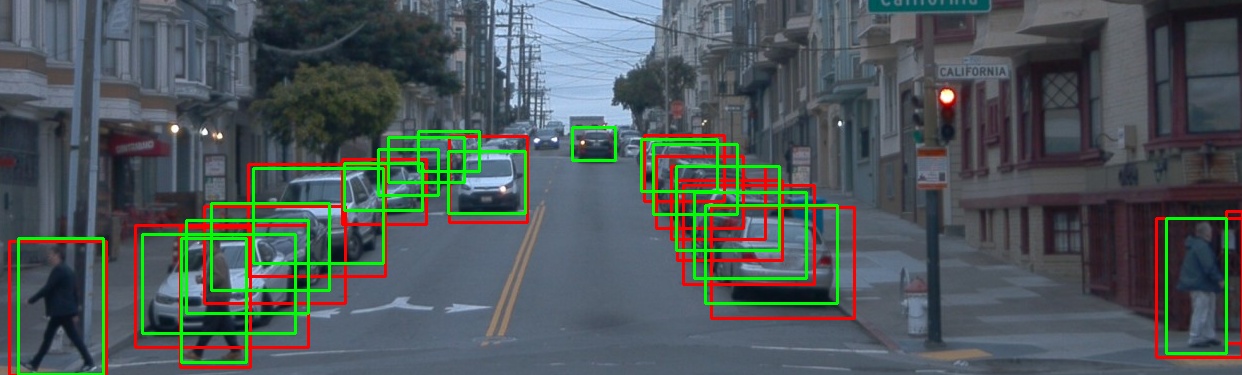}} &
\centeredcell{\includegraphics[trim=0 0 700 50,clip,height=2.2cm]{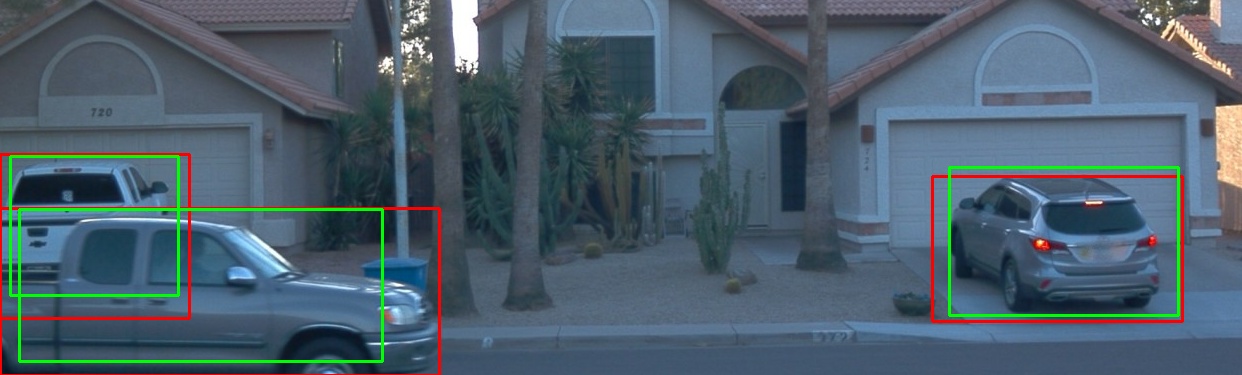}} &
\centeredcell{\includegraphics[trim=350 0 0 0,clip,height=2.2cm]{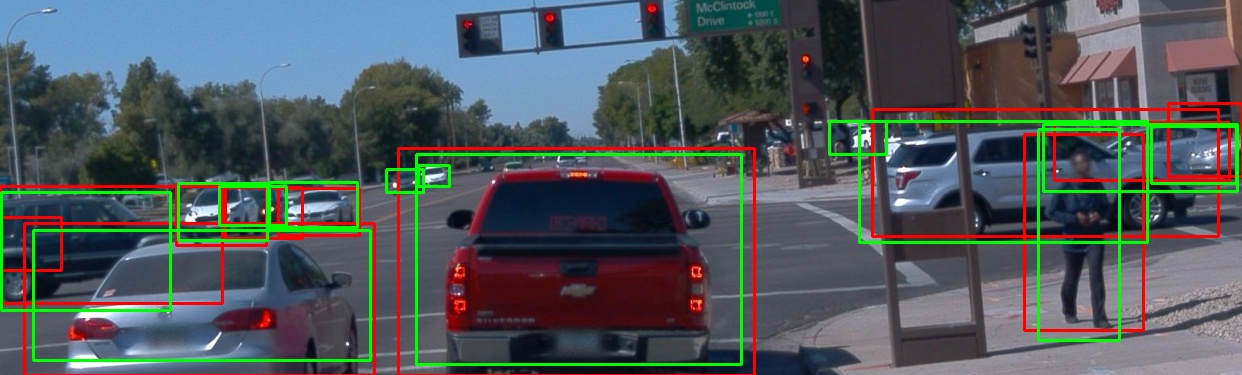}}
\end{tabular}
\caption{Analogous to Figure \ref{fig:final_detection_kitti} for $\Wdstest$ and $\VGWds$.} 
\label{fig:final_detection_waymo}
\end{figure*}

We finish by showing qualitative results on $\Kdstest$ (Figure \ref{fig:final_detection_kitti}) and $\Wdstest$ (Figure \ref{fig:final_detection_waymo}), for different object detectors. Comparing ground truth BBs and detections, we appreciate how it is confirmed what is expected from the quantitative results, {\ie} co-training combined with GAN-based image-to-image translation is providing the most accurate results.

%% file: sections/conclusion.tex
\section{Conclusion}
\label{sec:conclusion}

Motivated by the burden of manual data labeling when addressing vision-based object detection, we have explored co-training as SSL strategy for self-labeling objects in images. Moreover, we have focused on the challenging scenario where the initial labeled set is generated automatically in a virtual world; thus, co-training must actually perform UDA. We have proposed a specific co-training algorithm which is agnostic to the particular object detector used for self-labeling. We have devised a comprehensive set of experiments addressing the challenging task of on-board vehicle and pedestrian detection, using de facto standards such as KITTI and Waymo datasets, together with a virtual-world dataset introduced in this paper. Our qualitative results allow us to conclude that co-training and GAN-based image-to-image translation complement each other up to allow the training of object detectors without manual labeling, while still reaching almost the same performance as by totally relying on human labeling for obtaining upper-bound performances. These results show that the self-labeled objects are sufficient to train a well-performing object detector, but also that improving BB adjustment is convenient to improve its performance. Accordingly, our future work will address this point, for instance, by developing a multi-modal co-training which jointly explores RGB images (as in this paper) as well as depth information based on monocular depth estimation \cite{Godard:2020}, where object borders may be better localized.